\documentclass{article}
\usepackage{iclr2023_conference,times}
\iclrfinalcopy
\usepackage{amsmath,amsfonts,bm}

\def\eqref#1{equation~\ref{#1}}
\def\Eqref#1{Equation~\ref{#1}}
\def\1{\bm{1}}

\DeclareMathAlphabet{\mathsfit}{\encodingdefault}{\sfdefault}{m}{sl}
\SetMathAlphabet{\mathsfit}{bold}{\encodingdefault}{\sfdefault}{bx}{n}

\usepackage[utf8]{inputenc} % allow utf-8 input
\usepackage[T1]{fontenc}    % use 8-bit T1 fonts
\usepackage{hyperref}       % hyperlinks
\usepackage{url}            % simple URL typesetting
\usepackage{booktabs}       % professional-quality tables
\usepackage{amsfonts}       % blackboard math symbols
\usepackage{amssymb}
\usepackage{amsmath}
\usepackage{nicefrac}       % compact symbols for 1/2, etc.
\usepackage{microtype}      % microtypography
\usepackage{xcolor}         % colors
\usepackage{subcaption}
\usepackage{mathrsfs}
\usepackage{microtype}
\usepackage{array}
\usepackage{lipsum}
\usepackage{multirow}
\usepackage{placeins}
\usepackage{wrapfig}
\usepackage{tikz}

\newif\ifcomments
\commentstrue % Comment or set \commentsfalse to disable comments.

\newcommand{\blue}[1]{\textcolor{black}{#1}}
\newcommand{\red}[1]{\textcolor{black}{#1}}

\newcommand{\BEL}{BEL}
\newcommand{\CLL}{RLEL}
\newcommand{\LEL}{LEL}
\newcommand{\y}{1.51}
\newcolumntype{L}[1]{>{\raggedright\let\newline\\\arraybackslash\hspace{0pt}}m{#1}}
\newcolumntype{C}[1]{>{\centering\let\newline\\\arraybackslash\hspace{0pt}}m{#1}}
\newcolumntype{R}[1]{>{\raggedleft\let\newline\\\arraybackslash\hspace{0pt}}m{#1}}
\usepackage{algorithm}% http://ctan.org/pkg/algorithms
\usepackage{algpseudocode}% http://ctan.org/pkg/algorithmicx
\author{Deval Shah \& Tor M. Aamodt  \\
Department of Electrical and Computer Engineering\\
University of British Columbia, Vancouver, BC, Canada \\
\texttt{\{devalshah,aamodt\}@ece.ubc.ca} \\
}
\title{Learning Label Encodings for Deep Regression}

\begin{document}

\maketitle

\begin{abstract}
%Problem:
Deep regression networks are widely used to tackle the problem of predicting a continuous value for a given input. 
Task-specialized approaches for training regression networks have shown significant improvement over generic approaches, such as direct regression. 
More recently, a generic approach based on regression by binary classification using binary-encoded labels has shown significant improvement over direct regression.
The space of label encodings for regression is large.
Lacking heretofore have been automated approaches to find a good label encoding for a given application.
This paper introduces \emph{Regularized Label Encoding Learning (\CLL)} for end-to-end training of an entire network and its label encoding.
  \CLL{} provides a generic approach for tackling regression. 
%Key Insight: 
%We find that the search space of label encodings can be constrained and efficiently explored by using a combination of continuous approximation of the search space with real-valued label encodings and regularization functions. 
Underlying \CLL{} is our observation that the search space of label encodings can be constrained and efficiently explored by using a continuous search space of real-valued label encodings combined with a regularization function designed to encourage encodings with certain properties.  
These properties balance the probability of classification error in individual bits against error correction capability.
%We find that converting this discrete search space to a continuous space with real-valued label encodings enables the use of regularizers to search this design space efficiently.  
%We propose regularization rules based on properties of suitable real-valued label encodings for regression and provide corresponding regularization functions suitable for training. 
%Results: 
Label encodings found by \CLL{} result in lower or comparable errors to manually designed label encodings. 
Applying \CLL{ } results in $10.9\%$ and $12.4\%$ improvement in Mean Absolute Error (MAE) over direct regression and multiclass classification, respectively. 
Our evaluation demonstrates that \CLL{ }can be combined with off-the-shelf feature extractors and is suitable across different architectures, datasets, and tasks.  
Code is available at \url{https://github.com/ubc-aamodt-group/RLEL_regression}. 
\end{abstract}

\section{Introduction}
\label{sec:intro}
%\blue{What is the problem \\
%What are the 2-3 most closely related SOTA - just a brief selection. 2-3 closely related work- \\
%what they have not achieved?\\
% summary of what we do, how well it works, and then outline.\\
%}
Deep regression is an important problem with applications in several fields, including robotics and autonomous vehicles. 
\red{Recently, neural radiance fields (NeRF) regression networks have shown promising results in novel view synthesis, 3D reconstruction, and scene representation~\citep{liu2020neural,yu2021pixelnerf}.}
However, a typical generic approach to direct regression, in which the network is trained by minimizing the mean squared or absolute error between targets and predictions, performs poorly compared to task-specialized approaches~\citep{ssrnet,hopenet,agecnn,dorn}.
%%compared to task-specialized approaches~\citep{fsanet,ssrnet,hopenet,coralcnn,agecnn,dorn}.
Recently, generic approaches based on regression by binary classification have shown significant improvement over direct regression using custom-designed label encodings~\citep{ShahICLR2022}. 
% for regression problems~\citep{ShahICLR2022}. 
In this approach, a real-valued label is quantized and converted to an $M$-bit binary code, and these binary-encoded labels are used to train $M$ binary classifiers.
In the prediction phase, the output code of classifiers is converted to real-valued prediction using a decoding function. 
Moreover, binary-encoded labels have been proposed for ordinal regression~\citep{ordext,agecnn} and multiclass classification~\citep{multiclass,extremecode}. 
The use of binary-encoded labels for regression has multiple advantages. 
Additionally, predicting a set of values (e.g., classifiers' output) instead of one value (direct regression) introduces ensemble diversity, which improves accuracy~\citep{Song2021}.  
%%Such formulations allow the use of different loss functions to improve upon direct regression~\citep{fsanet,ssrnet}. 
Furthermore, encoded labels introduce redundancy in the label presentation, which improves error correcting capability and accuracy~\citep{ecoc}.

Finding suitable label encoding for a given problem is challenging due to the vast design space. 
Related work on ordinal regression has primarily leveraged unary codes~\citep{ordext,agecnn,dorn}.
%dorn 
Different approaches for label encoding design, including autoencoder, random search, and simulated annealing, have been proposed to design suitable encoding for multiclass classification~\citep{extremecode,ecoc,Song2021}. 
However, these encodings perform relatively poorly for regression due to differences in task objectives (Section~\ref{sec:rel}). 
More recently,~\citet{ShahICLR2022} analyzed and proposed properties of suitable encodings for regression.
They empirically demonstrated the effectiveness of manually designed encodings guided by these properties. 
While establishing the benefits of exploring the space of label encodings for a given task, they did not provide an automated approach to do so.  %search through this discrete space without human intervention for a given benchmark. 

In this work, we propose \emph{Regularized Label Encoding Learning (\CLL)}, an %automated 
end-to-end approach to train the network and label encoding together. 
%\devalh{Maybe the description and how we reach continuous space can be improved. First discrete is difficult: so we use relaxation. Further, we notice that the regularizer and decoder both use real values, so we do not need to binarize the output values.}
%The search space of binary label encodings is discrete. 
Binary-encoded labels have discrete search space. 
This work proposes to relax the assumption of using discrete search space for label encodings. 
%We find that 
Label encoding design can be approached by regularized search through a continuous space of real-valued label encodings, enabling the use of continuous optimization approaches. 
Such a formulation enables end-to-end learning of the network parameters and label encoding. 

We propose two regularization functions to encourage certain properties in the learned label encoding during training. 
Specifically, while operating on real-valued label encoding, the regularization functions employed by \CLL{} are designed to encourage properties 
previously identified as being helpful for binary-valued label encodings~\citep{ShahICLR2022}. 
The first property encourages the distance between learned encoded labels to be proportional to the difference between corresponding label values, which reduces the regression error. 
Further, each bit of label encoding can be considered a binary classifier. 
%%The second property proposes to reduce the complexity of each binary classifier's decision boundary by reducing the distance between label encodings for adjacent label values. 
The second property proposes to reduce the complexity of a binary classifier's decision boundary by reducing the number of bit transitions ($0\rightarrow 1$ and $1 \rightarrow 0$ transitions in the classifier's target over the range of labels) in the corresponding bit in binarized label encoding. 
%%%\blue{\cite{speactral} demonstrated showed that the complexity of a function can be measured by frequency components present in it. The frequency of bit transitions introduced the term \emph{spectral bias}, and demonstrated that neural networks prioritize learning low complexity (i.e., low frequency) functions. 
%%%Thus, lowering the frequency of bit transitions improves accuracy. }
  
\iftrue
{Figure~\ref{fig:intror} demonstrates the effect of proposed regularizers on the learned label encodings and regression errors. Figure~\ref {fig:intror1} plots the L1 distance between learned encodings for different target labels versus the difference between corresponding label values. The L1 distance between encodings for distant targets is low without the regularizer. In contrast, the proposed regularizer encourages the learned label encoding to follow the first design property. 
Figure~\ref{fig:intror2} plots learned label encoding (binarized representation for clarity). 
Each row represents encoding for a target value, and each column represents a classifier's output over the range of target labels. 
The use of regularizer R2 reduces the number of bit-transitions (i.e., $1\rightarrow 0$ and $0\rightarrow 1 $ transitions in a column) to enforce the second design property and consequently reduces the regression error.  }

\begin{figure}[t]
    \centering
    \begin{subfigure}[t]{0.38\linewidth}
        \centering
        \includegraphics[width=\textwidth]{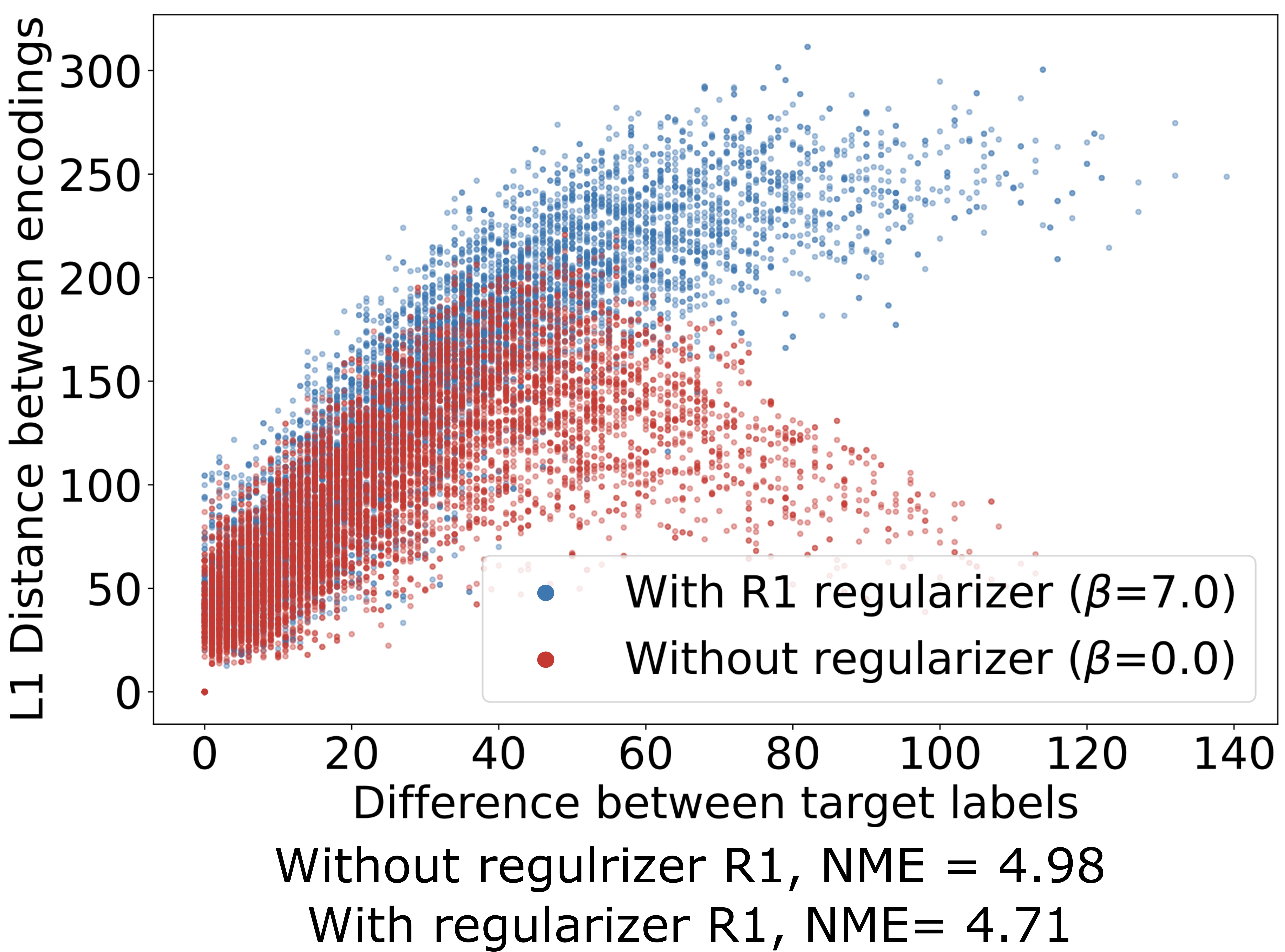}
        \caption{Effect of regularizer R1}
        \label{fig:intror1}
    \end{subfigure}
    \hfill
    \begin{subfigure}[t]{0.55\textwidth}
      \centering
      \includegraphics[width=\textwidth]{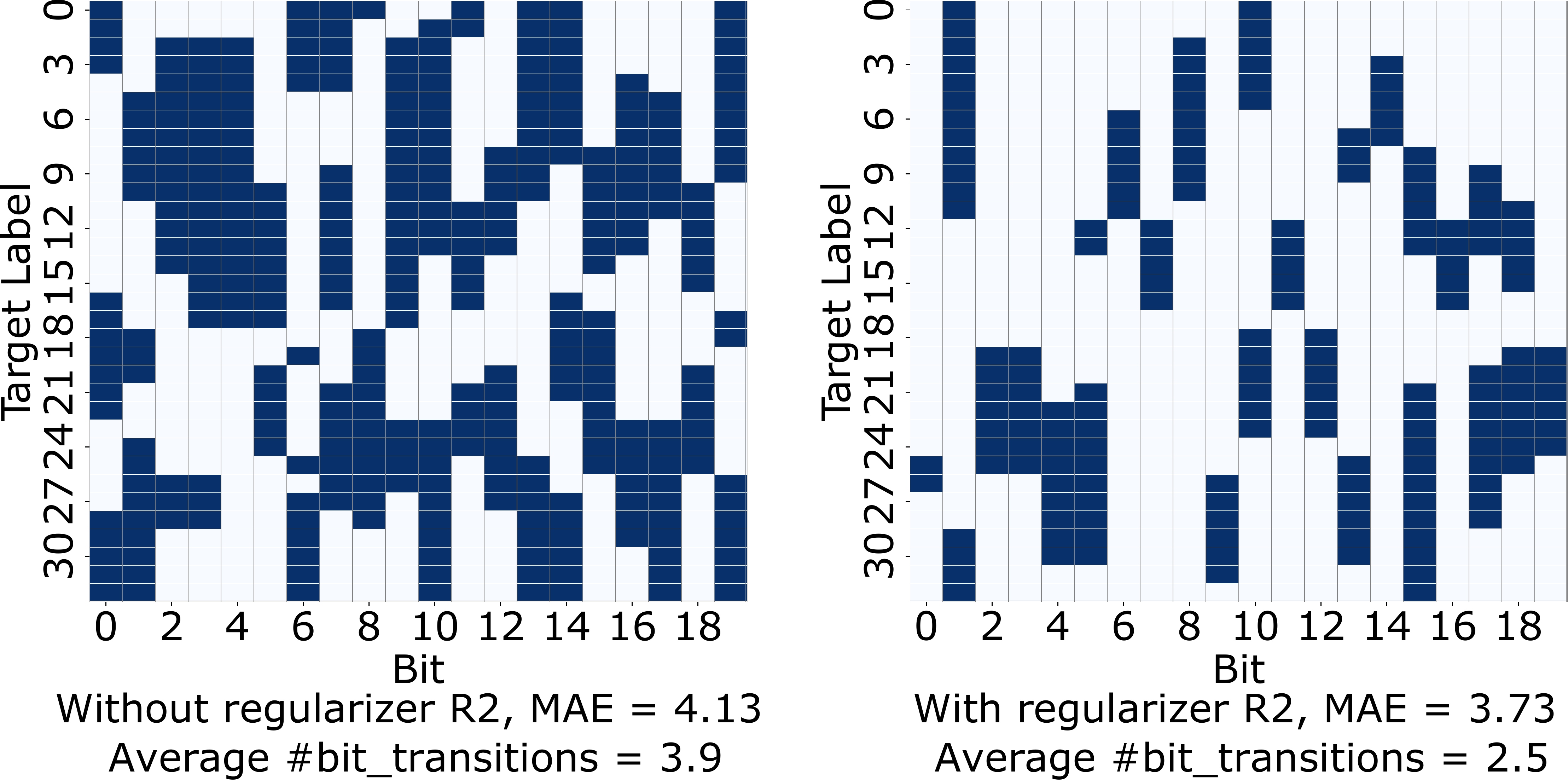}
      \caption{Effect of regularizer R2}
      \label{fig:intror2}
    \end{subfigure}
    \caption{(a) and (b) demonstrate the effect of proposed regularizers on learned label encodings and the regression error (NME/MAE) for FLD1\_s and LFH1 benchmarks (Table~\ref{tab:benchmarks}), respectively. (a) Regularizer R1 encourages the distance between learned encodings to be proportional to the difference between corresponding label values. (b) Regularizer R2 reduces the number of bit transitions per bit, reducing the complexity of decision boundaries to be learned by binary classifiers. Here blue and white colors represent $1$ and $0$, respectively.  }
  \label{fig:intror}
  \end{figure}
\fi
%Prior works have shown that using prior knowledge about the task to constrain or regularize the parameter space can improve accuracy and generalization. 
We demonstrate that the regularization approach employed by \CLL{} encourages the desired properties in the label encodings.  
We evaluate the proposed approach on $11$ benchmarks, covering diverse datasets, network architectures, and regression tasks, such as head pose estimation, facial landmark detection, age estimation, and autonomous driving. 
Label encodings found by \CLL{ }result in lower or comparable errors to manually designed codes and outperform generic encoding design approaches 
%Label encodings found by \CLL{ }outperform simulated annealing-, autoencoder-, and hand-designed label encodings
% by x\%, y\%, and z\%, respectively
~\citep{1057277,extremecode,ShahICLR2022}. 
Further, \CLL{ }results in lower error than direct regression and multiclass classification by $10.9\%$ and $12.4\%$, respectively, and even outperforms several task-specialized approaches.  
We make the following contributions to this work:
\begin{itemize}
    \item We provide an efficient label encoding design approach by combining regularizers with continuous search space of label encodings.
    \item We analyze properties of suitable encodings in the continuous search space and propose regularization functions for end-to-end learning of network parameters and label encoding. 
    \item We evaluate the proposed approach on $11$ benchmarks and show significant improvement over different encoding design methods and generic regression approaches.  
\end{itemize}
\section{Background and Related Work}
\label{sec:rel}
This section summarizes relevant background information on regression by binary classification approach and different code design approaches. 
%This section summarizes relevant background information on regression by binary classification approach and binary-encoded labels of regression. Further, we summarize related work on different code design approaches. 
%Task-specific regression approaches are also well-explored (summarized in Appendix~\ref{sec:a3}). 
Task-specific regression approaches are summarized in Appendix~\ref{sec:a3}. 
%However, task-specific approaches do not apply to all tasks, and a generic regression approach is desirable. 
However, a generic regression approach applicable to all tasks is desirable. 
\subsection{Regression by Binary Classification}
\label{subsec:prop}
%\paragraph{Regression by Binary Classification}
%\label{subsec:prop}
A regression problem can be converted to a set of binary classification subproblems. 
Prior works proposed to use $N$ binary classifiers for scaled and quantized target labels $\in \{1,2,...,N\}$~\citep{agecnn,dorn}. 
Here, classifier-$k$'s target output is $1$ if the target label is greater than $k$, else $0$. 
Figure~\ref{fig:cu} represents the target output of binary classifiers for this setup. 
%Here, each target label is converted to a unary code. 
\citet{ShahICLR2022} proposed \emph{Binary-encoded Labels (BEL)}, a generalized framework for regression by binary classification. 
In the proposed approach, a real-valued target label is quantized and converted to a binary code $B$ of length $M$ using an encoding function $\mathcal{E}$. 
$M$ binary classifiers are trained using binary-encoded target labels $B \in \{0,1\}^M$. 
During inference, the output of binary classifiers, i.e., predicted code, is converted to the real-valued prediction using a decoding function $\mathcal{D}$. 

{
\captionsetup[sub]{font=footnotesize, skip=5pt, belowskip=-5pt}
\newdimen\imageheight
\settoheight{\imageheight}{%
\includegraphics[width=0.22\textwidth]{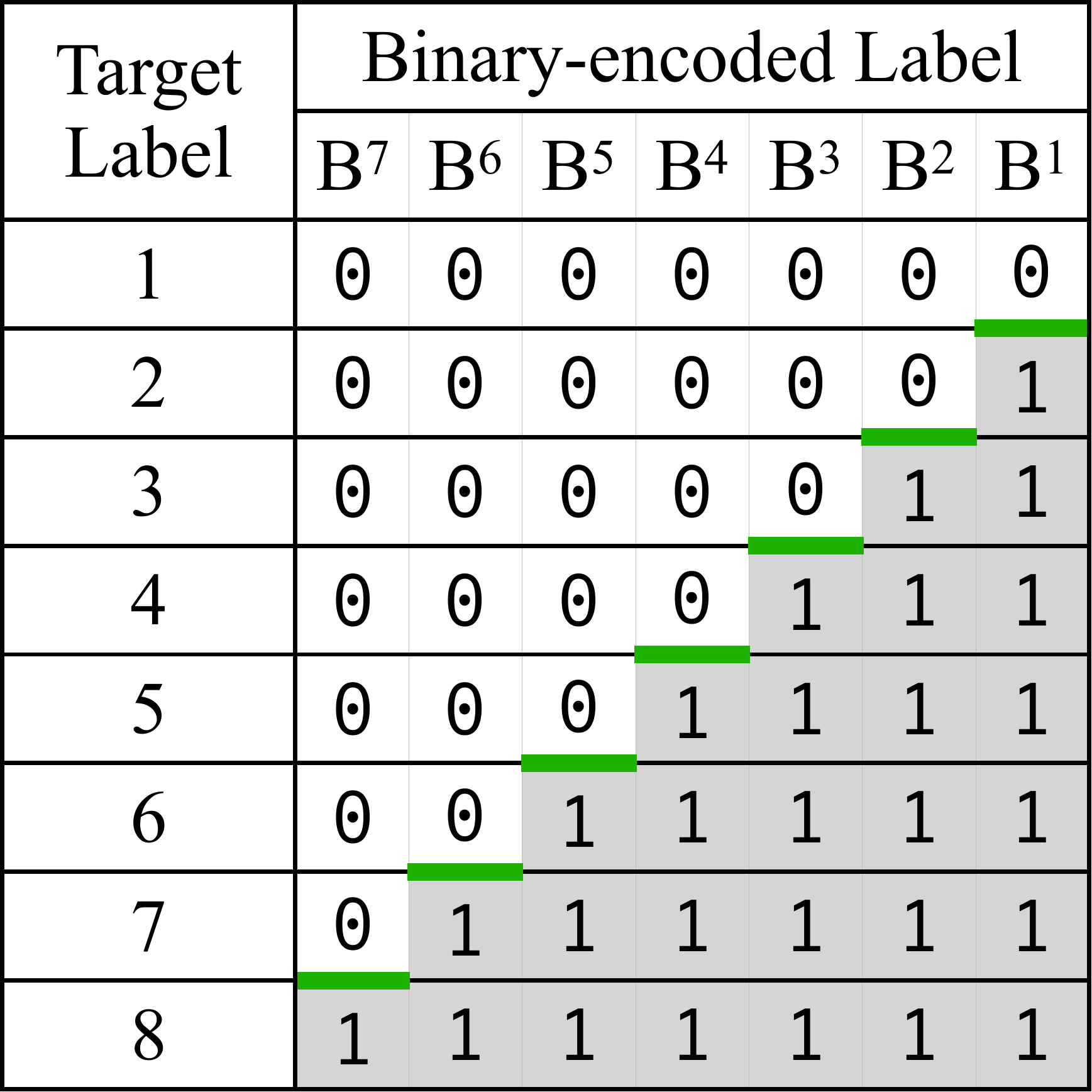}}
\begin{figure}[t]
  \centering
  \begin{subfigure}[t]{0.25\linewidth}
    \centering
  \includegraphics[height=\imageheight]{figures/codeu.pdf}
 \caption{Unary encoding}
 \label{fig:cu}
\end{subfigure}
\hfill
  \begin{subfigure}[t]{0.25\linewidth}
      \centering
    \includegraphics[height=\imageheight]{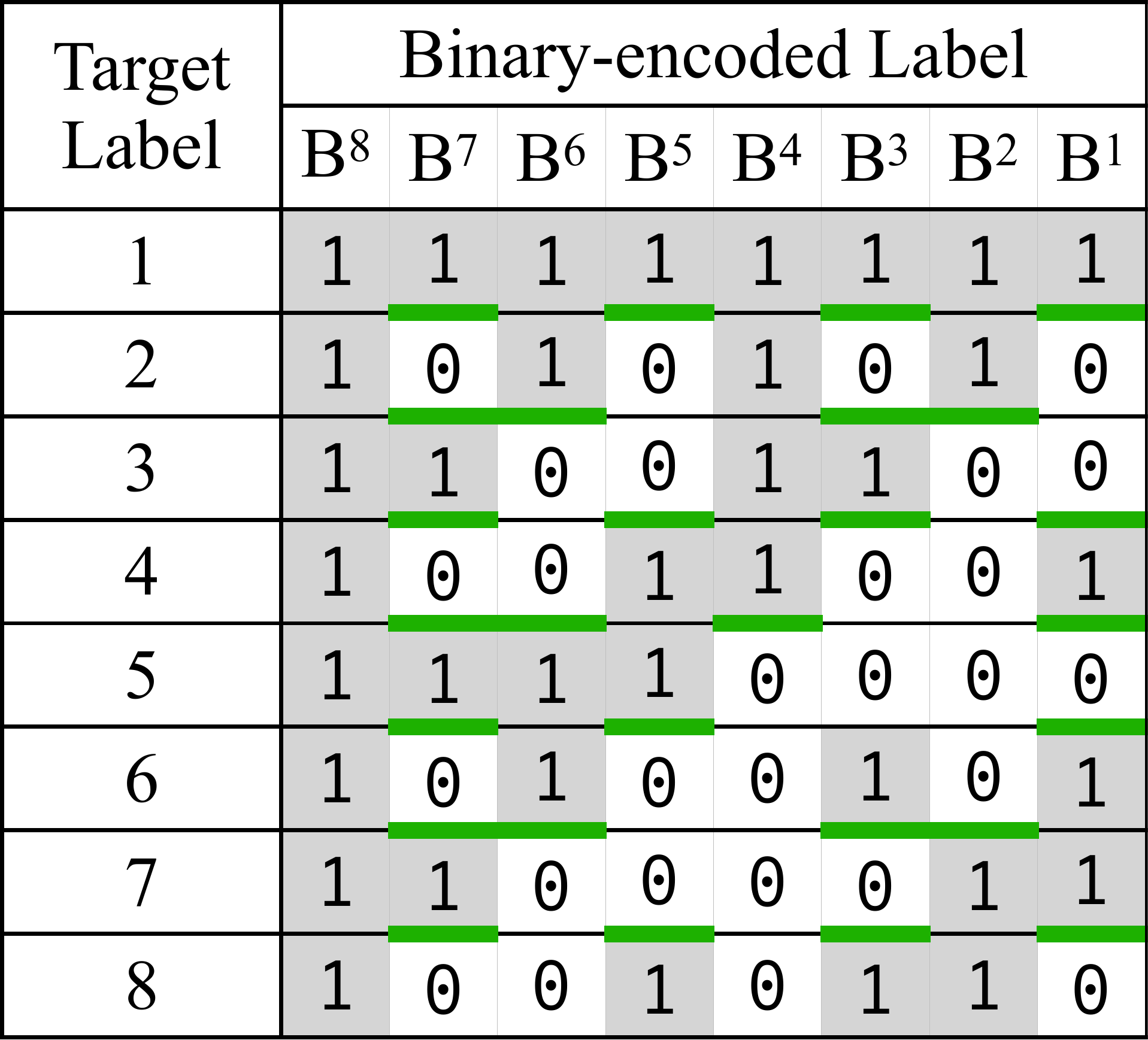}
   \caption{Hadamard encoding}
   \label{fig:ch}
  \end{subfigure}
  \hfill
  \begin{subfigure}[t]{0.20 \linewidth}
    \centering
  \includegraphics[height=\imageheight]{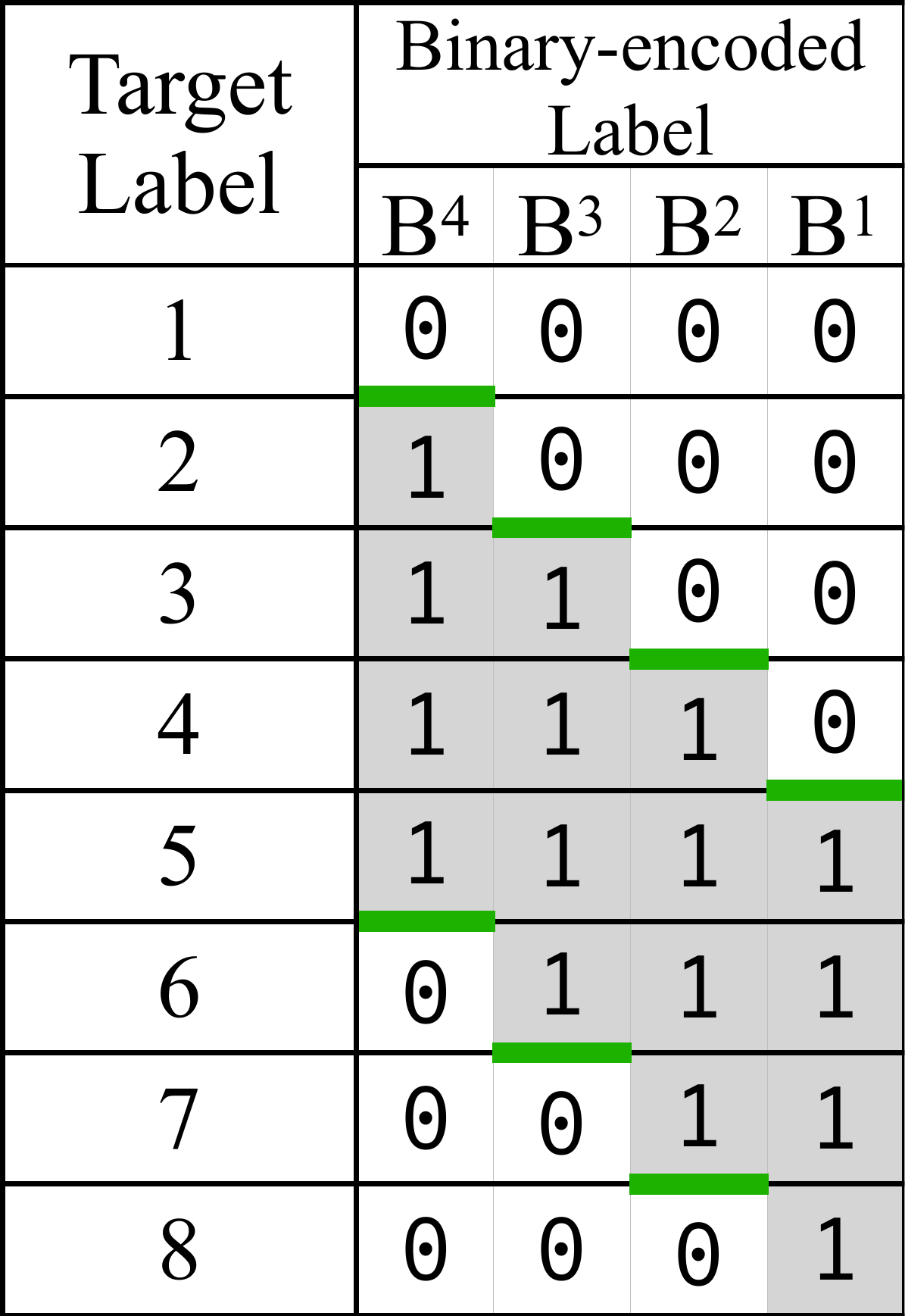}
 \caption{Johnson encoding}
 \label{fig:cj}
\end{subfigure}
\hfill
\begin{subfigure}[t]{0.20 \linewidth}
    \centering
  \includegraphics[height=\imageheight]{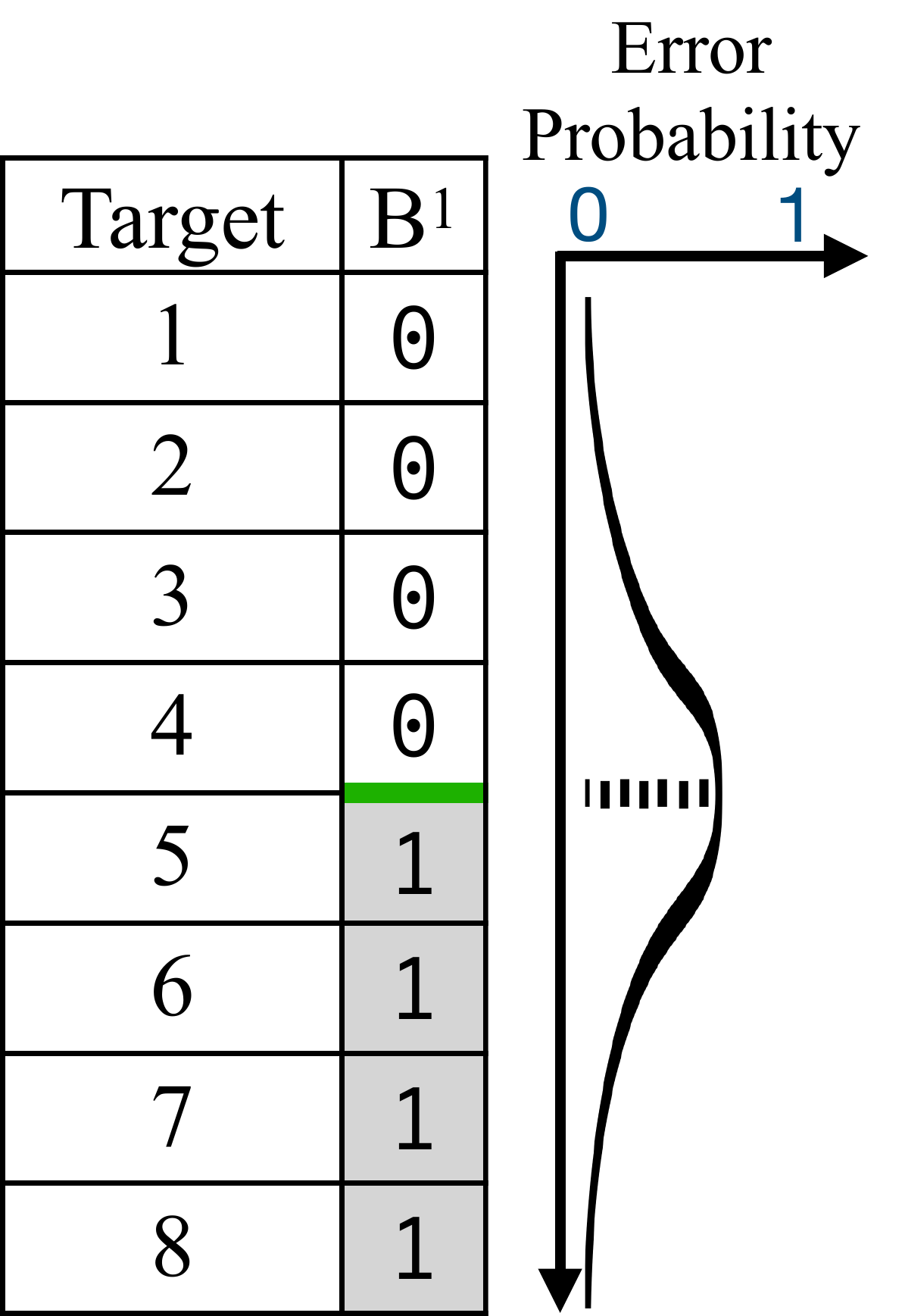}
 \caption{Error distribution}
 \label{fig:cdist}
\end{subfigure}
  \hfill
  \caption{(a-c) Examples of label encodings. Each row represents the binary-encoded target label for a given target. (d) represents the error probability distribution of the classifier-$1$ for different target values in Johnson encoding. Green lines represent the bit transitions of a classifier.}
 \label{fig:encoding}
\end{figure} 
}
Using encoded labels introduces error-correction capability, i.e., tolerance to classification error. %, depending upon the design of codes. 
%, where the label can be correctly decoded from predicted code with errors in a few bits. 
Hamming distance between two codes (number of differing bits) gives a measure of error-correction capability. 
Error-correcting codes, such as Hadamard codes (Figure~\ref{fig:ch}), have been proposed to encode labels in multiclass classification~\citep{ecoc,Verma2019}. 
BEL showed that such codes are not suitable for regression due to differences in task objectives and classifiers' error probability distribution, and proposed properties of suitable codes for regression.  

The first property suggests a trade-off between classification errors and error correction properties. 
Each classifier learns a decision boundary for \emph{bit transitions} from $1 \to 0$ and $0 \to 1$ in the classifier's target bit over the numeric range of labels (green lines within a column in Figure~\ref{fig:encoding}). 
\blue{For example, in Johnson encoding~\citep{libaw} (Figure~\ref{fig:cj}), the classifier for bit $B^3$ learns two decision boundaries for  bit transitions in intervals $(2,3)$ and $(6,7)$. The number of intervals for which the classifier has to learn a separate decision boundary increases with bit transitions, increasing its complexity.}
%Error-correcting 
Hadamard codes have excellent error-correction properties but have several bit transitions (Figure~\ref{fig:ch}); this increases the complexity of a classifier's decision boundary and reduces its classification accuracy compared to unary and Johnson codes (Figure~\ref{fig:cu} and Figure~\ref{fig:cj}). 
\red{\cite{spectral} introduced the term \emph{spectral bias} and demonstrated that neural networks prioritize learning low-frequency functions (i.e., lower local fluctuations). The spectral bias of neural networks provides insights into accuracy improvement with the reduction in the number of bit transitions. }

Second, Hamming distance between two codes should increase with the difference between corresponding label values to reduce the probability of making large absolute errors between predicted and target labels. 
\blue{The probability that erroneous predicted code for label X will be decoded as Y decreases as the hamming distance between codes for values X and Y increases. Thus the above rule reduces the regression error.  }  
Last, the encoding design should also consider the error probability of classifiers. 
BEL shows that the error probability of classifiers is not uniform for regression and increases near bit transitions, as shown for classifier $\text{B}^1$ in Figure~\ref{fig:cdist}.  
Here, the probability of predicting $8$ for target label $1$ is very low, as the bit differing between corresponding codes ($\text{B}^1$) has a very low classification error probability. 
BEL shows that this property can be exploited to design better codes for regression.
These three factors significantly affect the suitability of encodings. % for regression. 
BEL demonstrates that simple codes sampled based on these properties, such as unary or Johnson code, result in lower errors than widely used error-correcting Hadamard code.
%\subsection{Label Encodings}
%Prior works have proposed use of binary classification for ordinal regression~\cite{ordext,agecnn,dorn}, regression~\cite{ShahICLR2022}, and multiclass classification~\cite{Verma2019,Song2021,ecoc,extremecode,multiclass} using binary-encoded labels. 
%Ordinal regression is a combination of regression and classification problems, where the output is a discrete class from a set of ordinal classes, e.g., the rating of a movie. 
%Typically, an ordinal regression problem with labels $\in \{1,2,...,M\}$ is solved by $M$ binary classifiers, where classifier-$k$ outputs whether the label is greater than $k$ or not. 
%Other works have proposed the use of error-correcting codes for multiclass classification. 
%Instead of using one-hot encoding to represent the target label ($1$ for the correct class and $0$ for others), each label is encoded to an $M$-bit binary code, and $M$ binary classifiers are used to learn target encodings.  
%In the prediction stage, outputs of $M$ binary classifiers are combined to find the predicted class. 
%More recently, Shah et al.~\cite{ShahICLR2022} proposed a generalized framework for regression by binary classification. In regression by binary classification, the target label $y_i$ is quantized ($Q_i$), and converted to a binary code $B_i \in \{0,1\}^M$ using an encoding function $\mathcal{E}$. These binary-encoded labels are used to train $M$ binary classifiers. Similarly, the output predicted code $\hat{B}_i$ is passed through a decoding function, which outputs a real-valued continuous/quantized prediction $\hat{y}_i$. 
\subsection{Encoding design} 
Encoding design is a well-studied problem with applications in several fields. 
Iterative approaches, such as simulated annealing or random walk, have been proposed for code design~\citep{ecoc,Song2021}. 
%However, applying simulated annealing required an efficient way of approximating the error achieved by sampled encodings at each iteration, as training the network each iteration to measure error for sample encodings is computationally expensive and slow. 
However, iterative approaches are computationally expensive as each iteration requires full/partial training of the network to measure the error for sample encodings. 
Works on multiclass classification using binary classifiers have demonstrated the effectiveness of error-correcting codes such as Hadamard or random codes~\citep{Verma2019,ecoc}. \citet{extremecode} proposed an autoencoder-based approach to design compact codes for multiclass classification problems with a large number of classes. 
%%They demonstrate that the relational information between different classes can be used to design suitable encodings where labels with similar classes have similar encodings. 
However, these approaches do not consider the task objective and classifiers' nonuniform error probability distribution for regression. 
Deep hashing approaches aim to find binary hashing codes for given inputs such that the hashing codes preserve the similarities in the inputs space~\citep{deephashing,hashing,cnnh,ordhashing,sdh}. 
Deep supervised hashing approaches use the label information to design the loss function.  
%%~\citet{sdh} proposed an end-to-end approach for hashing and used a loss function to decrease the hamming distance between binary codes for images with the same label. 
In deep hashing, loss functions are designed to decrease the hamming distance between binary codes for similar images (e.g., same label). In contrast, label encoding design for regression aims to reduce the error between decoded output codes and target labels. 
Further, deep hashing approaches are designed for classification datasets and do not account for the nonuniform error probability distribution of classifiers observed in regression. 
As shown in prior work~\citep{ShahICLR2022}, classifiers' nonuniform probability significantly affects the design of suitable codes for regression. 
Thus, a naive adaptation of deep hashing approaches for regression problems performs poorly compared to codes designed by the proposed approach~\CLL{ }(Section~\ref{app:hash}). 
\section{Regularized Label Encoding Learning}
\label{sec:proposed} 
Regression aims to minimize the error between target labels $y_i$ and predictions $\hat{y}_i$ for a set of training samples $i$. 
In regression by binary classification, the network learns $M$-bit binary-encoded labels $B_i \in \{0,1\}^M$. 
During inference, the predicted code $\hat{B}_i$ is decoded to a real-valued label $\hat{y}_i$. 
We propose to relax label encodings' search space from a discrete ($\{0,1\}^M$) to a continuous space ($\mathbb{R}^M$), enabling the use of traditional continuous optimization methods. 
We propose regularizers to enable efficient search through this space. 
This work automates the search for label encoding using an end-to-end training approach that learns the network parameters and label encoding together.

This section explains the proposed label encoding learning approach \CLL. 
First, we explain the regression by binary classification formulation used in this work for end-to-end training of network parameters and label encoding. 
%Further, we explain the properties of suitable label encodings in the discrete space. 
Further, we introduce properties of suitable label encodings in continuous space. 
Lastly, we explain the proposed regularizers and loss function that accelerate the search for label encodings by encouraging learned label encoding to exhibit the proposed properties.  

\subsection{Label Encoding Learning} 
\begin{figure}[t]
    \centering
        \includegraphics[width=0.81\textwidth]{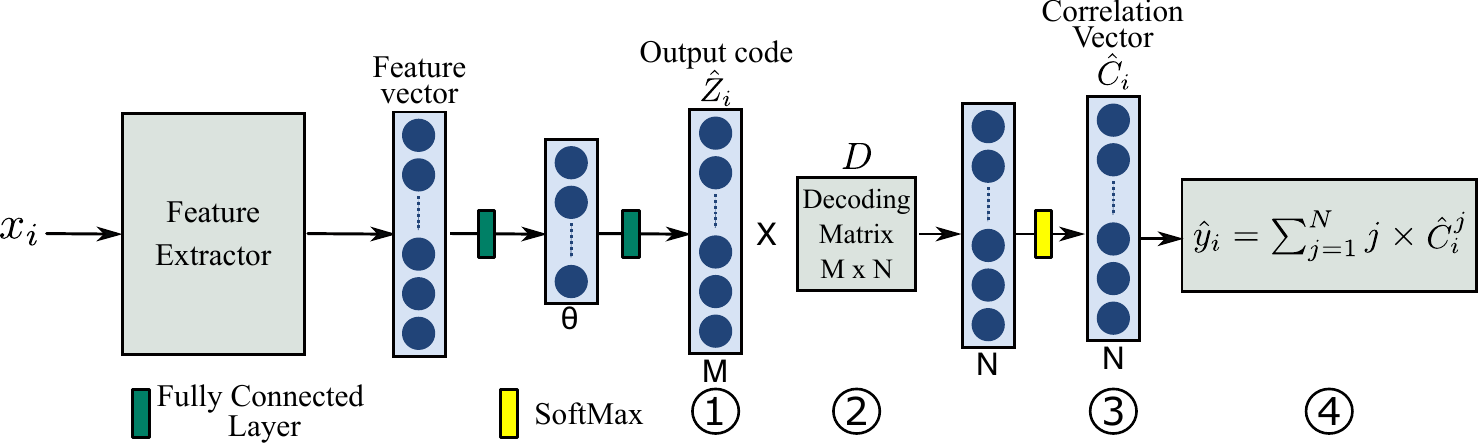}
   \caption{The flow for combined training of feature extractor and label encoding. }
   \label{fig:setup}
  \end{figure} 
%%The design space of encoding and decoding functions used for conversion between real-valued and binary-encoded labels is vast.  
\begin{table}[]
    \scriptsize
    \centering
    \caption{Summary of notations used in this work}
    \label{tab:notation}
    \begin{tabular}{ll}
    \toprule
    Notation & Description \\ \midrule
    $x_i$, $y_i$, $Q_i$     &  Input, real-valued target label, and quantized target label  for training example $i$. $y_i \in [1,N]$ and $Q_i \in \{1,2,...,N\}$ \\\hline     
    $N$                 & The range of target labels $y_i$; Number of quantization levels for $Q_i$ \\\hline     
    $M$                 & Number of bits/values for label encoding  \\\hline     
    $B_i$, $\hat{B}_i$  &Target and predicted binary-encoded labels (used for hand-crafted label encoding\\\hline     
   % $\mathcal{E}$       & Encoding function (used for hand-crafted label encoding)        \\\hline     
    $\hat{Z}_i$         & Predicted real-valued encodings; activation values of the output code layer in Figure~\ref{fig:setup} \\\hline     
    $E$                 & Learned label encoding through RLEL; calculated from $\hat{Z}_i$ for all training examples using~\eqref{eq:ez}                  \\\hline     
    $\hat{C}_i$         & Output correlation vector of length $N$. Here $\hat{C}^j_i$ gives a measure of the probability that predicted label value is equal to $j$ \\\hline     
    $D$                 & Decoding matrix that converts the predicted encodings to a correlation vector $\hat{C}_i$ \\ \bottomrule                           
    \end{tabular}
    \end{table}
\paragraph{Preliminaries: }Figure~\ref{fig:setup} represents the formulation used in this work for label encoding learning.  
$x_i$ and $y_i$ represent the input and the real-valued target label for sample $i$, respectively. 
We assume $y_i \in [1, N]$ for simplicity as the real-valued targets with any arbitrary numeric range can be scaled and shifted to this range. 
$Q_i \in \{1,2,...,N\}$ represents the quantized target label. 
The input $x_i$ is passed through a feature extractor and fully connected (FC) layers to generate the predicted encoding $\hat{Z}_i \in \mathbb{R}^M$~{\textcircled{\small 1}}. 
Here, an FC layer of size $\theta$ ($\theta < M $) is added between the feature vector and output code.
This layer reduces the number of parameters in FC layers and improves accuracy, as shown by previous work~\citep{ShahICLR2022}. 
Each neuron of the output code is a binary classifier, and the magnitude $\hat{Z}_i^k$  gives a measure of the confidence of the classifier-$k$~\citep{multiclass}.  
The output code and a decoding matrix $D \in \mathbb{R}^{M\times N} $ are multiplied~{\textcircled{\small 2}}, and the output is passed through a softmax function to give a correlation vector $\hat{C}_i \in \mathbb{R}^N$~{\textcircled{\small 3}}, where the value of $\hat{C}_i^k$ represents the probability that the predicted label $\hat{y}_i=k$. 
This correlation vector is then converted to a real-valued prediction by taking the expected value~{\textcircled{\small 4}}. \blue{Table~\ref{tab:notation} summarizes the notations used in this work. }

Prior works use custom-designed label encoding in this formulation. For example, \citet{ShahICLR2022} proposed a series of suitable label encodings $B_i=\mathcal{E}(Q_i) \in \{0,1\}^M$. The network can be trained using binary cross-entropy loss between $B_i$ and $\hat{Z}_i$, and these encodings are used as columns of the decoding matrix ($D_{:,i}=\mathcal{E}(i)$). %In this case, $D \in \{0,1\}^{M \times N}$ is binary.  
However, it is desirable to automatically find suitable encodings $B_i$ and decoding matrix $D$ without searching through a set of hand-designed encodings.  

The search space of binary label encodings is discrete and hence challenging to search using traditional continuous optimization methods~\citep{darabi2018regularized}. 
Hence, we relax the assumption of binarized label encodings and use a continuous search space. 
This relaxation, coupled with the proposed formulation, enables the use of traditional optimizers to learn label encoding and the decoding matrix $D$ with the entire network by optimizing the loss between targets and prediction.  
Let $\mathbb{S}_n$ represent the set of training samples with quantized target $Q_i = n$, and $E \in \mathbb{R}^{N\times M}$ represent a label encoding matrix, where each row $E_{n,:}$ is the encoding for target $Q_i = n$. $E$ is defined as:
\begin{equation}
    \label{eq:ez}
    E_{n,:} = \frac{1}{|\mathbb{S}_n|} \sum_{i \subset \mathbb{S}_n} \hat{Z}_i
\end{equation}
%from network output as $\mathcal{E}(n) = \frac{1}{|I_n|} \sum_{i \subset I_n} \hat{Z}_i $. 

However, training the network solely with the loss between  $\hat{y}_i$ and $y_i$ does not constrain the search space of label encodings ($E$). 
In regression, the label encoding ($E$) significantly impacts the accuracy, and label encodings that follow specific properties result in lower error~\citep{ShahICLR2022}. 
The following section explains desirable characteristics of output codes for regression and how these properties can be encouraged in learned label encoding using regularization functions. 

\subsection{Label Encoding Learning with Regularizers }
\label{subsec:const} 
%The search space of label encodings is vast; however, certain label encodings have been shown to be more suitable for regression tasks. 
Section~\ref{subsec:prop} summarizes the properties of suitable binary label encodings for regression proposed by prior works. 
These properties constrain the vast search space of label encodings. 
%% and encourage learning of better label encodings. 
We further propose two properties applicable to real-valued label encodings to narrow its search space. 

\textbf{R1 - Distance between encodings: } 
%For a regression problem, the difference between target values gives a measure of similarity between corresponding desired label encodings. 
A binary classifier's real-valued output represents its confidence (i.e., error probability). 
The L1 distance between real-valued predicted encodings gives more weight to classifiers that are more confident (i.e., higher value). 
{Thus, by considering the L1 distance between real-valued codes instead of the hamming distance between binary codes, we can combine the second and third design properties of binary label encodings (Section~\ref{subsec:prop}) into a single rule for real-valued label encodings. This gives the first regularization rule: 
\emph{ L1 distance between encodings for two labels should increase with the difference between two labels}}, i.e., $||{E}_{i,:} - {E}_{j,:}||_1 \propto |i-j|$. 
 
\textbf{{R2 - Regularizing bit transitions:}}
The number of bit transitions in a bit-position of label encoding gives a measure of the binary classifier's decision boundary's complexity. 
There are no $0 \to 1$ or $1 \to 0$ transitions in real-valued label encodings. 
Thus, we approximate the number of bit transitions by measuring the L1 distance between encodings for adjacent label values $Q_i =n \text{ and } Q_i = n+1$. 
The number of bit transitions for real-valued label encoding ${E}$ can be approximated as:
\begin{equation}
    \label{eq:bit}
    \sum_{i=1}^{M} \sum_{j=1}^{N-1} |{E}_{j,i} - E_{j+1,i}|
\end{equation}

This leads to the second regularization rule: 
{\emph{The L1 distance between encodings for adjacent target label values should be regularized to find a balance between the complexity of the decision boundary and the error-correction capability of designed codes for a given benchmark. }
\subsection{Loss Function Formulation}
\label{subsec:loss} 
We propose two regularizers applicable to learned label encoding ($E$) to limit its search space. 
$E$ is measured from the output codes $\hat{Z}_i$ over the complete training dataset (\Eqref{eq:ez}). 
However, deep neural networks are trained using mini-batches, where each batch consists of $K$ training examples sampled randomly from the (typically shuffled) training set. 
We extend the proposed regularization rules to apply to a minibatch-based loss function. 

\textbf{R1: } 
{Regularizer R1 can be approximated as the following for a batch with $K$ training examples:}
\begin{equation}
    \label{eq:r1}
    \mathcal{L}_1 = \sum_{i=1}^{K}\sum_{j=1}^{K} \text{max}(0, 2 \times |y_i-y_j|- ||\hat{Z}_i - \hat{Z}_j||_1 )
\end{equation}
{The above regularization considers $K^2$ pairs in a minibatch of $K$ samples, and penalizes a pair of training samples $i$ and $j$ if L1 distance between encodings $\hat{Z}_i$ and $\hat{Z_j}$ is less than twice the difference between corresponding label values. 
\blue{The scaling parameter is set to two as it encourages at least one bit difference between two binary codes. }
This encourages the L1 distance between encodings to be approximately proportional to the difference between corresponding target values. }

\textbf{R2: }
Regularizer R2 minimizes the L1 distance between encodings of adjacent label values. 
In a randomly formed minibatch consisting of only a subset of training examples, adjacent target labels might not be present. 
Hence it is nontrivial to apply this regularizer to the label encoding. 
However, we find that imposing this regularizer on the decoding matrix also helps with regularizing the bit transitions in the learned label encoding and can be added to the loss function irrespective of the batch formulation approach. \blue{Theorerical analysis and empirical verification for this approach are provided in Section~\ref{sec:r2proof} and Section~\ref{sec:abl}.}
{\Eqref{eq:r2} represents the proposed regularizer. } 
%%proposed based on empirical verification to reduce the bit transitions in learned label encoding. }
\begin{equation}
    \label{eq:r2}
    \mathcal{L}_2 = \sum_{i=1}^{M}\sum_{j=1}^{N-1} |D_{i,j}-D_{i,j+1} |
\end{equation}

\paragraph{Loss Function: }
We use the cross-entropy loss between $\hat{C}_i$ and soft target labels. 
\blue{Here, each bit of label encoding resembles a binary classifier. However, identifying the predicted label corresponding to the multi-bit label encoding can be treated as a multiclass classification problem. }
Soft target labels are probability distributions generated using the distance between different classes. 
Soft target labels can be used with cross-entropy loss and have shown improvement over typical classification loss between the correlation vector $\hat{C}_i$ and quantized target label $Q_i$ or regression loss between the expected prediction $\hat{y}_i$ and target label $y_i$ for ordinal regression~\citep{softlabel}. We use this loss function for \CLL{ }and multiclass classification. 
Complete loss function with regularizers (Equation~\ref{eq:r1} and~\ref{eq:r2}) can be written as:
\begin{multline}
    \label{eq:loss}
    \mathcal{L} = \sum_{i=1}^{K} \text{CE}(\hat{C}_i, \phi(y_i))+ \alpha \sum_{i=1}^{M}\sum_{j=1}^{N-1} |D_{i,j}-D_{i,j+1} | + \beta \sum_{i=1}^{K}\sum_{j=1}^{K} \text{max}(0, 2 \times |y_i-y_j|- ||\hat{Z}_i - \hat{Z}_j||_1 ), \\ \text{where } \phi^j(y_i) = \frac{e^{-|j-y_i|}}{\sum_{n=1}^{N} e^{-|n-y_i|}}
\end{multline}

Here, the first term is the loss between target and predicted labels. 
$\phi_i$ represents the target probability distribution generated from target $y_i$.  
The second and third terms are for regularizer R1 (\Eqref{eq:r1}) and regularizer R2 (\Eqref{eq:r2}), respectively.
%so that the L1 distance between label encodings increases with the difference between corresponding label values. 
%The third term is for regularizer R2 (\Eqref{eq:r2}) to regularize the bit transitions in learned label encodings. 

A trade-off exists between the proposed desirable properties of label encodings: Encouraging one design property comes at the cost of relaxing constraints imposed by other design properties. As demonstrated by \cite{ShahICLR2022}, finding the right balance between these properties for a given benchmark is crucial to finding the best label encoding for a given problem. 
%%A trade-off exists between proposed design properties of label encodings, and encouraging one design property comes at the cost of relaxing constraints posed by other design properties. As demonstrated by prior works, finding the right balance between these properties for a given benchmark is crucial to finding the best label encodings for a given problem~\citep{ShahICLR2022}. 
%There exists a trade-off between proposed design properties of label encodings, and finding the right balance between these properties for a given benchmark is crucial to finding the best label encodings for a given problem, as demonstrated by prior works~\citep{ShahICLR2022}.  
Thus, these design properties can be naturally applied as regularizers, and the search for balance between different properties can be seen as tuning the regularization parameters $\alpha$ and $\beta$. 
%In the proposed setup, classification loss between the correlation vector $\hat{C}_i$ and quantized target label $Q(y_i)$, or regression loss between the expected prediction $\hat{y}_i$ and target label $y_i$ can be used for training. 
% \cite{shah2022} proposed several encodings suitable for regression and proposed characteristics of suitable codes for regression based on empirical and analytical studies. However, it is desirable to search these encodings instead of carrying multiple training runs. 
%\input{sections/methodology.tex}
\section{Evaluation}
\label{sec:eval}
This section first provides the experimental setup used to evaluate the proposed approach, then we compare \CLL{ }with different label encoding design methods. We also compare \CLL{ }with different regression approaches to demonstrate its effectiveness as a generic regression approach. Last, we provide an ablation study to show the impact of proposed regularizers. 
\subsection{Experimental Setup}
\label{subsec:exp}
%% The table for benchmarks
\begin{table}[t]
    \centering
    \setlength\tabcolsep{4pt}
    \caption{Benchmarks used for evaluation} 
    \label{tab:benchmarks}
    \scriptsize
    \begin{tabular}{C{1.6cm}C{1.2cm}C{3.0cm}cC{2.0cm}C{0.3cm}}
      \toprule
     Task  & Feature Extractor  &  Dataset & Benchmark & Label range/ Quantization levels & $\theta$ \\ \midrule 
     \multirow{2}{\linewidth}{\centering Landmark-free 2D head pose estimation}  & \multirow{2}{\linewidth}{\centering ResNet50 \citep{resnet}}  & 300LP~\citep{300wlp}/AFLW2000~\citep{300wlp} & LFH1 & 0-200/200 & 10 \\ \cline{3-6}
     &    & BIWI~\citep{biwi} & LFH2  & 0-150/150 & 10\\  \cline{1-6}
     \multirow{3}{\linewidth}{\centering Facial Landmark Detection}  & \multirow{3}{\linewidth}{\centering HRNetV2-W18~\citep{hrnetface}}  & COFW~\citep{cofw} & FLD1/FLD1\_s (100\%/10\% training dataset) & 0-256/256 & 10 \\ \cline{3-6}
     &   & 300W~\citep{300w} & FLD2/FLD2\_s (100\%/10\% training dataset) & 0-256/256 & 10  \\  \cline{3-6}
     &   & WFLW~\citep{lab} & FLD3/FLD3\_s (100\%/10\% training dataset)  & 0-256/256 & 10\\ \cline{1-6} 
     \multirow{2}{\linewidth}{\centering Age estimation}  & \multirow{2}{\linewidth}{\centering ResNet50/ ResNet34} & MORPH-II~\citep{morphii} & AE1 & 0-64/64 & 10 \\ \cline{3-6}
     &  &   AFAD~\citep{agecnn} & AE2 & 0-32/32 & 10 \\  \cline{1-6}
    End-to-end autonomous driving & PilotNet\citep{pilotnet} & PilotNet & PN & 0-670/670 & 10 \\ 
     %\cline{1-7}
    %\multirow{2}{\linewidth}{\centering Age estimation}  & \multirow{2}{\linewidth}{\centering ResNet50 /ResNet34~\cite{resnet}} & \multirow{2}{\linewidth}{\centering Ordinal regression\\~\cite{coralcnn}} & MORPH-II~\cite{morphii} & AE1 & 0-64/64 & 10 \\ \cline{4-7}
    % &  &   & AFAD~\cite{agecnn} & AE2 & 0-32/32 & 10 \\  %\cline{1-7}
    %End-to-end autonomous driving & PilotNet & Direct regression~\cite{pilotnet} & PilotNet & PN & 0-670/670 & 10 \\ 
    \bottomrule
    %\cmidrule(r){1-4}
    \end{tabular}
    \vspace{-3mm}
    \end{table}
Table~\ref{tab:benchmarks} summarizes the regression tasks, feature extractors architecture (Figure~\ref{fig:setup}), and datasets for benchmarks used for evaluation. 
Selected benchmarks cover different tasks, datasets, and network architectures and have been used by prior works on regression due to the complexity of the task~\citep{softlabel,ShahICLR2022}.   
We also evaluated \CLL{ }on facial landmark detection tasks with smaller datasets to demonstrate its generalization capability. 
In this setup, a subset of training samples is used for training, whereas the complete test dataset is used to measure the test error. 

Landmark-free 2D head pose estimation (LFH) takes a 2D image as input and directly finds the pose of a human head with three angles: yaw, pitch, and roll. 
The facial landmark detection task focuses on finding $(x,y)$ coordinates of key points in a face image. 
The age estimation task is used to find a person's age from the given face image. 
In end-to-end autonomous driving, the car's steering wheel's angle is to be predicted for a given image of the road.
Normalized Mean Error (NME) or Mean Absolute Error (MAE) with respect to raw real-valued labels are used as the evaluation metrics. % for FLDx and the rest of the benchmarks, respectively.  

%%For comparison with other encoding design approaches, we also evaluate simulated annealing, autoencoder-based approaches (summarized in Appendix~\ref{sec:a2}), and manually designed codes~\citep{ShahICLR2022}. 
We compare with other encoding design approaches, including simulated annealing, autoencoder (summarized in Appendix~\ref{sec:a2}), and manually designed codes~\citep{ShahICLR2022}.  
%The autoencoder-based approach has been proposed for multiclass classification with a large number of classes~\citep{extremecode}. 
%The methodology used for simulated annealing and autoencoder-based encoding design is explained in Appendix~A2. \citet{ShahICLR2022} provided five custom-designed codes and showed that different encodings are suitable for different benchmarks. We also evaluate these encodings to compare automatically designed label encodings with hand-designed label encodings.   
We also compare \CLL{ }with generic regression approaches, such as direct regression and multiclass classification. 
For direct regression, L1 or L2 loss functions with L2 regularization are used. Label value scaling (hyperparameter) is used to change the numeric range of labels. 
%Hyperparameter tuning is used to determine the loss function, regularization, and scaling factor. 
For multiclass classification, we use cross-entropy loss between the softmax output and target labels. 

The feature extractor and regressor are trained end-to-end for all approaches.
The feature extractor architecture, data augmentation, and the number of training iterations are kept uniform across different approaches for a given benchmark. 
\blue{There is no notable difference between the training time for all approaches. } 
%%For hyperparameter tuning in all approaches (including $\alpha$ and $\beta$ values for \CLL), 
The training dataset is divided into $70\%$ training and $30\%$ validation sets for tuning hyperparameters. 
The network is trained using the full dataset after hyperparameter tuning. 
\blue{We use the same values for quantization levels as prior work~\citep{ShahICLR2022}. }
An average of five training runs with an error margin of $95\%$ confidence interval is reported. 
Appendix~\ref{sec:a3} provides details on datasets, training parameters, related work (task-specific approaches), and evaluation metrics. 
%\subsection{Label Encoding Learning}
%\label{sec:eval}
\subsection{Comparison of \CLL{ }with Encoding Design Approaches}
\label{sec:eval1}
Table~\ref{tab:compare_code} compares  different encoding design approaches. 
\CLL{ }results in lower error than simulated annealing and autoencoder-based approaches for most benchmarks. 
Both approaches are widely used for code design. 
However, for regression tasks, the suitability of label encoding depends upon the problem, including the task, network architecture, and dataset~\citep{ShahICLR2022}. 
Simulated annealing or autoencoder-based approaches do not optimize the encodings end-to-end with the regression problem., resulting in higher error. 
Furthermore, the gap between the error of learned label encoding with and without regularizers (RLEL and LEL) increases for smaller datasets, which suggests that \CLL{-}learned codes generalize better.  

%%\CLL{ }results in the lowest error for most benchmarks. 
{\CLL{ }can not be used with binary-cross entropy loss for training. We observe that for some benchmarks, the autoencoder-based approach outperforms (e.g., LFH1) as it can be used with binary-cross entropy loss. }
{The main objective of \CLL{ }is to automatically learn label encoding that can reach the accuracies of manually designed codes (\BEL), as using such codes is time and resource-consuming.
\blue{Hyperparameter search for \CLL{ } can be performed by off-the-shelf hypermeter tuners/libraries without manual efforts~\citep{hyperband,Falkner2018BOHBRA}. 
In contrast, hand-designed codes need human intervention to design codes. Also, multiple training runs are still required to find suitable codes for a given benchmark from a set of hand-designed codes.
}
% and \CLL{ }results in similar or marginally improved error over BEL for most benchmarks.  }
As shown in Table~\ref{tab:compare_code}, \emph{\CLL{ }results in lower or comparable errors to hand-designed codes. }
\begin{table}[t]
    \centering
    \setlength\tabcolsep{3pt}
    \caption{Comparison of \CLL with different label encoding design approaches. The bold and underlined numbers represent the first and second best errors, respectively.  } 
    \label{tab:compare_code}
    \footnotesize 
    \begin{tabular}{L{3.3cm}C{\y cm}C{\y cm}C{\y cm}C{\y cm}C{\y cm}C{\y cm}}
    \toprule
    & \multicolumn{6}{c}{ Error (MAE or NME) }   \\ \hline
    Approach  & LFH1 & LFH2 & FLD1 & FLD1\_s  & FLD2 & FLD2\_s         \\ \midrule
    Simulated annealing   & 4.32{$\pm$0.12}   &  5.03{$\pm$0.08} & 3.55{$\pm$0.01}  & 6.52{$\pm$0.05} & 3.59{$\pm$0.00} & 5.35{$\pm$0.01}  \\ \hline
    Autoencoder  & \textbf{3.38}{$\pm$0.01}   &  4.84{$\pm$0.02} &  3.39{$\pm$0.01} & 4.85{$\pm$0.03} & \underline{3.39}{$\pm$0.00}   &   \underline{4.20}{$\pm$0.05}  \\ \hline
    \LEL (w/o regularizers)               & 4.03{$\pm$0.15}  &  4.96{$\pm$0.08}&   \underline{3.36}{$\pm$0.01} &  4.98{$\pm$0.07} & \underline{3.39}{$\pm$0.01}      & 4.28{$\pm$0.05}  \\ \hline 
    \BEL (Manually designed)&  3.56{$\pm$0.11} & \textbf{4.77}{$\pm$0.05}  & \textbf{3.34}{$\pm$0.01} &  \textbf{4.63}{$\pm$0.03}   & 3.40{$\pm$0.02} &\textbf{4.15}{$\pm$0.01} \\ \hline
    \CLL                 & \underline{3.55}{$\pm$0.10}  &  \textbf{4.77}{$\pm$0.05}&   \underline{3.36}{$\pm$0.01} &  \underline{4.71}{$\pm$0.04} & \textbf{3.37}{$\pm$0.02}       & \textbf{4.15}{$\pm$0.05} \\ \bottomrule
\end{tabular}
\begin{tabular}{L{3.3cm}C{\y cm}C{\y cm}C{\y cm}C{\y cm}C{\y cm}C{\y cm}}
    \toprule
    Approach   & FLD3 & FLD3\_s & AE1 & AE2 & PN   &  \\ \midrule
    Simulated annealing      & 4.52{$\pm$0.02}  & 6.38{$\pm$0.01}  & 2.33{$\pm$0.01 }   &  3.17{$\pm$0.01} &  4.25{$\pm$0.01 }  &  \\ \hline
    Autoencoder        &  4.36{$\pm$0.01} & \underline{5.62}{$\pm$0.01} &  2.29{$\pm$0.00 }   &   3.19{$\pm$0.01 } &   {4.49$\pm$0.04 }  &  \\ \hline
    \LEL (w/o regularizers)              &  \textbf{4.35}{$\pm$0.02}&   5.68{$\pm$0.04} & 2.30{$\pm$0.01}  &  3.17{$\pm$0.01}&   3.22{$\pm$0.02}  & \\ \hline 
    \BEL (Manually designed)             &  4.36{$\pm$0.02}& \underline{5.62}{$\pm$0.00}  &  \textbf{2.27}{$\pm$0.01} & \textbf{3.11}{$\pm$0.00}  & \underline{3.11}{$\pm$0.01}&  \\ \hline
    \CLL             &  \textbf{4.35}{$\pm$0.01}&   \textbf{5.58}{$\pm$0.01} & \underline{2.28}{$\pm$0.01}  &  \underline{3.14}{$\pm$0.01}&  \textbf{3.01}{$\pm$0.03} & \\ \bottomrule
\end{tabular}
  \end{table}
\subsection{Comparison of \CLL{ }with Regression Approaches } 
\label{sec:eval2}
% Table for the main results
\begin{table}[t]
  \centering
  \setlength\tabcolsep{4pt}
  \caption{{Comparison of \CLL with different regression approaches and state-of-the-art task-specialized approaches (more details in Appendix~\ref{sec:a3}). ``/$x$M'' represents the model size.}  } 
  \label{tab:compare_regression}
  \footnotesize 
  \begin{tabular}{L{1.2cm}C{2.5cm}C{2.5cm}C{2.5cm}L{3.5cm}}
  \toprule
   &  {Direct regression} & Multiclass classification & \CLL  & \multicolumn{1}{c}{Task-specialized approach*}\\ \hline
  LFH1 &   { }4.22{$\pm$0.13}/23.5M & 4.49{$\pm$0.24}/24.2M & 3.55{$\pm$0.10}/23.6M & 3.30{$\pm$0.04}/69.8M \\ \hline
  LFH2 &   { }5.32{$\pm$0.12}/23.5M & 5.31{$\pm$0.05}/24.8M & 4.77{$\pm$0.05}/23.6M & 3.90{$\pm$0.03}/69.8M \\ \hline
  FLD1 &  { }3.60{$\pm$0.02}/10.2M & 3.48{$\pm$0.03}/25.6M & 3.36{$\pm$0.01}/10.6M & 3.34{$\pm$0.02}/10.6M \\ \hline
  FLD1\_s &  32.70{$\pm$1.37}/10.2M & 5.36{$\pm$0.03}/25.6M & 4.71{$\pm$0.04}/10.6M & - \\ \hline
  FLD2 &  { }3.54{$\pm$0.03}/10.2M & 3.46{$\pm$0.02}/45.2M & 3.37{$\pm$0.02}/11.2M & 3.07/25.1M \\ \hline
  FLD2\_s &  { }5.04{$\pm$0.02}/10.2M & 4.50{$\pm$0.04}/45.2M & 4.15{$\pm$0.05}/11.2M & -  \\ \hline
  FLD3 &   { }4.64{$\pm$0.03}/10.2M & 4.46{$\pm$0.01}/61.3M & 4.35{$\pm$0.01}/11.7M & 4.32/- \\ \hline
  FLD3\_s &  { }6.35{$\pm$0.07}/10.2M & 6.05{$\pm$0.01}/61.3M & 5.58{$\pm$0.01}/11.7M & - \\  \hline
  AE1 &   { }2.37{$\pm$0.01}/23.5M & 2.75{$\pm$0.03}/24.2M & 2.28{$\pm$0.01}/23.6M & 1.96/3.7M \\ \hline
AE2 &   { }3.16{$\pm$0.02}/23.5M & 3.38{$\pm$0.05}/24.8M & 3.14{$\pm$0.01}/23.6M & 3.47/21.3M \\ \hline
PN &   { }4.24{$\pm$0.45}/10.2M & 5.54{$\pm$0.03}/25.6M & 3.01{$\pm$0.03}/10.6M & 4.24/10.2M \\
 % AE1 &  2.44{$\pm$0.01}/23.1M & 2.75{$\pm$0.03}/23.1M &/23.1M \\ \hline
 % AE2 &  3.47/21.3M & 3.21{$\pm$0.02}/23.1M & 3.38{$\pm$0.05}/23.1M & /23.1M \\ 
 \bottomrule
\end{tabular}
\scriptsize{*This uses different network architecture, data augmentation, and training process. }
\end{table}
%% End of the results table
\CLL{ }is a generic regression approach that focuses on regression by binary classification and proposes a label encoding learning approach. 
We compare \CLL{ }with other generic regression approaches, including direct regression and multiclass classification % with soft labels~\citep{softlabel} 
as shown in Table~\ref{tab:compare_regression}. 
\CLL{ }problem formulation introduces more fully-connected layers after the feature extractor; hence, we also perform an ablation study on increasing the number of fully connected layers in Appendix~\ref{sec:a14}.  
{\emph{\CLL{ }consistently lowers the error compared to direct regression and multiclass classification with $10.9\%$ and $12.4\%$ improvement on average. }}
%BEL outperforms direct regression, \red{multiclass classification}, and application-specific regression approaches for the majority of the benchmarks. 
%\emph{\CLL{ }formulation results in lower error than regression approaches by imposing constraints on the output codes. }
  \subsection{Ablation Study}
  \label{sec:abl}
  Figure~\ref{fig:intror1} demonstrates that the use of regularizer R1 encourages the L1 distance between encodings to be proportional to the difference between target values. 
  The second regularizer R2 is introduced to regularize the number of bit transitions in encodings. 
  As mentioned in Section~\ref{subsec:loss}, we apply the regularization on the decoding matrix as it is nontrivial to apply this regularization on the output codes for randomly formed batches. 
  %We further observe the impact of the value of the $\alpha$ on bit transitions in the learned label encodings. 
  Table~\ref{tab:r2comp} summarizes the effect of $\alpha$ (i.e., the weight of R2) on the number of bit transitions in the decoding matrix and binarized/real-valued label encoding. 
  The second column is the number of bit transitions in the decoding matrix (\Eqref{eq:r2}), which is used as the regularization function. 
  The third and fourth columns are the total number of bit transitions in binarized and real-valued encodings (\Eqref{eq:bit}). 
  %The last column is an approximation of the total number of bit transitions measured using \Eqref{eq:bit} for real-valued encodings. 
  %The third column is the total number of bit transitions in binarized $E$. 
  %The last column is an approximation of the total number of bit transitions measured using \Eqref{eq:bit} for real-valued encodings. 
  %We measure learned label encodings $E$ for training samples using \Eqref{eq:ez}.
  The table shows that the proposed regularizer on the decoding matrix also encourages fewer bit transitions in the label encoding. 
  Figure~\ref{fig:intror2} shows the impact of regularizer R2 on learned binarized label encoding. 
  %  The second regularization term is introduced to regularize the number of bit transitions in label encodings. 
  
  As pointed out by prior works~\citep{ShahICLR2022}, there is a trade-off between the error probability and error correction capability of classifiers for regression. 
  Hence, depending upon the benchmarks, more bit transitions can be added as the advantage of increased error correction outweighs the increase in classification error. 
  We observe a similar trend, where adding R2 does not improve error for some benchmarks (FLD1\_s, FLD2\_s), as it constrains the number of bit transitions. 
  \begin{table}[t]
      \centering
      \setlength\tabcolsep{4pt}
      \caption{Effect of regularization R2 on bit-transitions in binarized and real-valued label encodings.} 
      \label{tab:r2comp}
      \scriptsize 
      \begin{tabular}{L{2cm}C{3.5cm}C{3.5cm}C{3.5cm}}
      \toprule
      $\alpha$ value & Proposed regularizer using Decoding matrix (\Eqref{eq:r2}) & \#Bit transitions in binarized label encoding & Approximated bit transitions in label encoding (\Eqref{eq:bit})  \\ \hline
      0 & 6816.1 & 5097  &391.88  \\ \hline 
      0.1 & 215.3  & 3596 & 168.52 \\ \hline 
      0.5 & 130.8  & 3180 & 104.19 \\  \bottomrule 
    \end{tabular}
    \end{table}
\iffalse
\textbf{Effect of hyperparameters in \CLL:}
    \iffalse
    \begin{wrapfigure}{R}{0.5\textwidth}
      \centering
      \includegraphics[width=0.49\textwidth]{figures/sweep_cofw_0p8.pdf}
      \caption{Effect of hyperparameters on NME for FLD1\_s benchmark}
      \label{fig:hypersweep}
      \end{wrapfigure}  
      \fi
    The \CLL{ }approach introduces two hyperparameters. 
    We first evaluate the sensitivity to these hyperparameters to determine the complexity of hyperparameter tuning. 
    Figure~\ref{fig:hypersweep} shows the NME for FLD1\_s benchmark for different values of $\alpha$ and $\beta$ values in \Eqref{eq:loss}. As shown in the figure, the error is not sensitive to small changes in these hyperparameters' values, suggesting that a sparse search in the hyperparameter space suffices. 
    Furthermore, several approaches have been proposed for efficient hyperparameter search~\citep{hyperband,Falkner2018BOHBRA}, and any off-the-shelf hypermeter tuners/libraries can be used to automatically find these values without manual efforts. 
    In contrast, hand-designed codes need human intervention to design codes. Also, multiple training runs are still required to find suitable codes for a given benchmark from a set of hand-designed codes. 
    On the other hand,  \CLL{ }provides an end-to-end automated approach for label encoding learning.
\fi
  %% various experiments to show the validity of design rules. By introducing negative the design rules in the . 
\section{Conclusion}
\label{sec:con}
This work proposes an end-to-end approach, Regularized Label Encoding Learning, to learn label encodings for regression by binary classification setup. 
We propose a combination of continuous approximation of binarized label encodings and regularization functions. This combination enables an efficient and automated search of suitable label encoding for a given benchmark using traditional continuous optimization approaches. 
The proposed regularization functions encourage label encoding learning with properties suitable for regression, and the learned label encodings generalize better, specifically for smaller datasets. 
Label encodings designed by the proposed approach outperform simulated annealing- and autoencoder-designed label encodings by 12.6\% and 2.1\%, respectively. 
\CLL{}-designed codes show lower or comparable errors to hand-designed codes. 
\CLL{ }reduces error on average by $10.9\%$ and $12.4\%$  over direct regression and multiclass classification. % with soft labels. 
%\CLL{ }also achieves new state-of-the-art accuracy for facial landmark detection (COFW). \\
%

\section{Acknowledgements}
%%We thank Dave Evans, Jonathan Lew, Saurabh Kumar, and the anonymous reviewers for their valuable comments on this work. 
%This research has been funded in part by the National Sciences and Engineering Research Council of Canada (NSERC) Strategic Project Grant number STP 506681-17. 
\red{This research has been funded in part by the National Sciences and Engineering Research Council of Canada (NSERC) through 
the NSERC strategic network on Computing Hardware for Emerging Intelligent Sensory Applications (COHESA) 
and through an NSERC Strategic Project Grant. 
Tor M. Aamodt serves as a consultant for Huawei Technologies Canada Co. Ltd and recently served as a consultant for Intel Corp. }

\textbf{Reproducibility:  } 
We have provided details on training hyperparameters, experimental setup, and network architectures in Appendix~\ref{sec:a3}. 
\red{Code is available at \url{https://github.com/ubc-aamodt-group/RLEL_regression}. We have provided the training and inference code with trained models. }
%%We have also submitted the training and inference code implementation with trained models in our supplemental material. 

\textbf{Code of Ethics: } 
Autonomous robotics and vehicles are major applications of deep regression networks. Thus improvement of regression tasks can accelerate the progress of these fields, which may lead to some negative societal impacts such as loss of jobs, privacy, and ethical concerns. 
%%Deval Shah is partly funded by the Four Year Doctoral Fellowship (4YF) provided by the University of British Columbia. 
\bibliography{ref}
\bibliographystyle{iclr2023_conference}

%%%%%%%%%%%%%%%%%%%%%%%%%%%%%%%%%%%%%%%%%%%%%%%%%%%%%%
\appendix
\section{Appendix}
This supplemental material provides additional results and ablation studies (Section~\ref{sec:a1}), methodology for baseline encodings design approaches (Section~\ref{sec:a2}), and related work on task-specialized approaches and experimental setup (Section~\ref{sec:a3}) for RLEL. 
Code is available at \url{https://github.com/ubc-aamodt-group/RLEL_regression}.

\subsection{Ablation Study}
\label{sec:a1}
Section~\ref{sec:a12}, Section~\ref{sec:a13}, and Section~\ref{sec:hype} provide an ablation study and supporting data on impact of proposed regularization functions and hyperparameters on label encoding learning.
Section~\ref{sec:a14} covers an ablation study on the impact of the number of fully connected layers in direct regression and multiclass classification.
Section~\ref{app:hash} explains and compares deep hashing approaches (adapted for regression) with~\CLL.
Section~\ref{sec:a11} provides results for geometric mean and Pearson coefficient as evaluation metrics. 
%For AE1 and AE2 benchmarks, the use of proposed regularizers results in slight improvement over label encodings learning without regularization (\LEL).
%We believe that this is due to the smaller label encodings ($62 \times 62$ for morph2, and $26 \times 26$ for AE2) compared to other benchmarks, and the proposed framework learns suitable label encodings without regularizers for a smaller search space of label encodings.

\subsubsection{Impact of Regularizer R1}
\label{sec:a12}
\begin{figure}[h]
 \centering
 \begin{subfigure}[t]{0.49\linewidth}
     \centering
     \includegraphics[width=\textwidth]{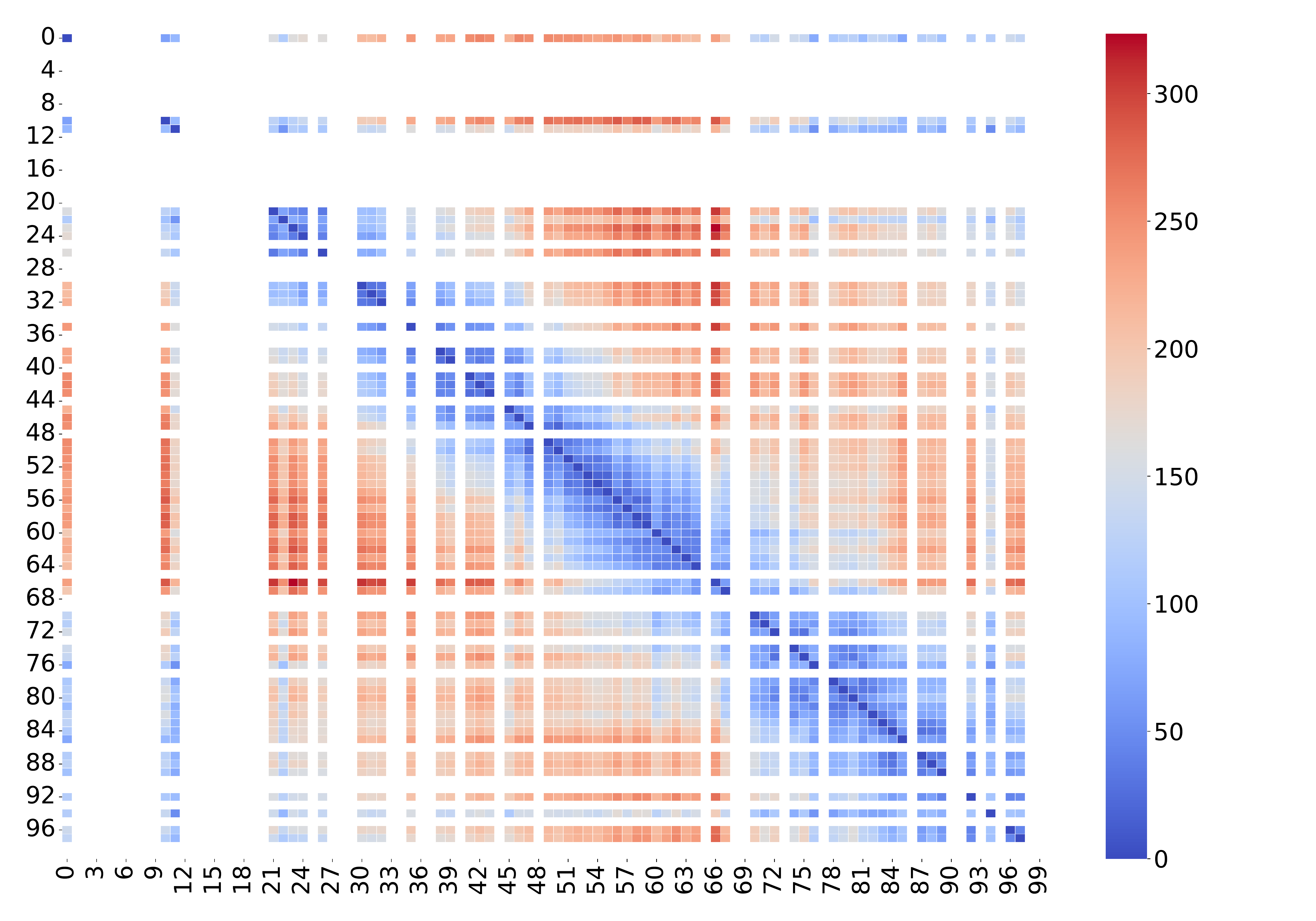}
     \caption{Without regularization: $\beta=0$}
     \label{fig:cofwr10}
 \end{subfigure}
 \begin{subfigure}[t]{0.49\textwidth}
   \centering
   \includegraphics[width=\textwidth]{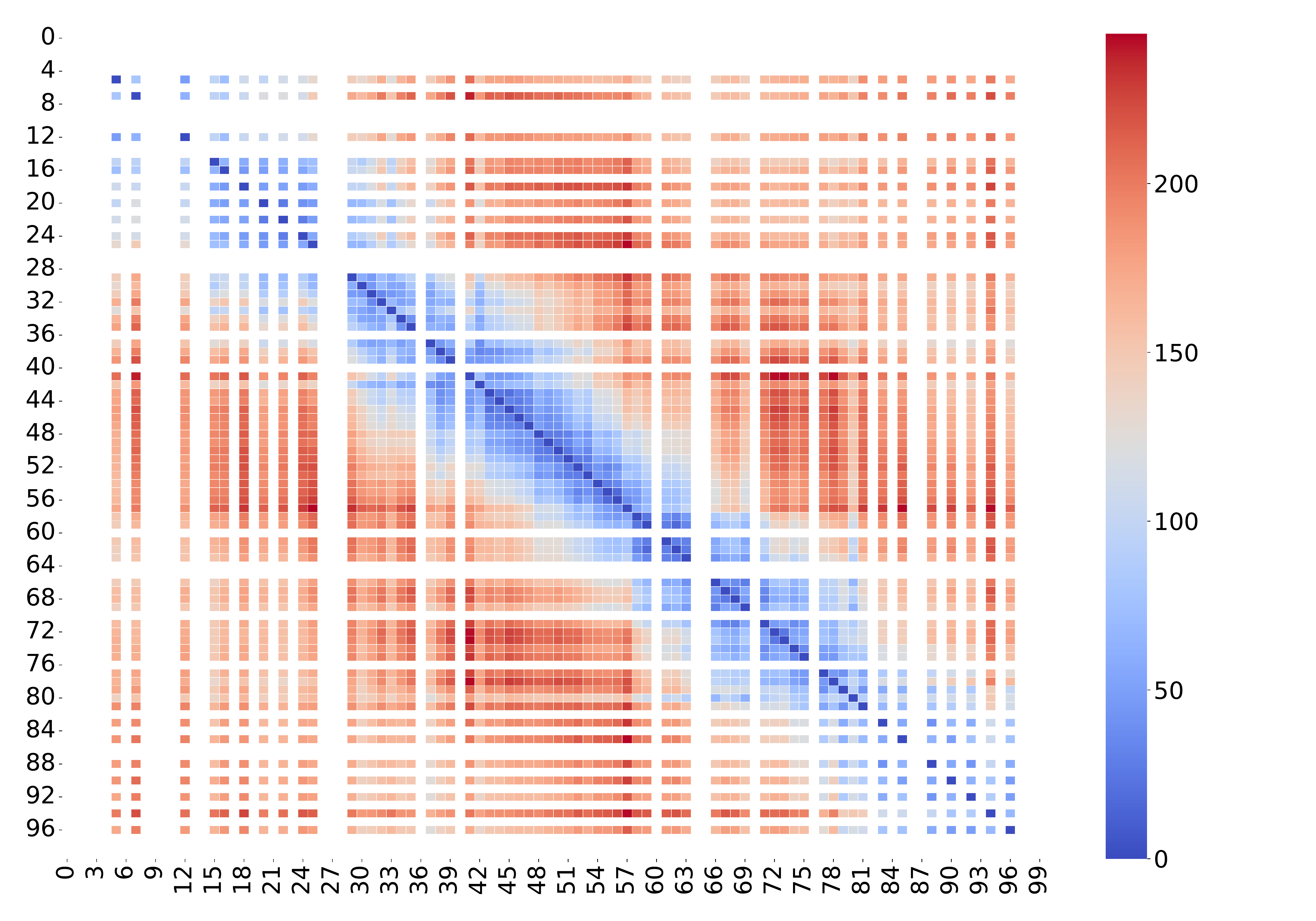}
   \caption{With regularization: $\beta=5.0$}
   \label{fig:cofwr15}
 \end{subfigure}
 \caption{(a) and (b) show the L1 distance between pairs of encodings for FLD1\_s benchmark for $\beta=0$ and $\beta=5.0$, respectively. Each cell (i,j) in this matrix represents the L1 distance between learned encodings for label $i$ and $j$, i.e., $||E_{i,:} - E_{j,:}||_1$.  }
\label{fig:r1abcomp}
\end{figure}
\begin{figure}[h]
 \centering
 \begin{subfigure}[t]{0.49\linewidth}
     \centering
     \includegraphics[width=\textwidth]{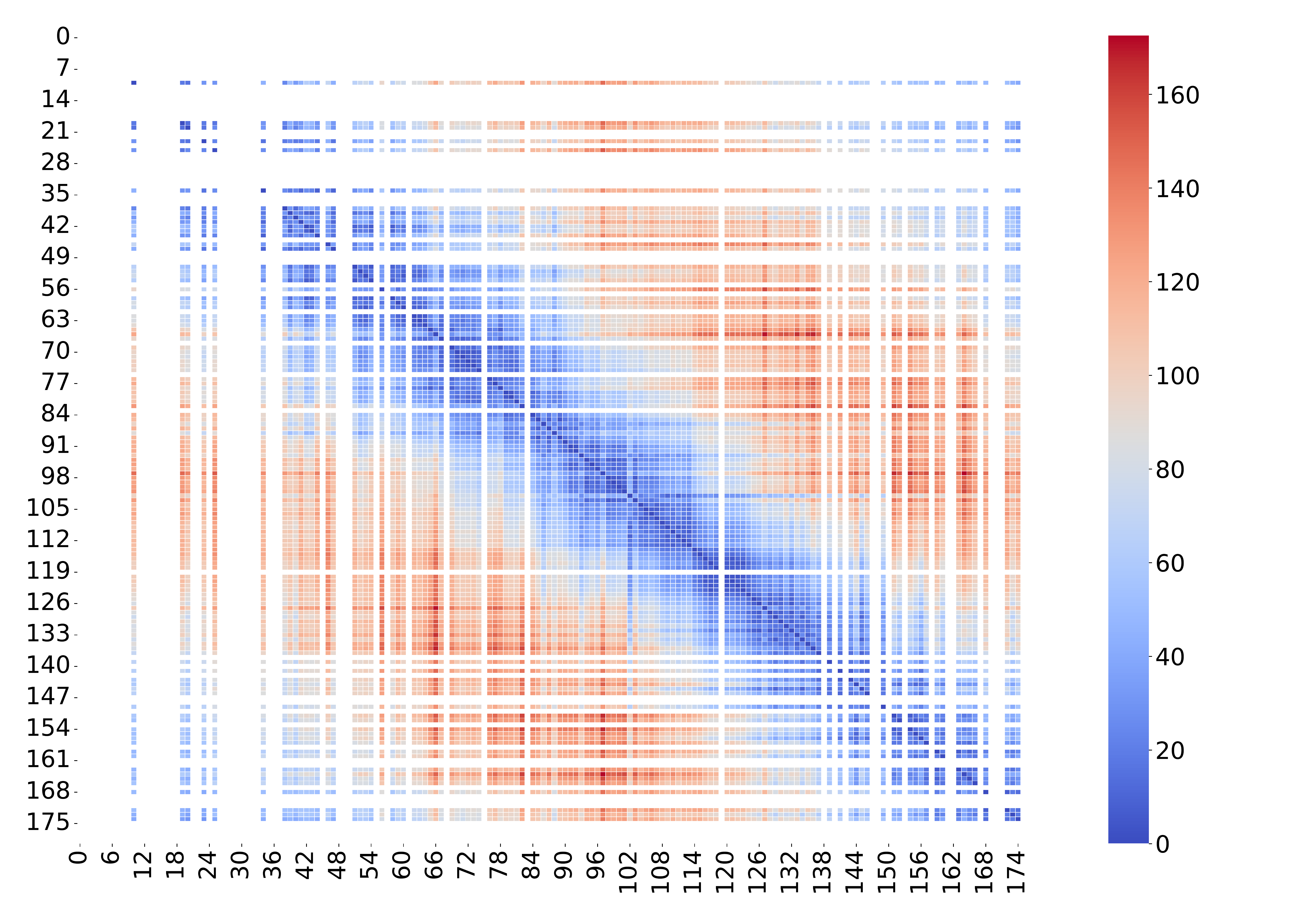}
     \caption{Without regularization: $\beta=0$}
     \label{fig:300wr10}
 \end{subfigure}
 \begin{subfigure}[t]{0.49\textwidth}
   \centering
   \includegraphics[width=\textwidth]{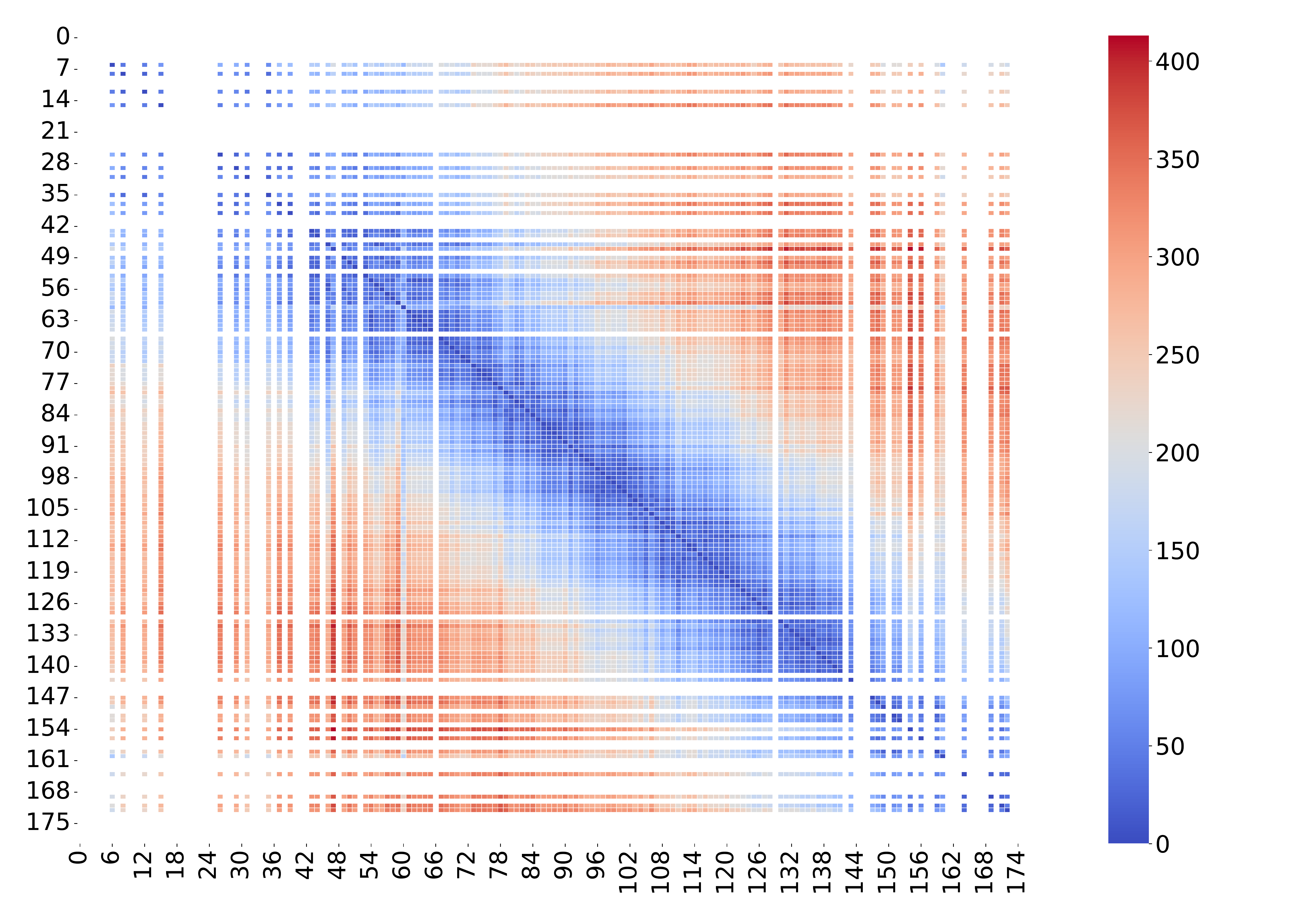}
   \caption{With regularization: $\beta=5.0$}
   \label{fig:300wr15}
 \end{subfigure}
 \caption{(a) and (b) show the L1 distance between pairs of encodings for FLD2\_s benchmark for $\beta=0$ and $\beta=5.0$, respectively. Each cell (i,j) in this matrix represents the L1 distance between learned encodings for label $i$ and $j$, i.e., $||E_{i,:} - E_{j,:}||_1$.  }
\label{fig:r1abcomp300w}
\end{figure}
We proposed regularization function R1 to encourage the L1 distance between encodings to be proportional to the difference between corresponding label values.
Figure~\ref{fig:cofwr10} and Figure~\ref{fig:cofwr15} represent the L1 distance between pairs of learned encodings for FLD1\_s benchmark without and with regularization, respectively.
The $X$-axis and $Y$-axis represent the label values.
Here, some columns and rows are replaced by white lines, as these label values are not present in the training dataset.
The data point at coordinates $(i,j)$ represent the L1 distance between encodings for label $i$ and $j$, i.e., $||E_{i,:} - E_{j,:}||_1$.
For example, in Figure~\ref{fig:cofwr10}, the L1 distance between encodings for label values $0$ and $97$ is $\sim 120$ (light-blue coloured point at coordinate ($0,97$)).
In Figure~\ref{fig:cofwr15}, the L1 distance between encodings for label values $4$ and $96$ is $\sim 170$ (red coloured point at coordinate ($4,96$)).

The first design property (Section~\ref{sec:proposed}) states that the L1 distance between encodings should increase with the difference between corresponding label values.
The difference between label values for pairs of encodings increases with the distance from the diagonal of this plot.
Thus, the value of data points (i.e., the L1 distance between encodings) should increase with the distance from the diagonal of this plot.
As shown in Figure~\ref{fig:cofwr10}, without regularization, the distance between encodings is less for faraway label values (blue-colored data points away from diagonal), which shows that learned encodings do not follow the proposed design property. As shown in Figure~\ref{fig:cofwr15}, the introduction of regularization function R2 remedies this and increases the L1 distance between encodings for faraway labels. Similar observations are made for FLD2\_s benchmarks, as shown in Figure~\ref{fig:300wr10} and Figure~\ref{fig:300wr15}.

\begin{figure}[h]
   \centering
   \begin{subfigure}[t]{0.49\linewidth}
       \centering
       \includegraphics[width=\textwidth]{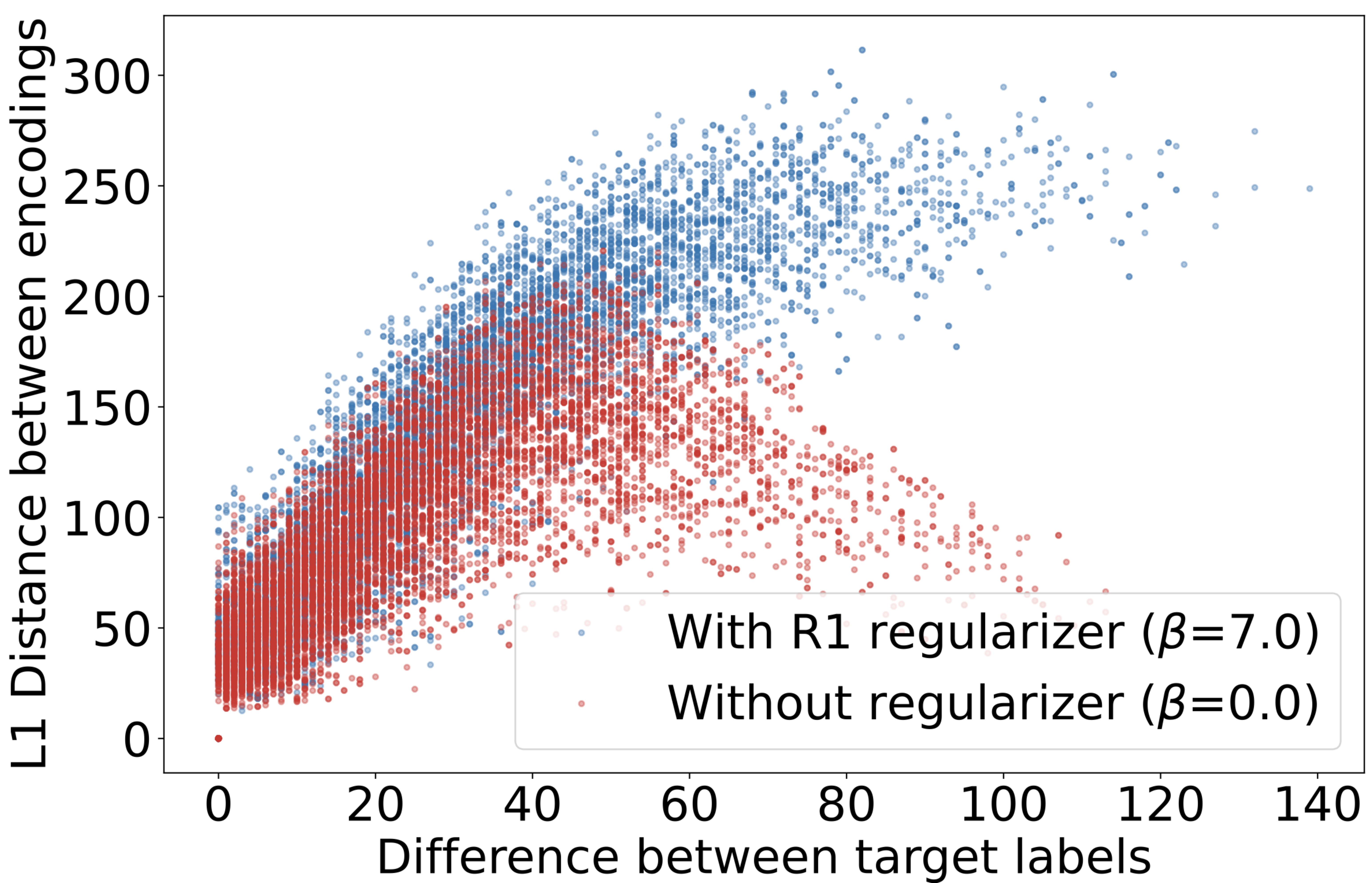}
       \caption{FLD\_1 benchmark}
       \label{fig:r1cofw}
   \end{subfigure}
   \begin{subfigure}[t]{0.49\textwidth}
     \centering
     \includegraphics[width=\textwidth]{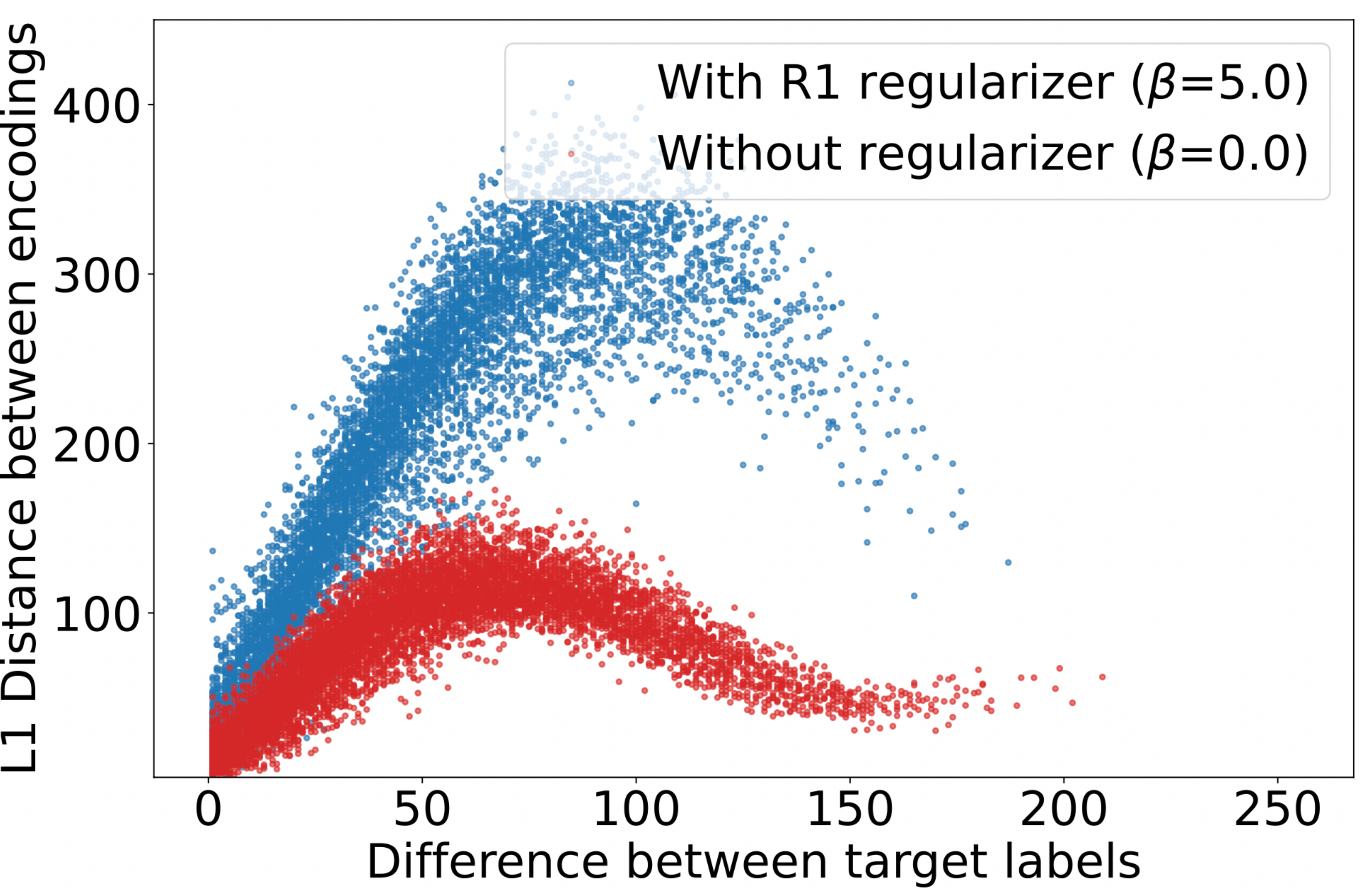}
     \caption{FLD\_1 benchmark}
     \label{fig:r1300w}
   \end{subfigure}
   \caption{(a) and (b) plot the L1 distance between pairs of encodings versus distance between corresponding label values for FLD1\_s and FLD2\_s benchmarks.  }
 \label{fig:r1abcompfull}
 \end{figure}
 Figure~\ref{fig:r1abcompfull} plots the L1 distance between encodings versus the difference between corresponding label values for benchmarks FLD1\_s and FLD2\_s. For both the benchmarks, the proposed regularizer R1 helps enforce the first design property for real-valued label encodings and results in better label encodings with lower error (Table~\ref{tab:compare_code}).

 \paragraph{Effect of the scaling parameter in~\eqref{eq:r1}}
 We use the scaling parameter $2$ in~\eqref{eq:r1}. Our intuition behind using the scaling parameter $2$ is based on binary-encoded labels. For two adjacent labels (i.e., $|y_i-y_j|=1$), the loss function encourages $||\hat{Z}_i-\hat{Z}_{j}||_1$ to be greater than $2$. Here, $\hat{Z}$ is the output encodings. In the case of binarized label encoding ($-1$ if $Z<0$ and $+1$ if $Z>0$), $||Z_i-Z_j||_1=2$ signifies that two encodings differ in at least one bit.

   We also analyzed the effect of changing this parameter for two benchmarks. Table~\ref{tab:scale} shows the impact of changing this scaling parameter for two benchmarks. We observe that the error is higher if the scaling parameter is too low, as encodings for two adjacent labels can not be discriminated against. If this parameter is set too high, the encoding space is more constrained and consequently the performance is degraded.
  
   Based on this intuition and empirical verification on two benchmarks, we use the value $2$ for all benchmarks.
   \begin{table}[]
       \centering
       \caption{Effect of the scaling parameter on error for FLD1$_s$ and FLD2$_s$ benchmarks. }
       \label{tab:scale}
       \begin{tabular}{lrr}
       \toprule
       \multicolumn{1}{l}{Value of the scaling parameter} & \multicolumn{1}{l}{NME (FLD1$_s$)} & \multicolumn{1}{l}{NME (FLD2$_s$)} \\ \midrule
       1                                                    & 4.89                              & 4.15                              \\ \hline
       2                                                    & 4.71                              & 4.15                              \\ \hline
       3                                                    & 4.83                              & 4.20                              \\ \hline
       4                                                    & 4.97                              & 4.27                              \\ \hline
       5                                                    & 4.95                              & 4.28                              \\ \hline
       6                                                    & 5.06                              & 4.41                              \\ \bottomrule
       \end{tabular}
       \end{table}
\subsubsection{Impact of Regularizer R2}
\label{sec:a13}
The regularization function R2 is proposed to reduce the number of bit transitions in the learned label encoding. Figure~\ref{fig:r2comp} compares the label encodings learned for LFH1 benchmark for different values of $\alpha$, where $\alpha$ is the weight of regularization function R2 (\Eqref{eq:loss}). Each row $k$ is the encoding for label value $k$. Each column $k$ represents the output of the encoding position $k$ for different label values.
The regularization function is proposed to decrease the transitions in an encoding bit (blue$\to$red and red$\to$blue) over the range of label values. Section~\ref{sec:abl} provided quantitative results to demonstrate that increasing the value of $\alpha$ reduces the number of bit transitions. We observe similar trends in the plots of learned label encodings shown in Figure~\ref{fig:r2comp}; increasing the value of $\alpha$ decreases bit transitions in the learned label encodings and improves MAE.
\begin{figure}[h]
 \centering
 \begin{subfigure}[t]{0.49\linewidth}
     \centering
     \includegraphics[width=\textwidth]{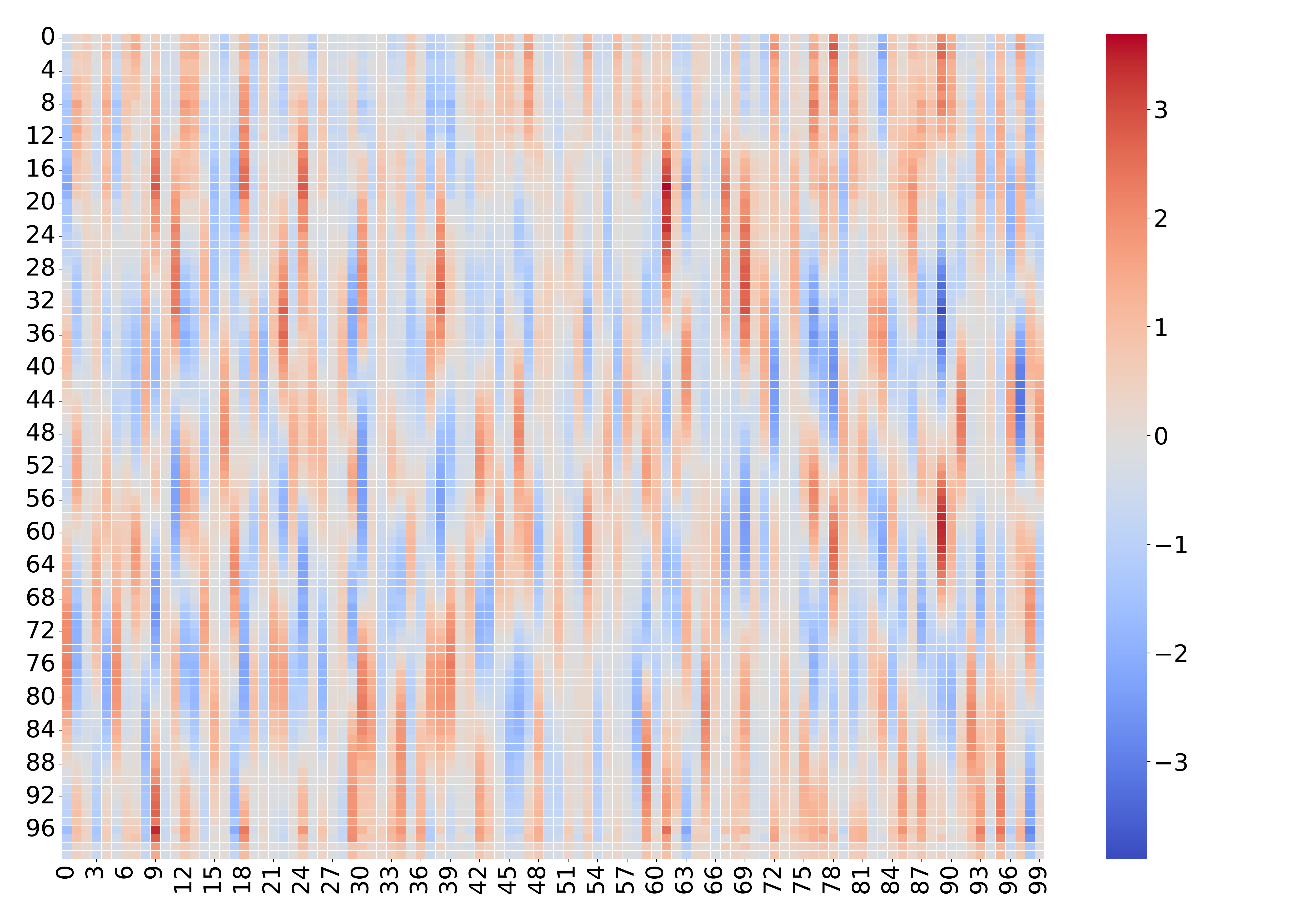}
     \caption{Without regularization: $\alpha=0$, MAE= $4.13$ }
     \label{fig:r20}
 \end{subfigure}
 \begin{subfigure}[t]{0.49\textwidth}
   \centering
   \includegraphics[width=\textwidth]{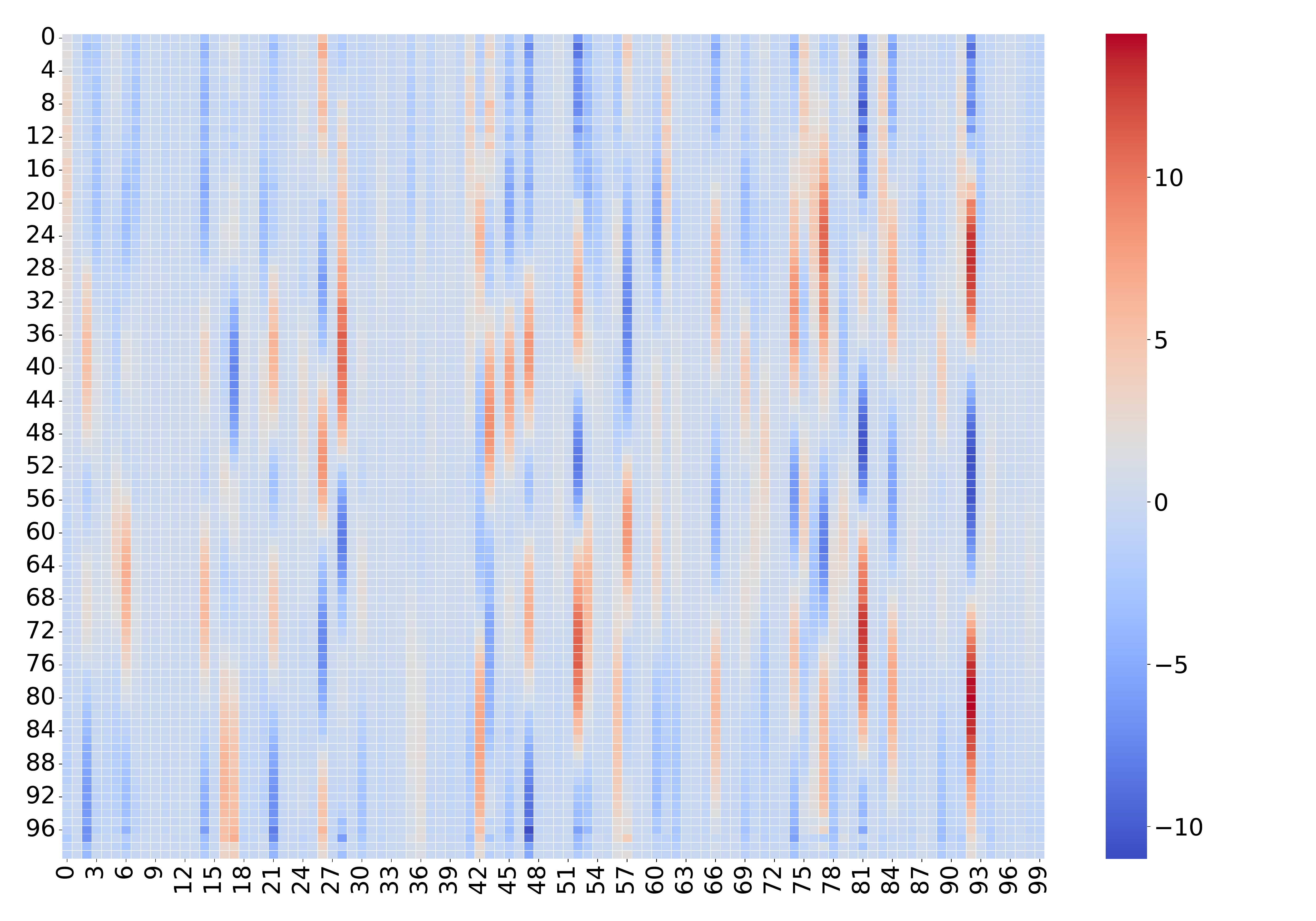}
   \caption{$\alpha=0.1$, MAE= $3.76$}
   \label{fig:r20o1}
 \end{subfigure}
 \\
 \begin{subfigure}[t]{0.49\linewidth}
    \centering
    \includegraphics[width=\textwidth]{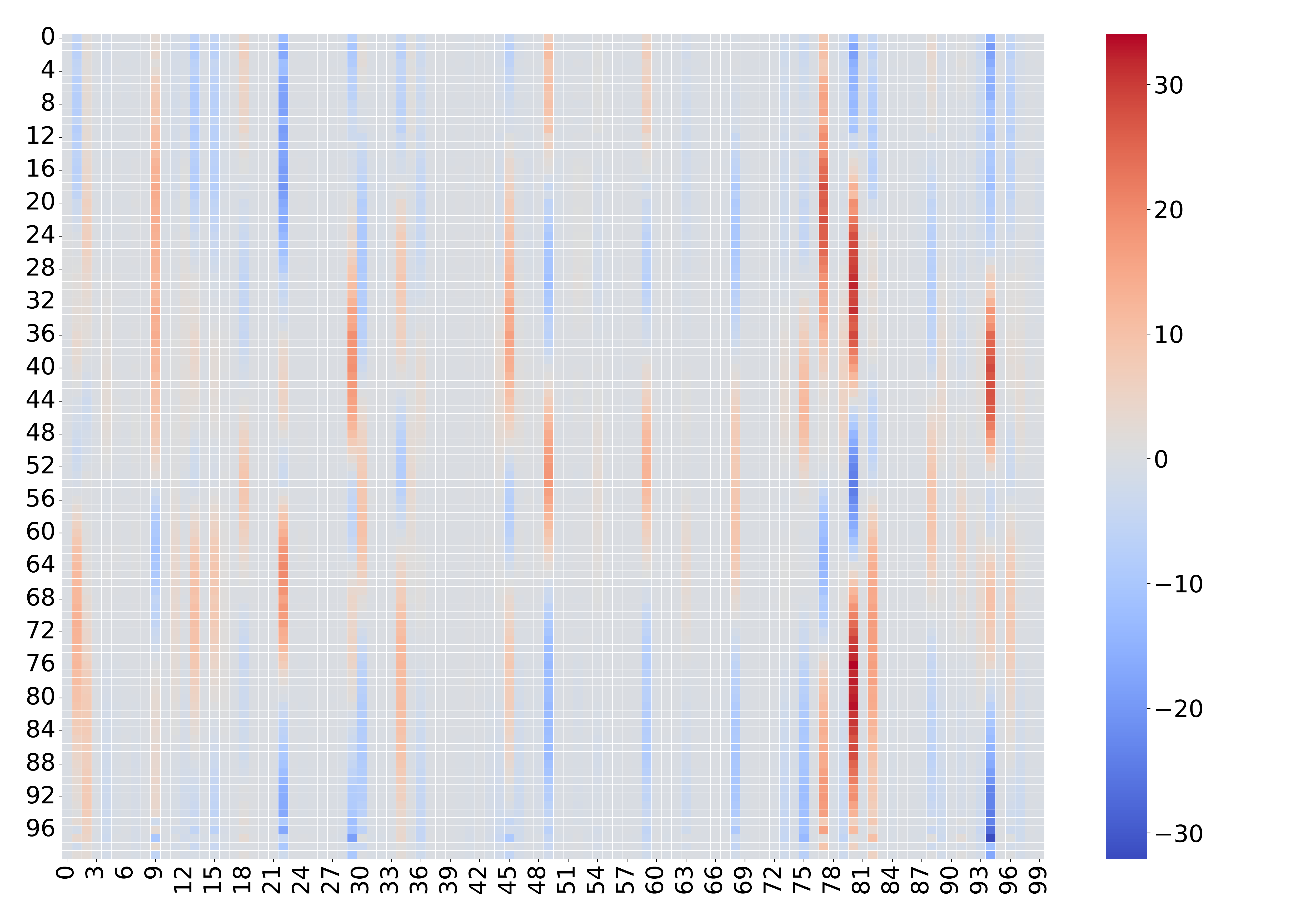}
    \caption{$\alpha=0.5$, MAE= $3.73$}
    \label{fig:r20o5}
\end{subfigure}
\begin{subfigure}[t]{0.49\textwidth}
  \centering
  \includegraphics[width=\textwidth]{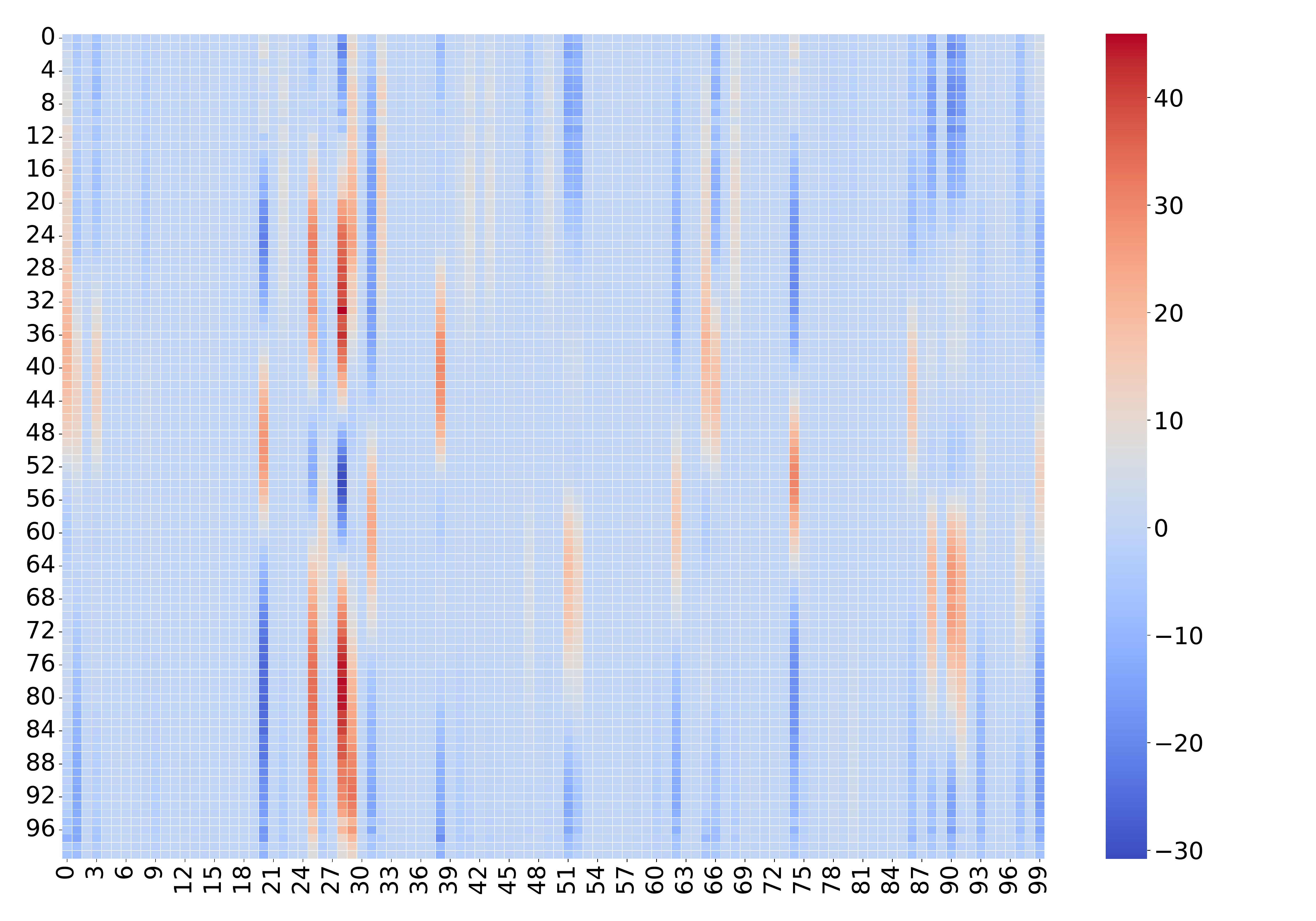}
  \caption{$\alpha=1.0$, MAE= $3.56$}
  \label{fig:r21o5}
\end{subfigure}
 \caption{(a)-(d) represent the label encodings learned by \CLL{ } for different values of weight $\alpha$ for regularizer R2 (\Eqref{eq:loss}).   }
\label{fig:r2comp}
\end{figure}
\subsubsection{Effect of hyperparameters in \CLL}
\label{sec:hype}
\begin{figure}[t]
   \centering
     \includegraphics[width=0.65\textwidth]{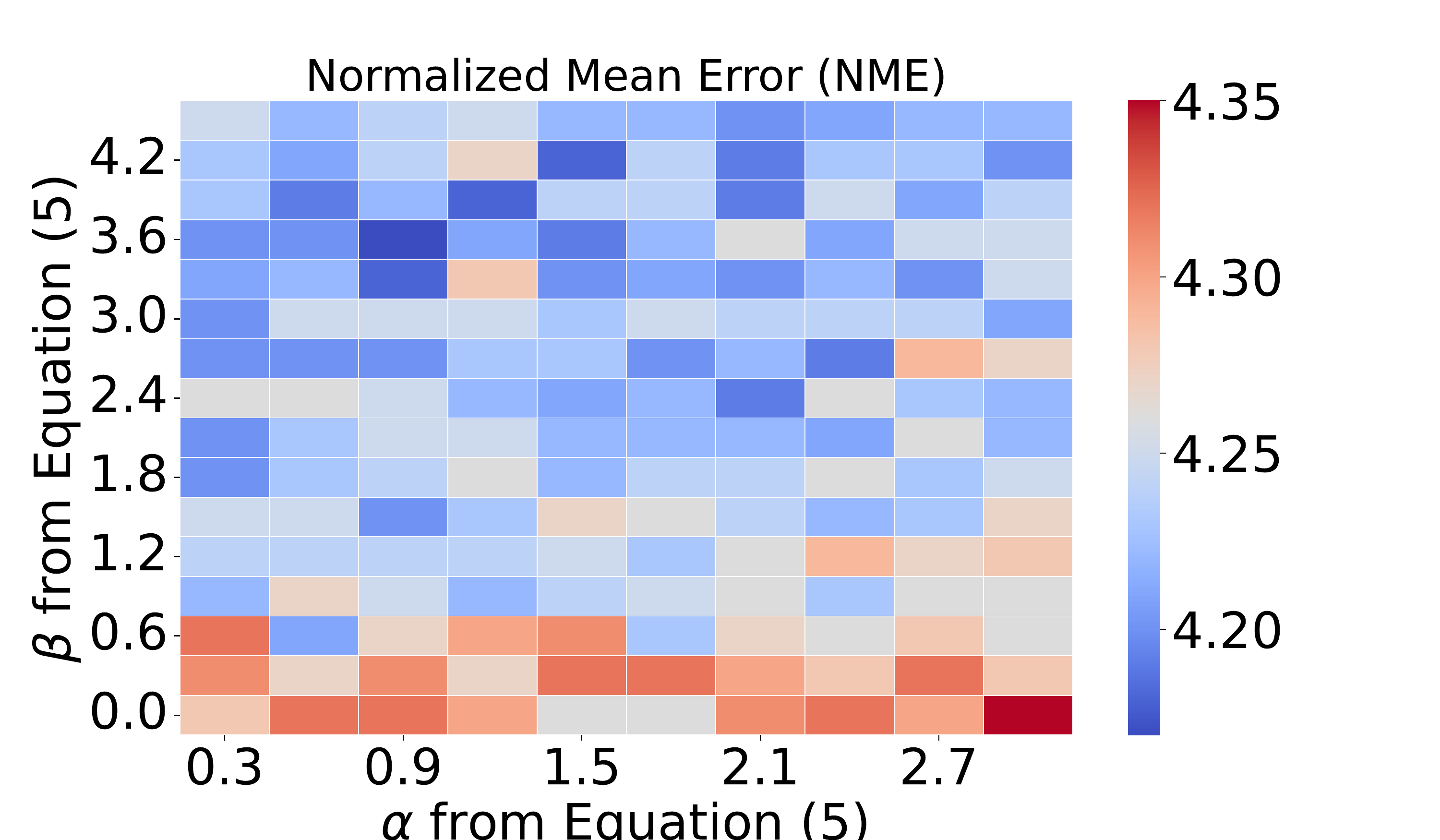}
   \caption{Impact of hyperparameters $\alpha$ and $\beta$ from \Eqref{eq:loss} on NME for FLD1\_s benchmark.  }
 \label{fig:hypersweep}
\end{figure}
   The \CLL{ }approach introduces two hyperparameters.
   We first evaluate the sensitivity to these hyperparameters to determine the complexity of hyperparameter tuning.
   Figure~\ref{fig:hypersweep} shows the NME for FLD1\_s benchmark for different values of $\alpha$ and $\beta$ values in \Eqref{eq:loss}. As shown in the figure, the error is not sensitive to small changes in these hyperparameters' values, suggesting that a sparse search in the hyperparameter space suffices.
   Furthermore, several approaches have been proposed for efficient hyperparameter search~\citep{hyperband,Falkner2018BOHBRA}, and any off-the-shelf hypermeter tuners/libraries can be used to automatically find these values without manual efforts.
   In contrast, hand-designed codes need human intervention to design codes. Also, multiple training runs are still required to find suitable codes for a given benchmark from a set of hand-designed codes.
   On the other hand,  \CLL{ }provides an end-to-end automated approach for label encoding learning.
\subsubsection{Impact of the number of fully-connected layers:}
\label{sec:a14}
\begin{table}[]
  \caption{Impact of the number of fully-connected layers in direct regression and multiclass classification on the error (MAE or NME). This table is reproduced from~\citep{ShahICLR2022}. }
  \label{tab:FCL1}
  \centering
  \scriptsize
  \begin{tabular}{lrrrr}
      \toprule
  \multirow{2}{*}{Benchmark} & \multicolumn{2}{c}{Direct regression} & \multicolumn{2}{c}{Multiclass classification}  \\ \cline{2-5}
   & \multicolumn{1}{c}{1 FC layer} & \multicolumn{1}{c}{2 FC layers} & \multicolumn{1}{c}{1 FC layer} & \multicolumn{1}{c}{2 FC layers} \\ \midrule
  LFH1 & 4.76 & 5.19 & 4.49 & 4.82  \\ \hline
  LFH2 & 5.65 & 5.59 & 5.31 & 5.42  \\ \hline
 %% HPE3 & 3.40 & 3.54 & 4.45 & 4.54  \\ \hline
%%  HPE4 & 4.14 & 4.22 & 5.14 & 5.45  \\ \hline
  FLD1 & 3.60 & 3.63 & {3.58} & 3.56   \\ \hline
  FLD2 & 3.54 & 3.58 & 3.51 & 3.62   \\ \hline
  FLD3 & 4.64 & 4.63 & 4.50 & 4.64   \\ \hline
  FLD4 & 1.51 & 1.51 & 1.56 & 1.53   \\ \hline
  AE1 & 2.44 &  2.35 & 2.75 &  2.81   \\ \hline
  AE2 & 3.21 & 3.14  & 3.38 & 3.40    \\ \hline
  PN & 4.24 &  4.33 & 4.56 &  5.74   \\ \bottomrule
  \end{tabular}
  \end{table}
For \CLL{ }, we use an extra fully connected bottleneck layer in the regressor as proposed by the prior work on regression by binary classification~\citep{ShahICLR2022}. We provide an ablation study (reproduced from~\citep{ShahICLR2022}) to show the impact of additional fully connected layers in direct regression and multiclass classification.
Table~\ref{tab:FCL1} provides the error (MAE or NME) for direct regression and multiclass classification with one or two fully connected layers after the feature extractor. As shown in the table, increasing the number of fully connected layers in direct regression and multiclass classification does not reduce the error for most benchmarks (possibly due to overparameterization).
\subsubsection{Comparison with Deep Hashing Approaches}
\label{app:hash}
Deep supervised hashing approaches use neural networks as a hash function and learn hash codes in an end-to-end manner.
The loss function for deep supervised hashing is designed to preserve the similarity between inputs in the hashing space.
Often, these approaches use the label information to determine the similarity between images (i.e., same label)~\citep{sdh,cnnh}.
Some deep hashing approaches have proposed to augment the loss function with classification loss to improve the performance.
We adapt two widely used deep-hashing approaches to regression and compare~\CLL{ }with deep hashing approaches.

\citet{sdh} proposed a deep supervised hashing (DSH) approach with a loss function based on the pairwise similarity between images.
The proposed approach introduces a loss function to preserve the similarity between output codes for similar training images (e.g., same class) and maximize discriminability between output codes for different training images (e.g., different class).
Further, they propose using relaxation on the binary output and a regularizer to encourage the output code to be close to discrete values $+1/-1$. The hamming distance between output codes can be computed for binary-like outputs using the L2 norm.
We use DSH for regression with some modifications (DSH-reg).
We used the quantized label to determine the class of a training sample.
%First, the number of samples in regression is typically lower, and the number of labels (Quantized levels) is higher than in classification datasets. Thus forming pairs of images with similar labels results in a significant imbalance between the number of image pairs with similar and different labels.
%To overcome this issue, we modify the loss function to encourage the hamming distance between codes for two images to be proportional to the similarity between corresponding labels (i.e., the difference between label values).

\citet{pairsdh} proposed a triplet ranking loss to learn a hash function that preserves relative similarities between images (SFLH). For images $(I, I+, I-)$, where $I$ is closer to $I+$ than $I-$, the loss function is designed to encourage higher hamming distance between codes for $(I,I-)$ than $(I,I+)$. For classification datasets, triplets are typically formed using two images from the same class and one from a different class~\citep{hashtriplet}. They proposed to use a piece-wise threshold function to encourage binary-like outputs.

We use the above approach (SFLH) for regression with a few modifications (SFLH-reg). To generate triplets, we pick sets of three images from a given batch and determine the similarity between images using differences between the label values.  We use $K^2$ triplets for a minibatch of $K$ training samples.

Further, for both DSH-reg and SFLH-reg, we augment the loss function with regression loss. We add a fully-connected layer between the output code and prediction. The MSE loss between the final outputs and target labels is added to the loss function (DSH-reg-L2, SFLH-reg-L2).
\begin{table}[h]
   \centering
   \setlength\tabcolsep{6pt}
   \caption{Comparison of \CLL with different deep hashing approaches adapted for regression. }
   \label{tab:deephash}
   \footnotesize
   \begin{tabular}{C{3cm}R{3cm}}
       \toprule
       Method & \multicolumn{1}{c}{MAE} \\ \midrule
       DSH-reg & 71.3 \\ \hline
       DSH-reg-L2 & 4.11 \\ \hline
       SFLH-reg & 69.8 \\ \hline
       SFLH-reg-L2 & 4.73 \\ \hline
       RLEL ( only R1 ) & 3.93 \\ \hline
       RLEL ( R1 + R2 ) & 3.55 \\ \bottomrule
       \end{tabular}
       \end{table}

Table~\ref{tab:deephash} compares the modified deep hashing approaches with \CLL. The gap between loss functions with and without regression loss is significant, which shows that a loss function that only aims to preserve the similarity between output codes is not sufficient and needs to account for the error between decoded output and target (i.e., regression loss). \CLL{ }results in a lower error as it is designed for regression problems that account for classifiers' nonuniform error probability distribution.

Regularizer R1 encourages the distance between output codes for images to be proportional to the difference between label values, similar to pairwise or ranking-based loss functions proposed by deep hashing. However, deep hashing approaches use the hamming distance between binary outputs.
As we show in Section~\ref{subsec:const}, the hamming distance between codes does not account for the error probability of classifiers. Thus we use the L1 distance between the real-valued outputs to account for the confidence of the classifiers. R1 does not use regularizer or nonlinear activation on the output codes to encourage binary-like outputs, as typically done in deep hashing approaches. In contrast, we show that suitable regression codes can be learned by not using this constraint. Thus \CLL{ }with only R1 regularizer results in lower error than deep hashing approaches.

\subsubsection{Evaluation}
\label{sec:a11}
\begin{table}[h]
 \centering
 \setlength\tabcolsep{4pt}
 \caption{Comparison of \CLL with different regression approaches using Geometric mean and Pearson coefficient as evaluation metrics. }
 \label{tab:evalmetrics}
 \footnotesize
 \begin{tabular}{lcC{1cm}cC{1cm}cC{1cm}}
 \toprule
 \multirow{2}{*}{} & \multicolumn{2}{C{3cm}}{RLEL} & \multicolumn{2}{c}{Direct Regression} & \multicolumn{2}{c}{Multiclass Classification} \\ \cline{2-7}
  & \multicolumn{1}{c}{GeoMean} & \multicolumn{1}{c}{Pearson Coeff.} & \multicolumn{1}{c}{GeoMean} & \multicolumn{1}{c}{Pearson Coeff.} & \multicolumn{1}{c}{GeoMean} & \multicolumn{1}{c}{Pearson Coeff.} \\ \midrule
 LFH1 & \multicolumn{1}{c}{1.95} & 97.68 & \multicolumn{1}{c}{2.91} & 97.10 & \multicolumn{1}{c}{2.30} & 94.60 \\ \hline
 LFH2 & \multicolumn{1}{c}{2.09} & 92.22 & \multicolumn{1}{c}{2.49} & 91.06 & \multicolumn{1}{c}{2.40} & 88.76 \\ \hline
 FLD1 & \multicolumn{1}{c}{0.96} & 99.94 & \multicolumn{1}{c}{1.07} & 99.93 & \multicolumn{1}{c}{1.04} & 99.93 \\ \hline
 FLD1\_s & \multicolumn{1}{c}{1.31} & 99.87 & \multicolumn{1}{c}{6.38} & 99.81 & \multicolumn{1}{c}{1.81} & 99.80 \\ \hline
 FLD2 & \multicolumn{1}{c}{1.92} & 99.97 & \multicolumn{1}{c}{2.12} & 99.97 & \multicolumn{1}{c}{2.07} & 99.97 \\ \hline
 FLD2\_s & \multicolumn{1}{c}{2.44} & 99.96 & \multicolumn{1}{c}{3.03} & 99.98 & \multicolumn{1}{c}{3.22} & 99.94 \\ \hline
 FLD3 & \multicolumn{1}{c}{0.96} & 99.99 & \multicolumn{1}{c}{1.05} & 99.99 & \multicolumn{1}{c}{1.01} & 99.97  \\ \hline
 FLD3\_s & \multicolumn{1}{c}{1.21} & 99.98 & \multicolumn{1}{c}{1.56} & 99.97 & \multicolumn{1}{c}{1.37} & 99.97 \\ \bottomrule
 \end{tabular}
\end{table}

{Table~\ref{tab:evalmetrics} compares \CLL{ }with direct regression and multiclass classification using geometric mean and Pearson coefficient as evaluation metrics. The geometric mean represents the geometric mean of absolute error for the test dataset. The Pearson coefficient represents the correlation between the target and predicted labels for the test dataset.  As shown in the table, \CLL{ }results in significant reduction in the error compared to other generic regression approaches. }

\subsubsection{Theoretical analysis of proposed regularization functions}
\label{sec:r2proof}
{
\begin{figure}[h]
   \centering
   \begin{subfigure}[t]{0.49\linewidth}
       \centering
       \includegraphics[width=\textwidth]{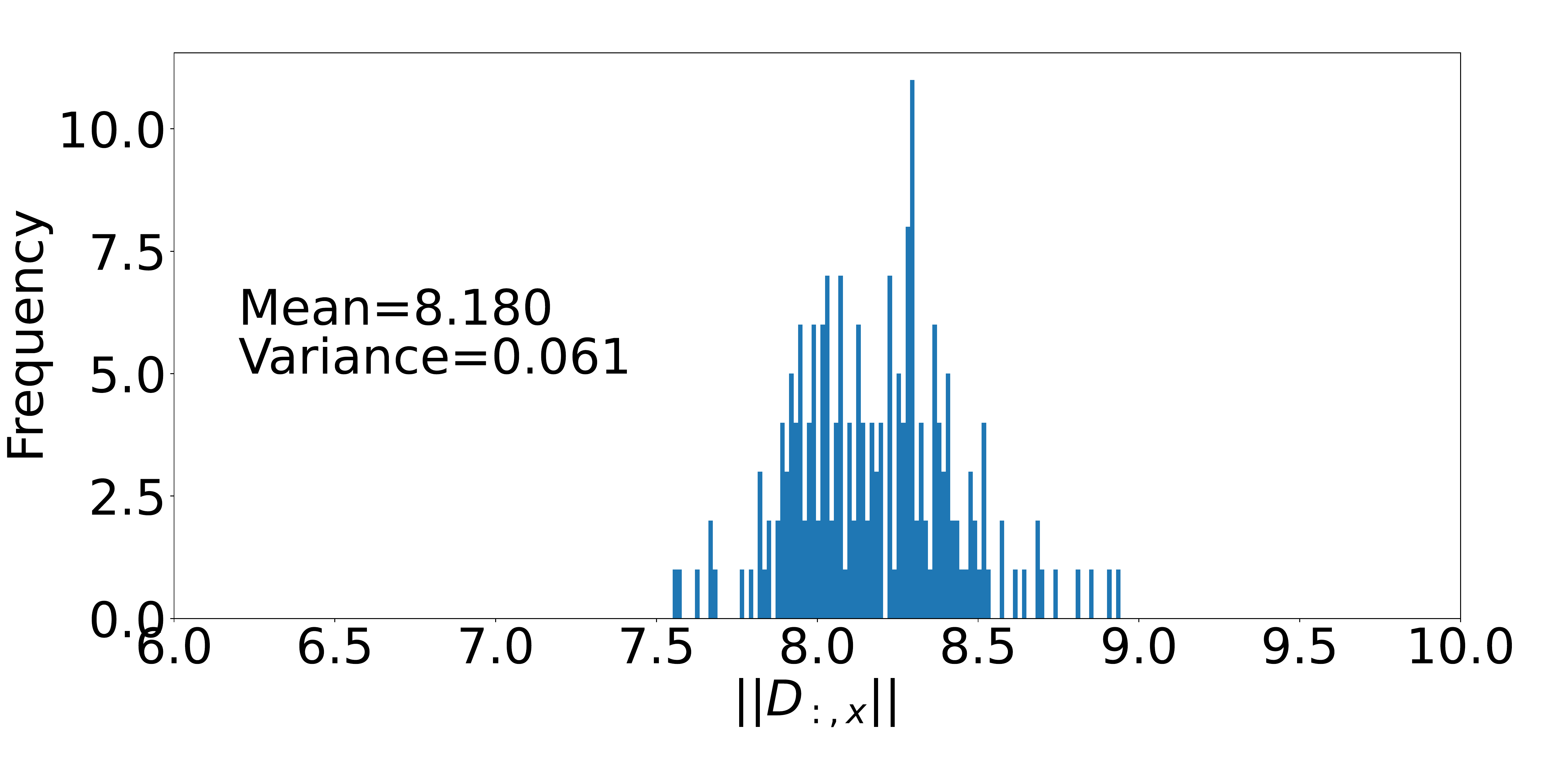}
       \caption{LFH2 benchmark - Pitch label}
       \label{fig:dlfh1}
   \end{subfigure}
   \begin{subfigure}[t]{0.49\textwidth}
     \centering
     \includegraphics[width=\textwidth]{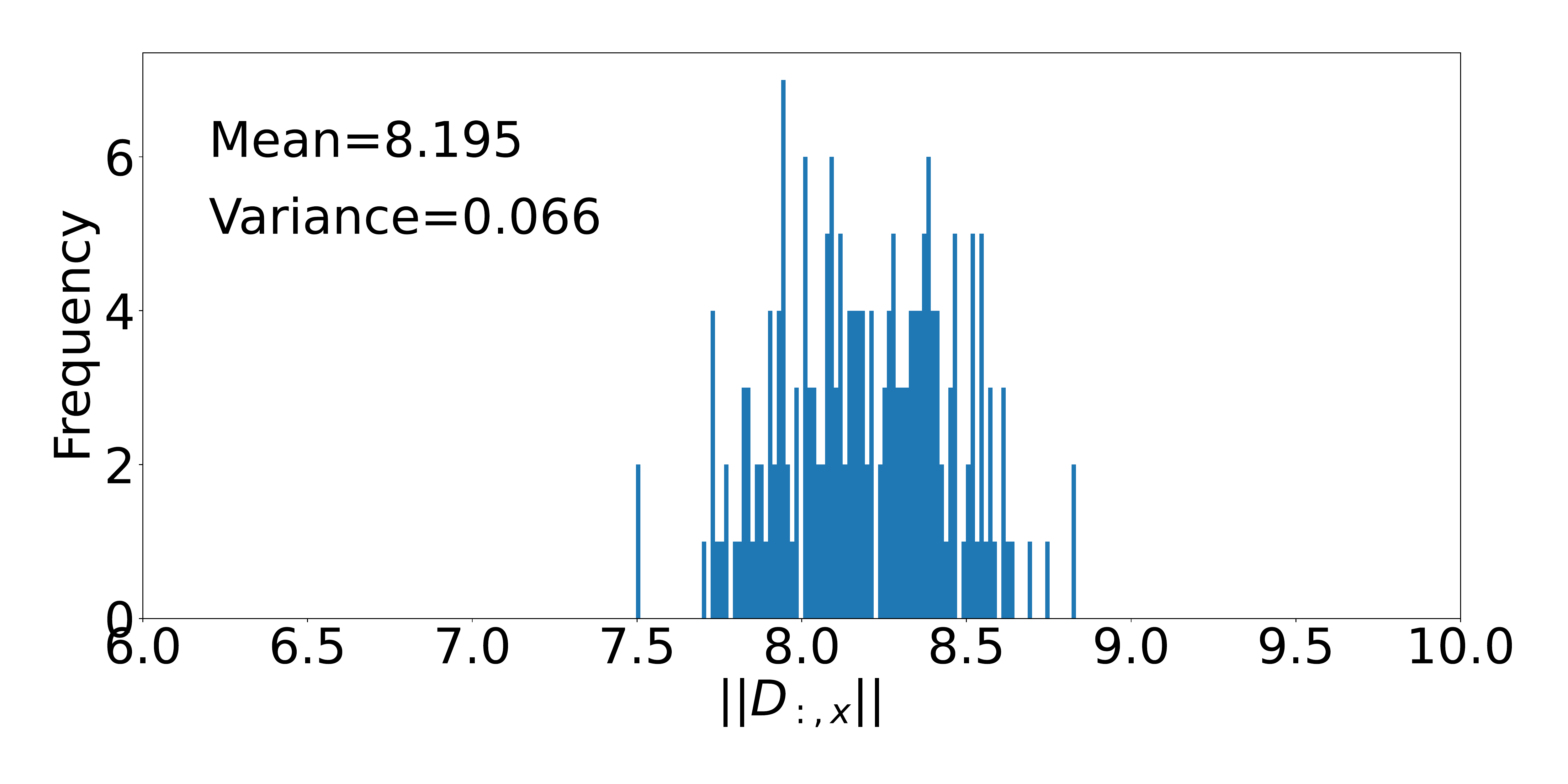}
     \caption{LFH2 benchmark - Roll label}
     \label{fig:dlfh1}
   \end{subfigure}
   \begin{subfigure}[t]{0.49\textwidth}
       \centering
       \includegraphics[width=\textwidth]{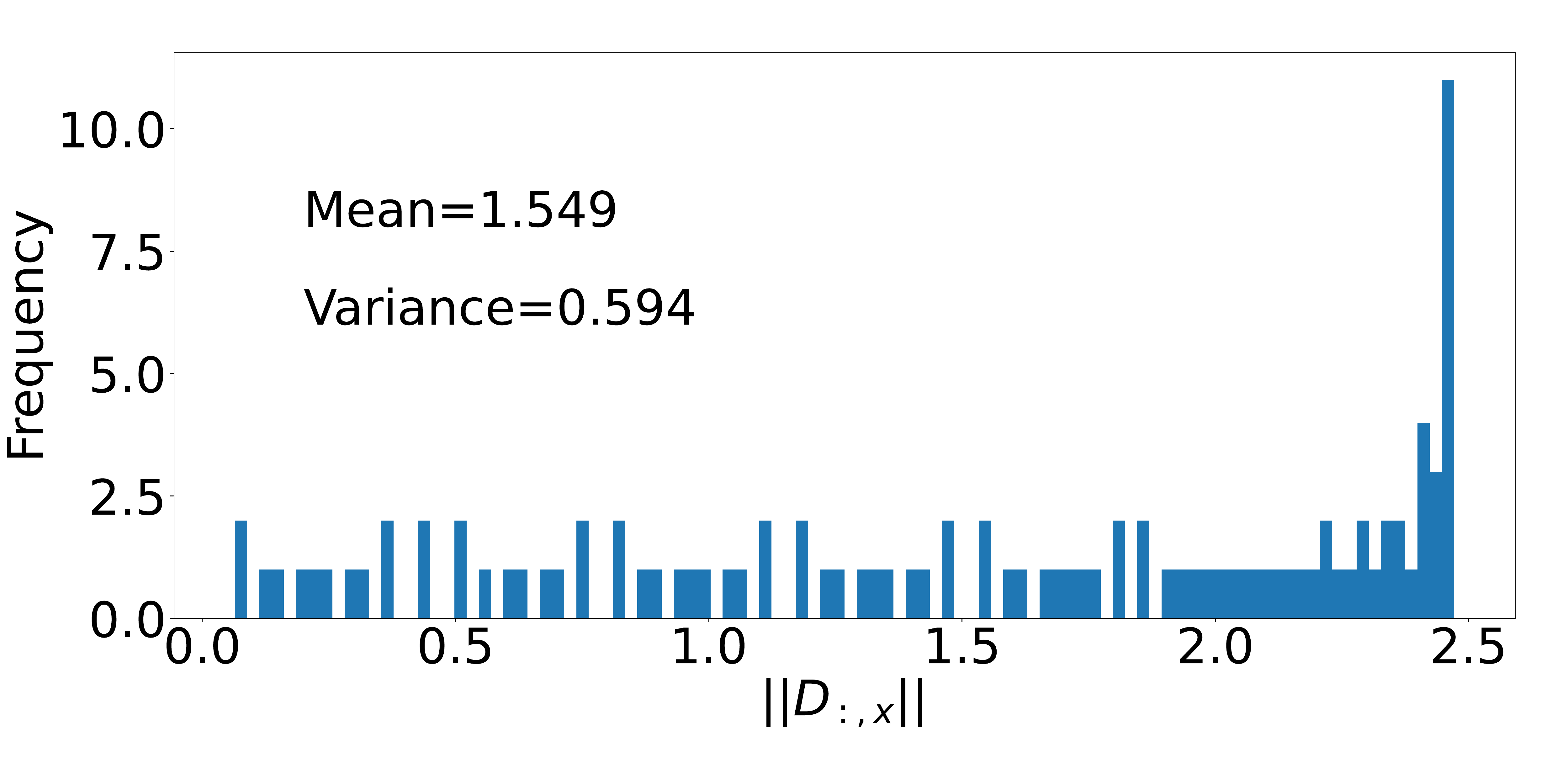}
       \caption{LFH1 benchmark - Roll label}
       \label{fig:dlfh1}
     \end{subfigure}
   \caption{(a) and (b) plot the distribution of $||D_{:,x}||$ for LFH2 benchmark. (c) plots the distribution of $||D_{:,x}||$ for LFH1 benchmark. Here the variance is very low, which suggests that the assumption $||D_{:,x}||\approx ||D_{:y}||, x\in[1,N],y\in[1,N]$ is valid. For the LFH1 benchmark, the variance is higher than LFH2. However, all outliers are for label values with very few (or even zero) training examples.  }
 \label{fig:dlfh1full}
 \end{figure}
 \subsubsubsection{\underline{Regularization function R2:}}
 \\

 We used matrix $D$ instead of label encoding $E$ to apply regularizer R2 in~\eqref{eq:r2}. We insight into this decision as follows. First, note the output encodings are multiplied with $D$ to generate the correlation vector $\hat{C}_i$ (Figure 3). We use the multiclass classification loss between $\hat{C}_i$ and the target labels for training. Due to this, label encoding $E$ and decoding matrix $D$ are related, and use of matrix $D$ proves to be effective for regularizer R2. We further explain this in detail below.
 \\

 Let $E$ represent an encoding matrix of size $N\times M$.
 Each row $E_{k:}$ represents the encoding output when the label is $k$.
 $D$ is the decoding matrix of size $M\times N$.
 Let $\hat{C}_{k}$ represent the output correlation row vector of size $1\times N$ when the target label is $k$. Here,  $\hat{C}_{k}$ is obtained by multiplying  $E_{k,:}$ with  $D$ (Figure~\ref{fig:setup}).
 \begin{equation}
   \label{eq:cref}
   \hat{C}_{k} = E_{k,:} D
 \end{equation}
 Since we apply softmax on the output vector to find the predicted label (Figure~\ref{fig:setup}), ideally, $\hat{C}^k_{k}$ should have the highest value as the target label value is $k$.
 \\ \\
 $\therefore \hat{C}^k_{k} > \hat{C}^x_{k} ,\text{ where, } x \neq k, x\in \{1,2,...,N\}$
 \\ \\
 $ \therefore E_{k,:} . D_{:,k} > E_{k,:} .  D_{:,x},\text{ where, } x \neq k, x\in \{1,2,...,N\}  \text{ (Using~\eqref{eq:cref})}$. 
  Let $\theta_{k,x}$ represent the angle between row vector $E_{k,:}$ and column vector $D_{: ,x}$. This leads to the below equation:
 \begin{equation}
     \label{eq:inter}
     ||E_{k,:}|| ||D_{:,k}|| \text{cos}(\theta_{k,k}) >  ||E_{k,:}|| ||D_{:,x}|| \text{cos}(\theta_{k,x}),\text{ where, } x \neq k, x\in \{1,2,...,N\}
 \end{equation}
  ~\cite{ShahICLR2022} used a hand-crafted decoding matrix $D$ with an equal number of $1$s and $0$s in each column for binary-encoded labels. Hence the L2 norm of each column is the same. In label encoding learning, parameters of matrix $D$ are learned during training and are not constrained to have the same L2 norm for each column. However, we observe a similar trend empirically. Figure~\ref{fig:dlfh1} plots the distribution of  $||D_{:,x}||$ for different benchmarks. As shown in the figure, for most benchmarks, we observe a small variance in the distribution of $||D_{:,x}||$. Based on this intuition and empirical validation, we assume that $||D_{:,x}|| \approx ||D_{:,y}|| $ for $x\in[1, N]$ and $y\in[1, N]$ to simplify the analysis.
 \\ \\
 Using this assumption in~\eqref{eq:inter} leads to the following inequality:
 \\ \\
 $\text{cos}(\theta_{k,k}) > \text{cos}(\theta_{k,x}),\text{ where, } x \neq k, x\in \{1,2,...,N\}$\\ \\
 Thus the cosine similarity between $E_{k,:}$ and $D_{:,k}$ should be the highest to predict the label $k$. The optimization process to reduce the loss between the target and prediction will try to maximize this cosine similarity. In the best case, the angle between  $E_{k,:}$ and $D_{:,k}$ will be zero, and both vectors are parallel.
 \\ \\
 This simplification leads to the following relation between $E$ and $D$. \\ \\
 $E_{k,:} = t D_{:,k}\text{, where } t > 0 $
 \\ \\
 Similarly, $E_{k+1,:} = t' D_{:,k+1}\text{, where } t' > 0 $
 \\ \\
 Since $t$ and $t'$ both are positive values, reducing $D_{i,k}-D_{i,k+1}$ also reduces $E_{k,i}-E_{k+1,i}$. \\ \\
 Regularizer rule R2 proposes to regularize the number of decision boundaries by regularizing $\sum_{i=1}^{M} \sum_{j=1}^{N-1} |{E}_{j,i} - E_{j+1,i}|$ as per~\eqref{eq:bit}. 
 Based on the analysis above, regularizing $\sum_{i=1}^{M} \sum_{j=1}^{N-1} |D_{i,j}-D_{i,j+1} | $ helps with the above goal as $E_{j,i}-E_{j+1,i}$ reduces with $D_{i,j}-D_{i,j+1}$.
  \subsubsubsection{\underline{Regularization function R1:}}
 \\

The first property suggests $||E_{i,:} - E_{j,:}||_1 \propto |i-j|$.
\\ \\
So ideally, $||E_{i,:} - E_{j,:}||_1 = \lambda |i-j|$

Since $E_{x,:}$ is average of $\hat{Z}_i$ for samples with label value $x$ (\eqref{eq:ez}), the above condition leads to: \\
\begin{equation}
   \label{eq:cond1}
   ||\hat{Z}_{i} - \hat{Z}_{j}||_1 = \lambda |y_i-y_j|
\end{equation}
%%$||\hat{Z}_{i} - \hat{Z}_{j}||_1 = \lambda |y_i-y_j|$

Based on this requirement, we add a regularization function max$(0,\lambda |y_i - y_j | - ||\hat{Z}_i - \hat{Z}_j ||_1)$, which penalizes the encodings if $||\hat{Z}_i - \hat{Z}_j ||_1$ < $\lambda |y_i-y_j|$. It does not strictly impose~\eqref{eq:cond1}. However, it approximately imposes the constraint as per shown in empirical verification in Section~\ref{sec:a12}.
\\

Our intuition behind using the scaling parameter $2$ is based on binary-encoded labels. For two adjacent labels (i.e., $|y_i-y_j|=1$), the loss function encourages $||\hat{Z}_i-\hat{Z}_{j}||_1$ to be greater than $2$. Here, $\hat{Z}$ is the output encodings. In the case of binarized label encoding ($-1$ if $Z<0$ and $+1$ if $Z>0$), $||Z_i-Z_j||_1=2$ signifies that two encodings differ in at least one bit.
 }
 \subsubsection{Impact of the number of quantization levels ($N$)}
 \label{sec:quantN}
 {
    % \color{blue}
 The number of quantization buckets is treated as a design parameter for binary-encoded labels.~\cite{ShahICLR2022} showed that the error changes with the number of quantization levels. Fewer levels introduce quantization error, and more levels increase the number of bits in the encoding. They showed a trade-off between these two factors to decide the number of quantization levels.

Our work focuses on the design space of encoding and decoding functions. Hence we use the same values for the quantization levels ($N$) as BEL~\cite{ShahICLR2022}. Parameter $N$ tuning can be integrated into hyperparameter tuning or included in the optimization process.

We further analyze the effect of the number of quantization levels for RLEL. Table~\ref{tab:quantN}  shows the NME (Normalized Mean Error) for different values of $N$ for FLD1 benchmark.
\begin{table}[t]
   \caption{Impact of the number of quantization levels on error for FLD1 benchmark }
   \label{tab:quantN}
   \centering
   \begin{tabular}{lr}
       \toprule
   \multicolumn{1}{l}{Quantization levels (N)} & \multicolumn{1}{l}{NME} \\ \midrule
   32                                          & 3.49                    \\ \hline
   64                                          & 3.36                    \\ \hline
   128                                         & 3.36                    \\ \hline
   256                                         & 3.36                    \\ \hline
   384                                         & 3.37                    \\ \hline
   512                                         & 3.37                   \\ \bottomrule
   \end{tabular}
   \end{table}

This suggests that the proposed method RLEL is less sensitive to the number of quantization levels for higher values. For RLEL, the decoding matrix that converts the encodings to the predicted label is also learned during the training (Figure 3). This matrix is of size $M\times N$, where each column represents the weight parameters for one quantization level. One possible reason for the above results is that matrix $D$ adaptively learns the number of quantization levels suitable for this problem.

There is a potential to adaptively learn the number of quantization levels and non-uniform quantization using the proposed RLEL framework. For example, in Figure 3- step (4), fixed parameters $j$ are used to scale the correlation vector $\hat{C}^j_i$ and find the expected prediction $\hat{y}_i$. These parameters represent quantization levels. One possible approach to learning the quantization levels is to make these parameters trainable. In this case, L1/L2 loss between the expected prediction $\hat{y}_i$ and target labels $y_i$ can be used to train the network.
 }
\subsubsection{Impact of dataset size on error for RLEL and BEL}
\label{sec:datasetsize}
{
 %  \color{blue}
In order to compare the effect of dataset size on encoding design, we run BEL and RLEL approaches with the same training loss function (cross entropy loss in~\eqref{eq:loss}). We take the dataset FLD1 and use a fraction of the dataset for training. The entire test dataset is used for testing here. Table~\ref{tab:datasetsize} summarizes the error achieved by RLEL and BEL for different fractions of the training dataset. The evaluation shows that the gap between the performance of RLEL and BEL decreases with the increase in dataset size, which suggests that RLEL might be able to achieve lower error for larger datasets.

\begin{table}[t]
   \centering
   \caption{Effect of dataset size on the error for FLD1 benchmark.}
   \label{tab:datasetsize}
   \begin{tabular}{lrrr}
   \toprule
   \multicolumn{1}{l}{\%Dataset used} & \multicolumn{1}{l}{RLEL} & \multicolumn{1}{l}{BEL} & \multicolumn{1}{l}{Difference (RLEL-BEL)} \\ \midrule
   100                               & 3.36                     & 3.35                    & 0.01                                      \\ \hline
   80                                & 3.43                     & 3.42                    & 0.01                                      \\ \hline
   60                                & 3.53                     & 3.47                    & 0.06                                      \\ \hline
   40                                & 3.77                     & 3.72                    & 0.05                                      \\ \hline
   20                                & 4.08                     & 4.04                    & 0.04                                      \\ \hline
   10                                & 4.71                     & 4.63                    & 0.08                    \\ \bottomrule                 
   \end{tabular}
   \end{table}
}
\subsubsection{Comparison of learned and manually designed encodings}
\begin{figure}[h]
   \centering
   \begin{subfigure}[t]{0.49\linewidth}
       \centering
       \includegraphics[width=\textwidth]{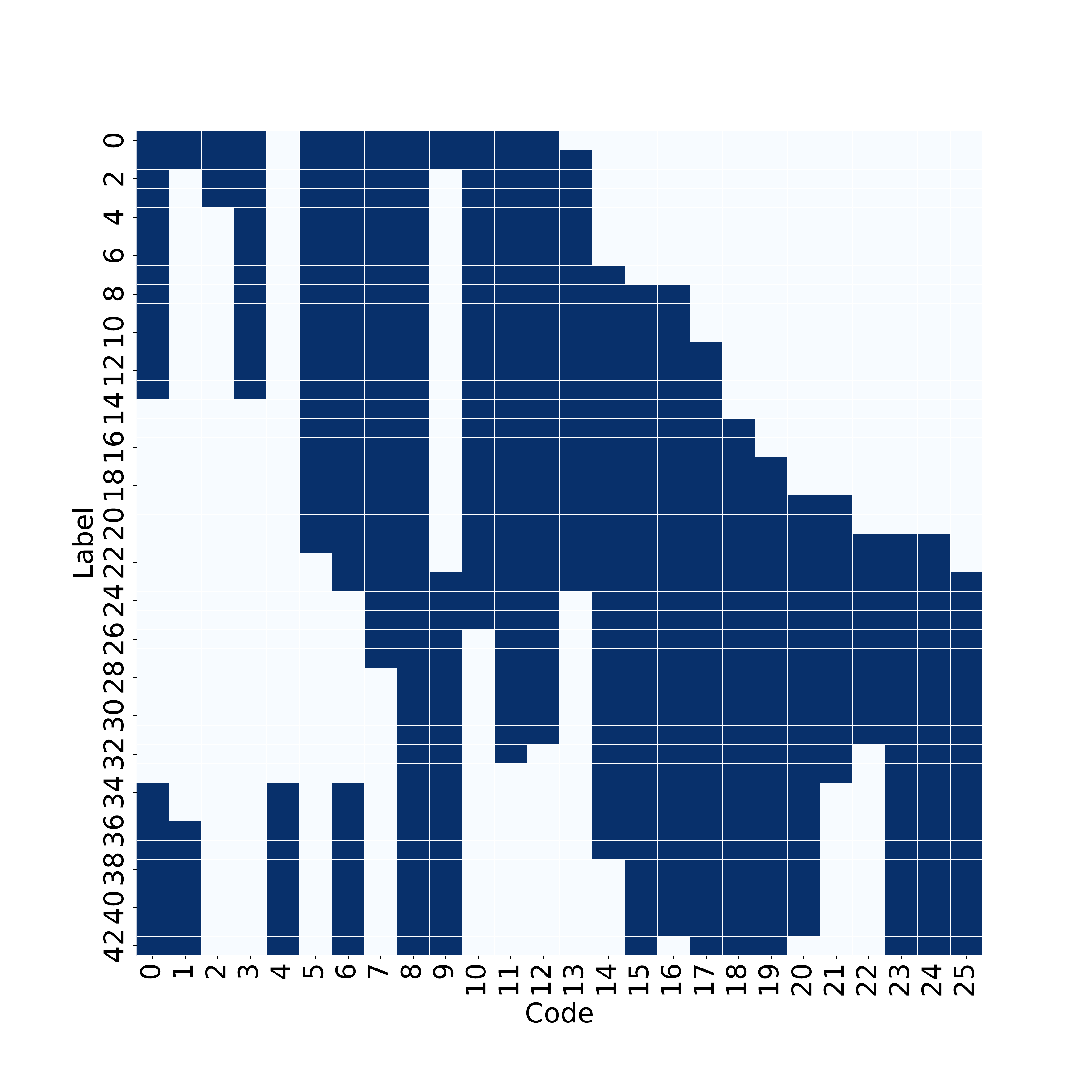}
       \caption{Learned encoding (RLEL)}
       \label{fig:learned}
   \end{subfigure}
   \begin{subfigure}[t]{0.49\textwidth}
     \centering
     \includegraphics[width=\textwidth]{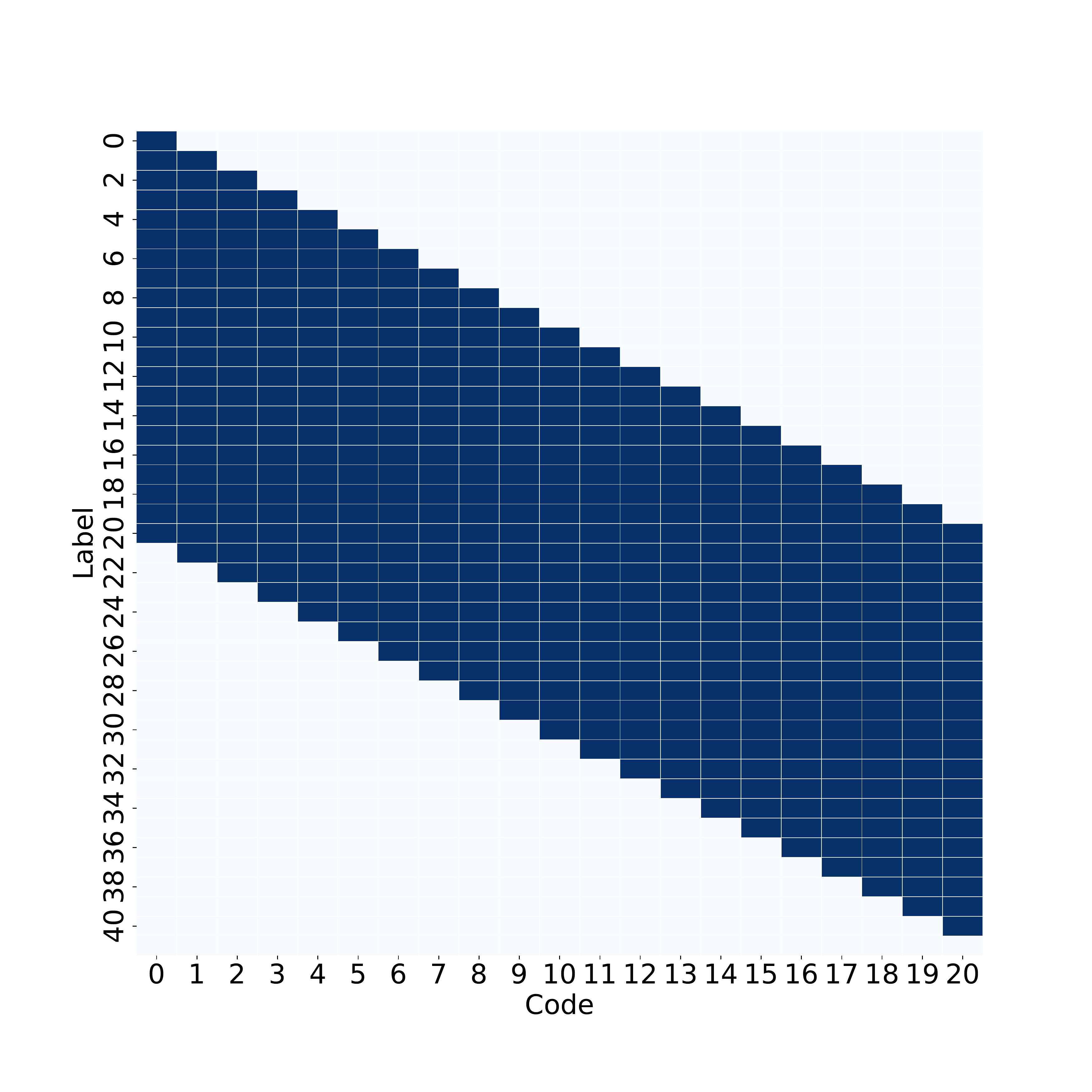}
     \caption{Hand-crafted encoding (BEL)}
     \label{fig:hand}
   \end{subfigure}
   \caption{(a) and (b) give examples of learned and hand-crafted encodings. Here, row $k$ represents an encoding for label $k$. Column $j$ represents the bit  values for classifier-k over the numeric range of labels.  }
 \label{fig:encoding_compare}
 \end{figure}
 {
  % \color{blue}
 We visually compare the encoding learned by RLEL with BEL manually designed code for one benchmark. Figure~\ref{fig:encoding_compare} shows the learned and manually designed encodings. Here, row $k$ represents an encoding for label $k$. Column $j$ represents the bit  values for classifier-k over the numeric range of labels. We notice some common characteristics between both encodings. For example, the codes for nearby labels differ by fewer bits than faraway labels. Both the codes also have fewer bit transitions ($0\rightarrow1$ and $1\rightarrow0$ transitions in a column). These characteristics in the learned encodings are encouraged by the proposed regularizers R1 and R2. There are a few differences between learned and hand-crafted encodings. In contrast to hand-crafted labels, encodings for adjacent labels do not differ in some cases, where hand-crafted encoding assures at least one or two bits of difference between adjacent labels.
 }

\subsection{Label Encoding Design}
\label{sec:a2}
We evaluate different label encoding design approaches, including simulated annealing and autoencoder. These approaches have been used to design encodings for multiclass classification by prior works~\citep{Song2021,extremecode}.
We adapt these approaches to design encodings for regression tasks and compare \CLL{ } with these code design techniques. This section provides the methodology for simulated annealing and autoencoder-based label encoding design.
\subsubsection{Simulated Annealing}
Simulated annealing is a probabilistic approach to find a global optimum of a given function.
It is often used for combinatorial optimization, where the search space is discrete.
\begin{algorithm}
   \caption{Simulated annealing for encodings design}\label{alg:des}
   \textbf{Input:} K$_{\text{max}}$, T, $M$, $N$;\\
   \textbf{Output:} C $\in \{0,1\}^{M \times N}$;\\
   \begin{algorithmic}[1]
   \State  C = C$0$ $\in \{0,1\}^{N\times M}$, where $\text{Pr}(\text{C}0_{i,j}=0) = \text{Pr}(\text{C}0_{i,j}=1)$ \label{al1}
   \State t = T \label{al2}
   \For{\text{k} $\in$ K$_{\text{max}}$}\label{alg1:ms}
       \State C$_\text{new}$ = Move(C) \label{almove}
       \State D = E(C$_\text{new}$) - E(C) \label{alen}
       \If{D < 0 or  e$^{\frac{-D}{t}}$ > Random(0,1) } \label{alif}
           \State C = C$_\text{new}$
       \EndIf \label{alifend}
       \State t = T / (k + 1) \label{altemp}
   \EndFor
   \end{algorithmic}
 \end{algorithm}
Algorithm~\ref{alg:des} represents the pseudo-code for label encoding design using simulated annealing. This algorithm takes two hyperparameters, K$_{\text{max}}$ (number of iterations) and T (initial temperature). It designs a code matrix {\tt C} of size $N \times M$, where $N$ is the number of values and $M$ is the number of bits. Each row $k$ in this code matrix represents encoding for value $k$.
Code matrix {\tt C} is initialized with a random matrix of $0$ and $1$ (Line~\ref{al1}).
 
For each iteration, a new code matrix {\tt C$_\text{new}$} is sampled from the current code matrix {\tt C} using a {\tt Move} function (Line~\ref{almove}). For example, a move function can be designed to randomly flip a few bits in {\tt C}.
The difference between \emph{the errors} of the current and new code matrix is measured (Line~\ref{alen}).
The error of a code matrix, i.e., expected regression error for this problem, is measured using function {\tt E}.
For example, {\tt E} can be replaced by training a regression network for a given code matrix to measure the regression error.
Finally, the current code matrix {\tt C} is updated with the new matrix {\tt C$_\text{new}$} probabilistically. The probability is determined using the decrease in regression error and current temperature {\tt t} (Line~\ref{alif}-\ref{alifend}).
The current temperature is updated for each iteration (Line~\ref{altemp}).
 
There are mainly two design parameters in the above algorithm: the error measurement function {\tt E} and the move function {\tt Move}. We further explain the design of these functions.
 
\subsubsubsection{\textbf{Error measurement: }}
 
We used the expected absolute error between targets and decoded predictions for a given code matrix as its error, as the goal is to design a code matrix that results in lowest regression error.
However, training a regression network for each sample code matrix to measure its regression error is computationally expensive and time-consuming ($\sim 200$ training runs).
Hence we use an analytical model to estimate the regression error for a given code matrix.
 
Regression error is the absolute error between targets $Q_i$ and decoded predictions $\hat{Q}_i$. For a given target $Q_i$ and target code $B_i = C_{Q_i,:}$, the predicted code ($\hat{B_i}$) will be erroneous due to classification errors. This erroneous predicted code is decoded to a predicted value ($\hat{Q_i}$). The following equation is used to predict $\hat{Q_i}$ in expected-correlation-based decoding~\citep{ShahICLR2022}.
\begin{equation}
\label{eq:genex}
   \mathcal{D}^{\text{GEN-EX}}(\hat{B}_i,C) = \sum_{k=1}^N k \sigma_k, \text{where } \sigma_k = \frac{e^{\hat{B}_i \cdot C_{k, :}}}{\sum_{j=1}^N e^{\hat{B}_i \cdot C_{j, :}} }
 \end{equation}
 
The regression error can be estimated given sufficient samples of $B_i$ and $\hat{B}_i$.

\citet{ShahICLR2022} provided an approximate model of classification errors. They showed that for each classifier, its error probability distribution can be approximated using a combination of $p$ Gaussian distributions, where $p$ is the number of bit transitions. Each Gaussian distribution is centered around a bit transition. For example, for bit-$k$ in unary code with bit transition between $Q = k$ and  $Q = k+1$, the error probability of the classifier-$k$ for different target labels $Q_i$ can be approximated as:
\begin{equation}
   \label{eq:temperror}
  e_{k}(Q_i) = r  f_{\mathcal{N}(\mu_{k},\sigma^2)}(Q_i) , \text{where, }\mu_{k} =  k + 0.5
  \end{equation}
 
$\hat{B}_i$ can be sampled for the given $Q_i$ and $C$ using the above error-probability model. Equation~\ref{eq:genex} is then used to find the decoded prediction $\hat{Q_i}$. We measure the expected absolute error between $\hat{Q}_i$ and $Q_i$ using $100\times N$ samples.
 
\begin{figure}[t]
   \centering
       \includegraphics[width=0.7\textwidth]{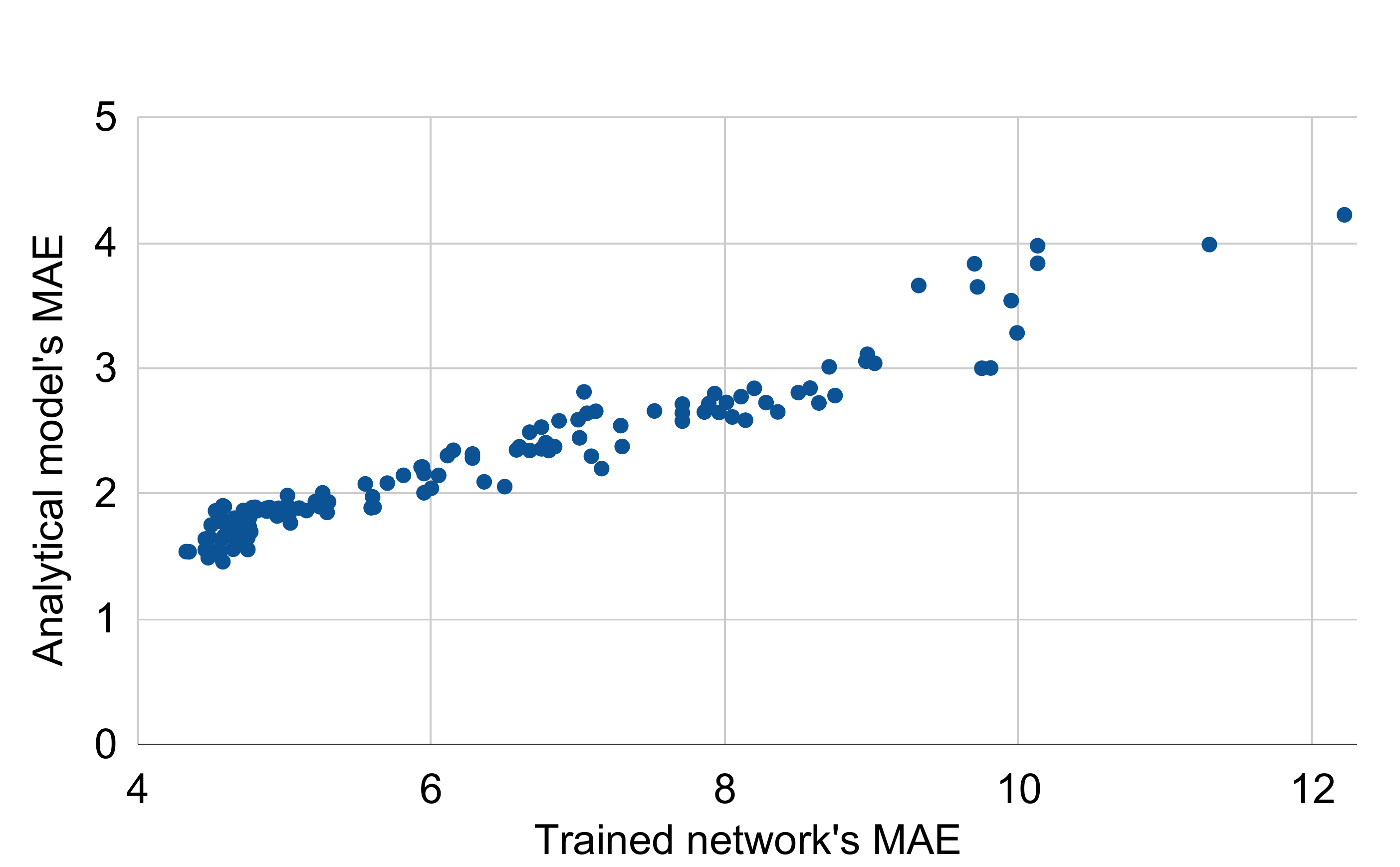}
  \caption{Comparison of Mean Absolute Error (MAE) approximated by the proposed analytical model and trained network for different code matrices.Each point in this plot is a different code matrix.  }
  % (b) Examples of two lookup table-based encoding and decoding functions. $\hat{Q}_i$ represents the decoded quantized level.  }
  \label{fig:simcorr}
 \end{figure}
We further verify the validity of this analytical model by finding the correlation between regression error measured by this model and trained networks. Figure~\ref{fig:simcorr} plots the analytical regression error versus actual regression error for FLD\_1 benchmarks. Here, each point is for a different code matrix. The $Y$-axis represents the absolute error approximated by the proposed analytical model. The $X$-axis represents the absolute test error of a trained network for a given code matrix. The figure shows that the proposed analytical model for error measurement approximates error with significant speedup.
 
\subsubsubsection{\textbf{Move function: }}
The move function flips some bits in the current code matrix to sample a new one. A naive approach would be to randomly flip $b$ bits. 
We further optimize the move function to consider the regression task objective. For a given code matrix, using the proposed analytical model, we find a matrix $F$ of size $N \times N$, where $F_{i,j}$ = $|i-j| \times Pr(\text{Round}(\mathcal{D}(\hat{B},C))= j | Q=i,B = C_{i,:})$. Thus, each element represents a pair ($C_{i,:},C_{j,:}$) of encodings' contribution to expected error. We select top-$b$ pairs from this matrix. For each pair of encodings, we find bit-positions that have equal bit-values between two encodings, and a randomly selected bit-position from this list is flipped in encoding $C_{i,:}$. This procedure increases the hamming distance between pairs of encodings that contribute the highest to the regression error.
Figure~\ref{fig:movecomp} compares the convergence of the proposed move function and a random-flip-based move function. Here the $Y$-axis represents the approximated error for the current code matrix, and $X$-axis represents the iteration identifier. The figure shows that the proposed error-based move function results in faster convergence and lower error.
 
\begin{figure}[t]
   \centering
   \includegraphics[width=0.7\textwidth]{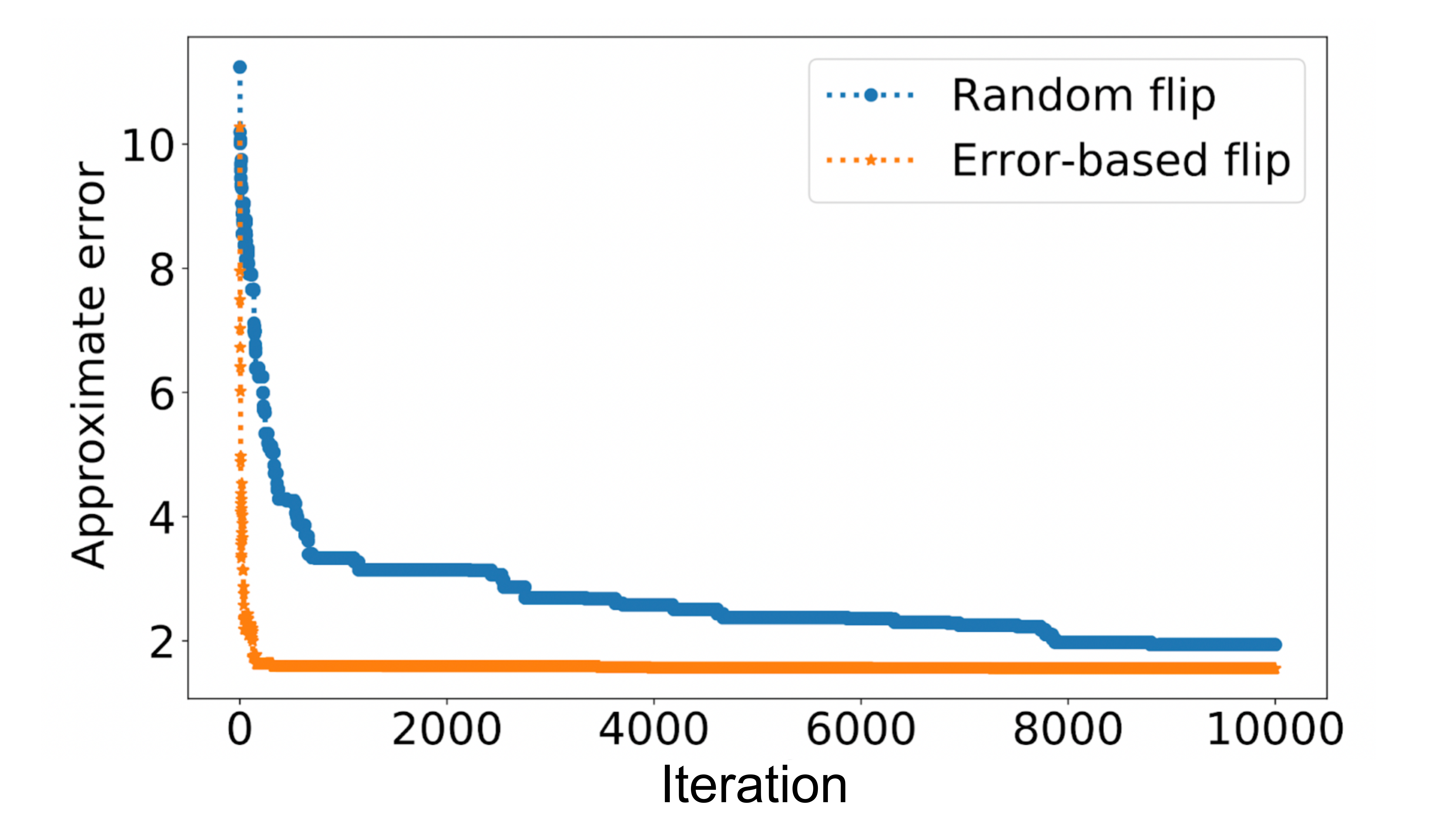}
  \caption{Comparison of convergence of random-flip and proposed error-based flip move functions.   }
  % (b) Examples of two lookup table-based encoding and decoding functions. $\hat{Q}_i$ represents the decoded quantized level.  }
  \label{fig:movecomp}
 \end{figure}
 
 We use the proposed move function with the analytical model to approximate regression error in Algorithm~\ref{alg:des} to design label encoding for regression using simulated annealing.
\subsubsection{Autoencoder}
\begin{figure}[t]
    \centering
        \includegraphics[width=0.7\textwidth]{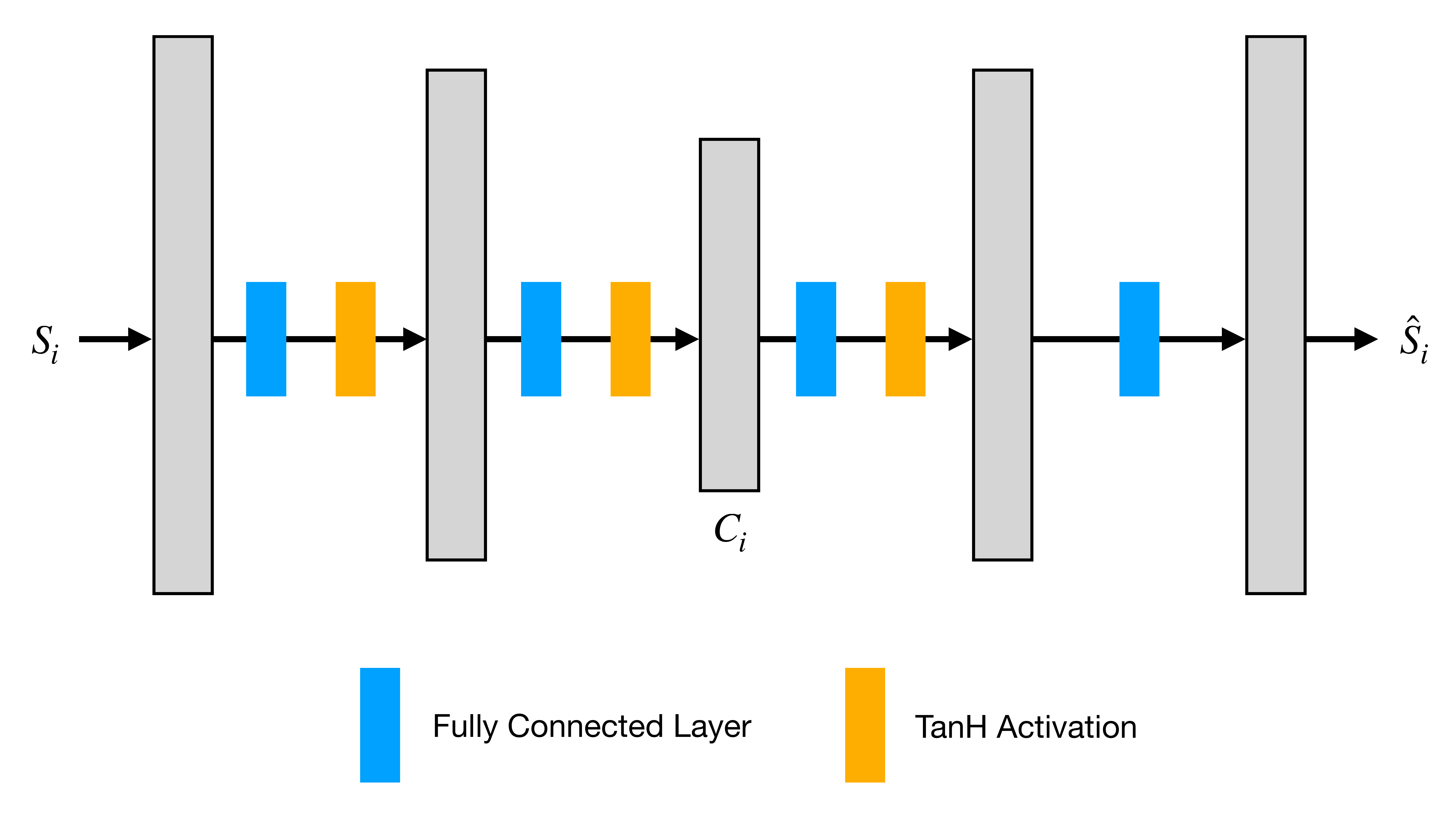}
   \caption{Network architecture for autoencoder-based encodings design. }
   % (b) Examples of two lookup table-based encoding and decoding functions. $\hat{Q}_i$ represents the decoded quantized level.  }
   \label{fig:auto}
  \end{figure}
Ciss{\'e} et al.~\citep{extremecode} proposed an autoencoder-based approach to design encodings for a multiclass classification problem. Figure~\ref{fig:auto} represents the network architecture used for encodings design. Input $S_i$ is an $N$-dimensional vector for class $i$. Here, each element $S_i[j]$ represents the similarity between class $i$ and $j$. The output of the bottleneck layer $C_i$ is the designed encodings for class $i$.
 
For regression problems, we set $S_i[j] = |i-j|$. Let $W$ represent the weight parameters of the network. The network is trained using SGD optimization, where each batch consists of randomly sampled $i$ and $j$. The following loss function is used for training:
\begin{equation}
   \label{eq:autoloss}
   \mathcal{L} = ||\hat{S}_i - S_i||^2 + ||\hat{S}_j - S_j||^2 + \beta \text{max}(0,b-||C_i - C_j||_1) + \gamma ||W||^2
\end{equation}
Here, the first and second terms represent reconstruction losses for inputs $S_i$ and $S_j$. The third term encourages a minimum distance of $b$ between any pair of encodings to yield unique encodings for different classes. The fourth term is an L2-regularizer.
 
Once the network is trained, the real-valued encodings $C_i$ are converted to binary encodings such that it has equal numbers of $0$s and $1$s.
This formulation introduces three hyperparameters.
We determine the number of bit transitions in the designed label encodings and select hyperparameters that result in the lowest number of bit transitions. 

Note that this autoencoder network is decoupled from the regression network and design codes agnostic to classifiers' characteristics for a given regression problem. 
%%\subsubsection{Visual Comparison of Label encodings}
\subsection{Experimental Methodology}
\label{sec:a3}
We use $11$ benchmarks covering four different regression tasks for evaluation. This section summarizes the experimental setup, including datasets, evaluation metrics, hyperparameters, and related work for each of these tasks.
We also report the training time using an Nvidia RTX 2080 Ti GPU with 11GB of memory for each benchmark.
\subsubsection{Head Pose Estimation}
\label{sec:hpe}
In landmark-free 2D head pose estimation, for a given 2D image, the head pose of a human is directly estimated in terms of three angles: yaw, pitch, and roll.
We use loose cropping around the center with random flipping for data augmentation.
We use the ResNet50 network as the feature extractor.
This network is initialized using pre-trained parameters for ImageNet~\citep{imagenet15} dataset. During the training for \CLL{ }the entire network, including the feature extractor, is trained.
\paragraph{Datasets:}
We use the evaluation methodology followed by prior works~\citep{hopenet,fsanet}. Two protocols are used for evaluation.

\underline{Protocol 1 (LFH1):}
This protocol uses the BIWI~\citep{biwi} dataset for training and evaluation. This dataset consists of $15,128$ frames of $20$ subjects. Random $70\%-30\%$ splits are used for training and evaluation. The ranges of yaw, pitch, and roll angles are $[-75^{\circ},75^{\circ}]$, $[-65^{\circ},85^{\circ}]$, and $[-55^{\circ},45^{\circ}]$, respectively.

\underline{Protocol 2 (LFH2):}
In this protocol, the network is trained using the 300W-LP~\citep{300wlp} dataset consisting of $122,450$ samples. AFLW2000~\citep{300wlp} dataset is used for evaluation. The range of all labels is  $[-99^{\circ},99^{\circ}]$ in this setting.
\paragraph{Evaluation metrics:}
We report the Mean Absolute Error (MAE) between the targets ($y_i$) and predictions ($\hat{y}_i$). Let $N$ represent the number of samples, and $P$ represent the number of labels (three in head pose estimation). The MAE is defined as:
\begin{equation}
  \label{eq:mae}
  \text{MAE} = \frac{1}{N} \sum_{i=1}^{N} \frac{1}{P} \sum_{j=1}^{P} |y_{i,j} - \hat{y}_{i,j}|
\end{equation}
\paragraph{Network architecture and training parameters: }
Table~\ref{tab:trp1} summarizes the hyperparameters used for \CLL{ }. The learning rate of the decoding matrix $D$ is kept $10\times$ higher than the learning rate of the feature extractor. L2 regularization with weight of 0.0001 is used for direct regression. 
\begin{table}[h]
\begin{center}
    \setlength\tabcolsep{3pt}
    \caption{Training parameters for LFH1. }
    \label{tab:trp1}
    \scriptsize
    \begin{tabular}{L{1cm}L{2.7cm}C{1.7cm}cC{0.7cm}C{1cm}C{2cm}C{0.5cm}C{0.5cm}C{1cm}}
    \toprule
    Approach &  Label range/ Quantization levels & Optimizer & Epochs & Batch size &  Learning rate & Learning rate schedule  & $\beta$ & $\alpha$ & Training time (GPU hours) \\  \midrule
    LFH1 & Yaw: $[-75^{\circ},75^{\circ}]/150$, Pitch:$[-65^{\circ},85^{\circ}]/150$ , Roll: $[-55^{\circ},45^{\circ}]/100$ & Adam, weight decay=0, momentum = 0 & 50 & 8 & 0.0001 & 1/10 after 30 Epochs & 0.5 & 1.0 & 2 \\ \hline
    LFH2 & $[-99^{\circ},99^{\circ}]/200$ &Adam, weight decay=0, momentum = 0 & 20 & 16 & 0.00001 & 1/10 after 10 Epochs & 2.0 & 0.0 & 4 \\ \hline
    \bottomrule
    \end{tabular}
\end{center}
\end{table}
\paragraph{Related work}
Head pose estimation is a widely studied problem.
Existing task-specialized approaches propose different loss formulations or feature extractors to improve the error.
HopeNet~\citep{hopenet} proposed a combination of regression and classification loss.
SSR-Net~\citep{ssrnet} and FSA-Net ~\citep{fsanet} proposed stage-wise soft regression.
QuatNet~\citep{quatnet} proposed to use MSE loss with custom ordinal regression loss.
RAFA-Net~\citep{rafanet} proposed an attention-based feature extractor architecture.
Table~\ref{tab:hpe2} and Table~\ref{tab:hpe1} compare the performance of \CLL{ }with related work.
\begin{table}[h]
  \begin{center}
  \setlength\tabcolsep{6pt}
  \caption{Landmark-free 2D Head poses estimation evaluation for protocol 1 (HPE1 and HPE3). }
  \label{tab:hpe2}
  \scriptsize
  \begin{tabular}{L{3.2cm}C{1.4cm}C{1.1cm}llll@{}}
    \toprule
    Approach & Feature Extractor  & \#Params (M) &  Yaw & Pitch & Roll & MAE  \\ \midrule
  SSR-Net-MD~\citep{ssrnet} { }{ }  (Soft regression) & SSR-Net & 1.1  & 4.24  & 4.35 & 4.19 & 4.26 \\ \hline
  FSA-Caps-Fusion~\citep{fsanet} (Soft regression)  & FSA-Net & 5.1 & \underline{2.89} & 4.29 & 3.60 & 3.60    \\  \hline 
  RAFA-Net~\citep{rafanet} (Direct Regression)  & RAFA-Net & 69.8 & \underline{3.07} & 4.30 & \underline{2.82} & \underline{3.40} \\ \hline
  Direct regression (L2 loss) & ResNet50  & 23.5  &  3.80  & 4.63  & 4.28q & 4.22 $\pm$ 0.35  \\ \hline
  BEL~\citep{ShahICLR2022}  & ResNet50 & 23.6  & \textbf{3.32} & {3.80}  & \textbf{3.53}  & {3.56} $\pm$ 0.11\\ \hline
  \CLL  & ResNet50  & 23.6  &  3.41  & \textbf{\underline{3.20}}  &  3.97  & \textbf{3.55} $\pm$ 0.10\\
  \bottomrule
  \end{tabular}
\end{center}
\end{table}
\begin{table}[h]
  \scriptsize
  \begin{center}
  \setlength\tabcolsep{6pt}
  \caption{Landmark-free 2D Head poses estimation evaluation for protocol 2 (HPE2 and HPE4). }
  \label{tab:hpe1}
  \scriptsize
  \begin{tabular}{L{3.2cm}C{1.4cm}C{1.1cm}llll@{}}
    \toprule
    Approach & Feature Extractor  & \#Params (M) &  Yaw & Pitch & Roll & MAE  \\ \midrule
  SSR-Net-MD~\cite{ssrnet} { }{ }  (Soft regression) & SSR-Net & 1.1 & 5.14 & 7.09 & 5.89 & 6.01 \\ \hline
  FSA-Caps-Fusion~\cite{fsanet} (Soft regression)  & FSA-Net & 5.1 & 4.50 & 6.08 & 4.64 & 5.07  \\   \hline
  RAFA-Net~\cite{rafanet} (Direct Regression)  & RAFA-Net (HPE4) & 69.8 & 3.60 & {{4.92}} & 3.88 & \underline{4.13} \\ \hline 
  HopeNet* ($\alpha$ = 2)~\cite{hopenet} (classification + regression loss) &  ResNet50 &  23.9 & 6.47 & 6.56 & 5.44 & 6.16 \\ \hline \hline
  Direct regression (L2 loss) & ResNet50  & 23.5  &  5.61 & 6.13 & 4.18  & 5.32 $\pm$ 0.12   \\ \hline
  BEL~\cite{ShahICLR2022}  & ResNet50  & 23.6 & \textbf{4.54} & \textbf{5.76} & {3.96}  & \textbf{4.77} $\pm$ 0.05  \\ \hline
  \CLL  & ResNet50  & 23.6 & 4.69  &  5.79  & \textbf{3.86}  & \textbf{4.77} $\pm$ 0.05  \\
 \bottomrule
  \end{tabular}\\
\end{center}
  \end{table}
\subsubsection{Facial Landmark Detection}
\label{sec:fld}
Facial landmark detection focuses on finding the $(x,y)$ coordinates of facial keypoints for a given 2D image.
\paragraph{Evaluation metrics:}
We report the Normalized Mean Error (NME) between the targets $y_i$ and predictions $\hat{y}_i$. Inter-ocular distance normalization is used for all datasets. For $N$ test samples, $P$ facial landmarks, and $L$ normalization factor, the NME is defined as:
\begin{equation}
\label{eq:nme}
\text{NME} = \frac{1}{N} \sum_{i=1}^{N} \frac{1}{P} \cdot \frac{1}{L} \sum_{j=1}^{P} |y_{i,j} - \hat{y}_{i,j}|_2
\end{equation}
\paragraph{Datasets: }
We use three datasets widely used for facial landmark detection: COFW~\citep{cofw}, 300W~\citep{300w}, and WFLW~\citep{lab}. HRNetV2-W18 network architecture for feature extraction~\citep{hrnetface} and the modified regressor architecture for label encoding proposed by BEL~\citep{ShahICLR2022} are used in this work.
Random flipping, scaling ($0.75-1.25$), and rotation ($\pm 30$) are used for data augmentation.
The COFW dataset consists of $1,345$ training and $507$ testing images annotated with $29$ landmarks.
The training set of the 300W dataset has $3,148$ images annotated with $68$ facial landmarks. This dataset provides four test sets:  full test set, common subset, challenging subset, and the official test set with $300$ indoor and $300$ outdoor images. WFLW dataset is a comparatively larger dataset with $7,500$ training and $2,500$ testing images. Each image is annotated with $98$ facial landmarks. The test set is divided into six subsets: large pose, expression, illumination, make-up, occlusion, and blur.
\paragraph{Training parameters: }
Table~\ref{tab:trfdl} provides a summary of all the training parameters.
The learning rate of the decoding matrix $D$ is kept $20\times$ higher than the learning rate of the feature extractor. The HRNetV2-W18 network is initialized with pretrained weight parameters for the ImageNet dataset. 
We refer to HRNetV2-W18 evaluated on COFW as \textbf{FLD1/FLD1\_s}, on 300W as \textbf{FLD2/FLD2\_s}, and on WFLW as \textbf{FLD3/FLD3\_s}.
\begin{table}[h]
\begin{center}
\setlength\tabcolsep{6pt}
\caption{Training parameters for facial landmark detection for HRNetV2-W18 feature extractor.}
\label{tab:trfdl}
\scriptsize
\begin{tabular}{C{1.5cm}C{1.5cm}C{0.5cm}C{0.5cm}C{1.5cm}C{2cm}C{0.5cm}C{0.5cm}C{1.5cm}}
  \toprule
Dataset/ Benchmark & Optimizer & Epochs & Batch size &  Learning rate (BEL/Direct regression/Multiclass classification) & Learning rate schedule  &$\beta$&$\alpha$& Training time (GPU hours) \\  \midrule
COFW/ FLD1  & Adam, weight decay=0, momentum = 0 & 60 & 8 & 0.0005/0.0003/ 0.0003 & 1/10 after 30 and 50 Epochs & 3.0 & 0.0& $\frac{1}{2}$ \\ \hline
COFW/ FLD1\_s  & Adam, weight decay=0, momentum = 0 & 60 & 8 & 0.0005/0.0003/ 0.0003 & 1/10 after 30 and 50 Epochs & 4.0 & 0.0& $\frac{1}{2}$ \\ \hline
300W/ FLD2 & Adam, weight decay=0, momentum = 0 & 60 & 8 & 0.0007/0.0003/ 0.0003 & 1/10 after 30 and 50 Epochs & 5.0&1.0 & 3  \\ \hline
300W/ FLD2\_s & Adam, weight decay=0, momentum = 0 & 60 & 8 & 0.0007/0.0003/ 0.0003 & 1/10 after 30 and 50 Epochs &5.0 &0.05 & 3  \\ \hline
WFLW/ FLD3 & Adam, weight decay=0, momentum = 0 & 60 & 8 & 0.0003/0.0003/ 0.0003 & 1/10 after 30 and 50 Epochs & 0.0& 0.1& 5 \\ \hline
WFLW/ FLD3\_s & Adam, weight decay=0, momentum = 0 & 60 & 8 & 0.0003/0.0003/ 0.0003 & 1/10 after 30 and 50 Epochs & 5.0&0.1 & 5 \\
\bottomrule
\end{tabular}
\end{center}
\end{table}
\paragraph{Related work}
Facial landmark detection is a widely studied problem.
A common approach is to use heatmap regression, where the target heatmaps are generated by assuming a Gaussian distribution around the ground truth landmark location.
Prior works proposed the use of binary heatmaps with pixel-wise binary cross-entropy loss~\citep{binaryheatmap}.
HRNet~\citep{hrnetface} proposed a feature extractor that maintains high-resolution representations and uses heatmap regression.
AWing~\citep{awing} proposed a modified heatmap regression loss function with adapted wing loss.
AnchorFace ~\citep{anchorface} used anchoring of facial landmarks on templates.
%Common regression approaches for this tasks includes regression using MSE loss~\cite{ssdm,8099876}, cascaded regression~\cite{dsrn,7298989,ccl,fpd}, and coarse-to-fine regression~\cite{fpd,cfss,cfan}. State-of-the-art methods for this task learn heatmaps by regression to find facial landmarks. %~\cite{hrnetface,san,dan,lab}.
%SAN~\cite{san} augments training data using temporal information and GAN-generated faces.
%DVLN~\cite{dvln}, CFSS~\cite{cfss}, LAB~\cite{lab}, DSRN~\cite{dsrn} take advantage of correlations between facial landmarks. % to improve performance.
%DAN~\cite{dan} introduces a progressive refinement approach using predicted landmark heatmaps. % passed across stages of a neural network.
%LAB~\cite{lab} also exploits extra boundary information to improve the accuracy.
LUVLi~\citep{luvli} proposed a landmark's location, uncertainty, and visibility likelihood-based loss.
% SAN~\cite{san} augments training data using temporal information and GAN-generated faces, respectively. DVLN~\cite{dvln}, CFSS~\cite{cfss}, LAB~\cite{lab}, DSRN~\cite{dsrn} take advantage of correlations between facial landmarks to improve performance. DAN~\cite{dan} introduces a progressive refinement approach using predicted landmark heatmaps passed across stages of a neural network. HRNet~\cite{hrnetface} proposes a new CNN architecture to maintain high-resolution representations across the network.
Table~\ref{tab:fldcofw}-~\ref{tab:fldwflw} compare \CLL{ }with related work.
\begin{table}[h]
\begin{center}
\setlength\tabcolsep{6pt}
\caption{Facial landmark detection results on COFW dataset (FLD1). }
\label{tab:fldcofw}
\scriptsize
\begin{tabular}{@{}L{4.1cm}cC{1.5cm}ll}
\toprule
Approach &  Feature Extractor & \#Params/ GFlops & Test NME & FR$_{0.1}$   \\  \midrule
LAB (w B)~\citep{lab}         & Hourglass   & 25.1/19.1 & 3.92 & 0.39   \\ \hline
AWing~\citep{awing}*         & Hourglass & 25.1/19.1 & 4.94 & -  \\ \hline
\hline
HRNetV2-W18~\citep{hrnetface} (Heatmap regression) & HRNetV2-W18 & 9.6/4.6 & 3.45 &  0.19 \\   \hline
Direct regression (L2 loss) & HRNetV2-W18 & 10.2/4.7  &  3.96 $\pm$ 0.02 & 0.29  \\ \hline
Direct regression (L1 loss) & HRNetV2-W18 & 10.2/4.7  &  3.60 $\pm$ 0.02 & 0.29 \\ \hline
BEL~\citep{ShahICLR2022}          & HRNetV2-W18 & 10.6/4.6  &   \textbf{\underline{3.34}} $\pm$ 0.02  & 0.40 \\ \hline
\CLL          & HRNetV2-W18 & 10.6/4.6  &   {{3.36}} $\pm$ 0.01  & 0.20 \\ \bottomrule
\end{tabular}
\scriptsize{\\$*$Uses different data augmentation for the training}
\end{center}
\end{table}
\begin{table}[h]
\begin{center}
\setlength\tabcolsep{3.4pt}
\caption{Facial landmark detection results on 300W dataset (FLD2). }
\label{tab:fld300W}
\scriptsize
\begin{tabular}{@{}L{3.1cm}cC{1.3cm}llll}
\toprule
Approach &  Feature Extractor & \#Params/ GFlops & Test & Common & Challenging & Full  \\  \midrule
DAN~\citep{dan}               & -   & - & -    &  3.19 & 5.24 & 3.59  \\ \hline
LAB (w B)~\citep{lab}         & Hourglass   & 25.1/19.1 & - & 2.98 & 5.19 & 3.49   \\ \hline
AnchorFace~\citep{anchorface} & ShuffleNet-V2   & - & - & 3.12 & 6.19 & 3.72   \\ \hline
AWing~\citep{awing}*         & Hourglass & 25.1/19.1 & - & \underline{2.72}& \underline{4.52} & \underline{3.07}  \\ \hline
LUVLi~\citep{luvli}   & CU-Net & - &  - & 2.76 & 5.16 & 3.23   \\ \hline
\hline
HRNetV2-W18~\citep{hrnetface} (Heatmap regression) & HRNetV2-W18 & 9.6/4.6 &- & \textbf{2.87} & \textbf{5.15} & \textbf{3.32}  \\   \hline
Direct regression (L2 loss) & HRNetV2-W18 & 10.2/4.7  & 4.40 & 3.25 & 5.65 & 3.71 $\pm$ 0.05  \\ \hline
Direct regression (L1 loss) & HRNetV2-W18 & 10.2/4.7  & 4.26 & 3.10 & 5.42 & 3.54 $\pm$ 0.03 \\ \hline
BEL~\citep{ShahICLR2022}             & HRNetV2-W18 & 11.2/4.6  & {{4.09}} & {{2.91}} &  5.50  &  {3.40} $\pm$ 0.02   \\ \hline
\CLL             & HRNetV2-W18 & 11.2/4.6  & \textbf{\underline{4.03}} & {{2.90}} &  5.39  &  {3.37} $\pm$ 0.02   \\
\bottomrule
\end{tabular}
\scriptsize{\\$*$Uses different data augmentation for the training }
\end{center}
\end{table}
\begin{table}[h]
\begin{center}
\setlength\tabcolsep{2.5pt}
\caption{Facial landmark detection results (NME) on WFLW test (FLD3) and 6 subsets: pose, expression (expr.), illumination (illu.), make-up (mu.), occlusion (occu.) and blur. $\theta=10$ is used for \BEL.  }
\label{tab:fldwflw}
\scriptsize
\begin{tabular}{@{}L{2.6cm}cC{1.3cm}L{1.6cm}cccccc}
\toprule
Approach &  Feature Extractor & \#Params/ GFlops & Test & Pose & Expr. & Illu. & MU & Occu. &Blur  \\  \midrule
LAB (w B)~\citep{lab} & Hourglass   & 25.1/19.1 & 5.27 & 10.24 & 5.51 & 5.23 & 5.15 & 6.79 & 6.32    \\ \hline
AnchorFace~\citep{anchorface}* & HRNetV2-W18   & -/5.3 & \underline{4.32} & 7.51&4.69 &\underline{4.20} &{4.11} &\underline{4.98} & \underline{4.82} \\ \hline
AWing~\citep{awing}*         & Hourglass & 25.1/19.1 & 4.36 & \underline{7.38} & \underline{4.58}& 4.32& 4.27& 5.19& 4.96    \\ \hline
LUVLi~\citep{luvli}   & CU-Net & - &  4.37  &- &- &-  &-  &-  &- \\ \hline
\hline
HRNetV2-W18~\citep{hrnetface} (Heatmap regression) & HRNetV2-W18 & 9.6/4.6 & 4.60 & 7.94 &4.85 &4.55 &4.29 &5.44 & 5.42 \\   \hline
Direct regression (L1 loss) & HRNetV2-W18 & 10.2/4.7  & 4.64 $\pm$ 0.03  & 8.13& 4.96& 4.49& 4.45& 5.41&5.25\\ \hline
BEL~\citep{ShahICLR2022}        & HRNetV2-W18 & 11.7/4.6 &  {{4.36}} $\pm$ 0.02 &\textbf{{7.53}} &{4.64} &\textbf{4.28} & \textbf{{4.19}} &\textbf{5.19} &\textbf{5.05}\\  \hline
\CLL        & HRNetV2-W18 & 11.7/4.6 &  {\textbf{4.35}} $\pm$ 0.01 &{{7.57}} &\textbf{4.57} &{4.36} & \textbf{{4.19}} &{5.25} &{5.07}\\  %%\hline
\bottomrule
\end{tabular}
\scriptsize{\\$*$Uses different data augmentation for the training}
\end{center}
\end{table}
\FloatBarrier
\subsubsection{Age Estimation}
\label{sec:ae}
This task focuses on predicting a person's age from a given 2D image. MAE (\Eqref{eq:mae}) and Cumulative Score (CS)  are used as the evaluation metrics, and ResNet50~\citep{resnet} is used as the feature extractor. CS$\theta$ is the percentage of test samples with absolute error less than $\theta$ years.
\paragraph{Datasets}
MORPH-II~\citep{morphii} and AFAD~\citep{agecnn} datasets are used for evaluation.
We follow the protocols for preprocessing and data augmentation of datasets as per prior works~\citep{ShahICLR2022,mlxtend}.
MORPH-II dataset consists of $55,608$ images with random split of $39,617$ training, $4,398$ validation, and $11,001$ test images. The AFAD dataset consists of $164,432$ images with random split of $118,492$ training, $13,166$ validation, and $32,763$ test images.
\paragraph{Training parameters: }
Table~\ref{tab:trage} summarizes the training parameters for \textbf{AE1} (MORPH-II) and \textbf{AE2} (AFAD) benchmarks.  The learning rate of the decoding matrix $D$ is kept $10\times$ higher than the learning rate of the feature extractor. L2 regularization with weight of 0.001 is used for direct regression. 
Training for AE1 and AE2 consumes $\sim 2$ and $\sim 7$ hours, respectively.
\begin{table}[h]
\begin{center}
  \setlength\tabcolsep{6pt}
  \caption{Training parameters for age estimation using MORPH-II and AFAD dataset }
  \label{tab:trage}
  \scriptsize
  \begin{tabular}{L{1cm}L{2cm}C{1cm}C{1cm}cC{2cm}C{1cm}C{0.5cm}C{0.5cm}}
  \toprule
  Benchmark& Optimizer & Epochs & Batch size &  Learning rate & Learning rate schedule & $\beta$ & $\alpha$ \\  \midrule
  AE1 & Adam, weight decay=0, momentum=0 & 50 & 64 & 0.0001 & - & 0.0 & 2.0
  \\ \hline
  AE2 & Adam, weight decay=0, momentum=0 & 50 & 64 & 0.0001 & - & 0.0 & 5.0
  \\ \bottomrule
  \end{tabular}
\end{center}
\end{table}
\paragraph{Related work}
Different approaches including ordinal regression~\citep{agecnn, coralcnn,mvloss, dldl}, soft stage-wise regression~\citep{ssrnet,fsanet}, soft labels~\citep{softlabel} have been proposed for age estimation.
OR-CNN~\citep{agecnn} and CORAL-CNN~\citep{coralcnn} proposed ordinal regression by binary classification with threshold-based encodings (i.e., unary codes).
DLDL~\citep{dldl} augmented the loss function with KL-divergence between softmax output and soft target probability distributions.
MV-Loss~\citep{mvloss} proposed to penalize the prediction based on the variance of the age distribution.
We compare CLL with related work in Table~\ref{tab:age} and Table~\ref{tab:agee}.
\begin{table}[h]
  {
  \centering
  \setlength\tabcolsep{4pt}
  \caption{Age estimation results on MORPH-II dataset (AE1).}
  \label{tab:age}
  \scriptsize
  \begin{tabular}{L{4.4cm}cR{2cm}L{2cm}L{2cm}}
  \toprule
  Approach & Feature extractor & \#Parameters (M) & MORPH-II (MAE)& MORPH-II (CS$\theta = 5$)\\ \midrule
  OR-CNN~\citep{agecnn} (Ordinal regression by binary classification ) & - & 1.0 & 2.58 & 0.71 \\ \hline
  MV Loss~\citep{mvloss} (Direct regression) & VGG-16 & 138.4 & 2.41 & 0.889 \\ \hline
  DLDL-v2~\citep{dldl} (Ordinal regression with multi-class classification) & ThinAgeNet & 3.7 & \underline{1.96}* & - \\ \hline
  CORAL-CNN~\citep{coralcnn} (Ordinal regression by binary classification) & ResNet34 & 21.3 & 2.49 & - \\ \hline \hline
  Direct Regression (L2 Loss) & ResNet50 & 23.1 & 2.37 $\pm$ 0.01 & 0.903 $\pm$ 0.002 \\ \hline
  BEL~\citep{ShahICLR2022} & ResNet50 & 23.1 & \textbf{{2.27}} $\pm$ 0.01& \textbf{\underline{0.928}} $\pm$ 0.001\\ \hline
  \CLL & ResNet50 & 23.1 & {{2.28}} $\pm$ 0.01& {{0.901}} $\pm$ 0.002\\   \bottomrule
  \end{tabular}
  }
\end{table}
\begin{table}[h]
  {
  \centering
  \setlength\tabcolsep{4pt}
  \caption{Age estimation results on AFAD dataset (AE2).}
  \label{tab:agee}
  \scriptsize
  \begin{tabular}{L{4.4cm}cR{2cm}L{2cm}L{2cm}}
  \toprule
  Approach & Feature extractor & \#Parameters (M) & AFAD (MAE)& AFAD (CS$\theta = 5$)\\ \midrule
  OR-CNN~\citep{agecnn} (Ordinal regression by binary classification ) & - & 1.0 & 3.51 & 0.74 \\ \hline
  CORAL-CNN~\citep{coralcnn} (Ordinal regression by binary classification) & ResNet34 & 21.3 & 3.47 & - \\ \hline \hline
  Direct Regression (L2 Loss) & ResNet50 & 23.1 & 3.16 $\pm$ 0.02 & 0.810 $\pm$ 0.02 \\ \hline
 BEL~\citep{ShahICLR2022} & ResNet50 & 23.1 & \textbf{\underline{3.11}} $\pm$ 0.01& \textbf{\underline{0.823}} $\pm$ 0.001\\ \hline
 \CLL & ResNet50 & 23.1 & {{3.14}} $\pm$ 0.01& {{80.78}} $\pm$ 0.002\\ \bottomrule
  \end{tabular}
  }
  %\vspace{-3mm}
\end{table}
\subsubsection{End-to-end Self Driving}
\label{sec:pn}
For the regression task of end-to-end autonomous driving, we use the NVIDIA PilotNet dataset, and PilotNet model ~\citep{pilotold}. In this task, for a given image of the road, the angle of the steering wheel that should be taken next is predicted. MAE (\Eqref{eq:mae}) is used as the evaluation metric.
\paragraph{Dataset}
The PilotNet driving dataset consists of $45,500$ images taken around Rancho Palos Verdes and San Pedro, California~\citep{sully}. We use the data augmentation technique used by prior works~\citep{pilotold}.
\paragraph{Training parameters}
Table~\ref{tab:tauto} summarizes the training parameters.  The learning rate of the decoding matrix $D$ is kept $10\times$ higher than the learning rate of the feature extractor.
\begin{table}[h]
\begin{center}
  \setlength\tabcolsep{6pt}
  \caption{Training parameters for end-to-end autonomous driving using PilotNet. }
  \label{tab:tauto}
  \scriptsize
  \begin{tabular}{L{3cm}C{1cm}C{1cm}C{2cm}C{2cm}C{0.5cm}C{0.5cm}}
  \toprule
  Optimizer & Epochs & Batch size &  Learning rate & Learning rate schedule & $\beta$ & $\alpha$  \\  \midrule
  SGD with weight decay=1e-5, momentum=0 & 50 & 64 & 0.1 & 1/10 at 10, 30 epochs & 0.0 & 2.0
  \\ \bottomrule
  \end{tabular}
\end{center}
\end{table}
\paragraph{Related work}
End-to-end autonomous driving is an interesting task with increasing attention. PilotNet~\citep{pilotnet} used a small, application-specific network.
We compare \CLL{ }with the baseline PilotNet architecture in Table~\ref{tab:pilot}.
\begin{table}[h]
  {
  \begin{center}
  \setlength\tabcolsep{4pt}
  \caption{End-to-end autonomous driving results on PilotNet dataset (PN) and architecture~\citep{pilotnet,pilotold}.}
  \label{tab:pilot}
  \scriptsize
  \begin{tabular}{L{5.3cm}ccc}
  \toprule
  Approach & Feature extractor & \#Parameters (M) & MAE \\ \midrule
  PilotNet~\citep{pilotnet} & PilotNet & 1.8 & 4.24 $\pm$ 0.45 \\ \hline
  BEL~\citep{ShahICLR2022} & PilotNet & 1.8 & {{3.11}} $\pm$ 0.01 \\ \hline
  \CLL & PilotNet & 1.8 & \textbf{\underline{2.94}} $\pm$ 0.01 \\
  \bottomrule
  \end{tabular}
  \end{center}
  }
\end{table}

\end{document}